\documentclass[twoside,11pt]{article}

 \PassOptionsToPackage{numbers, compress}{natbib}

\usepackage[preprint]{jmlr2e}

\usepackage[multiple]{footmisc}
\usepackage[utf8]{inputenc} 
\usepackage[T1]{fontenc}    
\usepackage{url}            
\usepackage{amsfonts}       
\usepackage{nicefrac}       
\usepackage{microtype}      
\usepackage{bookmark}
\usepackage{latexsym}
\usepackage{appendix}
\usepackage{amsmath}
\usepackage[makeroom]{cancel}
\usepackage{colortbl} 
\usepackage{hyperref} 
\usepackage{graphicx}
\usepackage{array,multirow}
\usepackage{xcolor}
\usepackage{tabularx}
\usepackage{bm}
\usepackage{multirow}
\usepackage{mathtools}
\usepackage{varwidth}
\usepackage{subcaption}
\usepackage{empheq}
\usepackage{wrapfig}
\usepackage{defs}
\usepackage{enumitem}
\usepackage{pifont}
\usepackage{amssymb}
\definecolor{medGray}{RGB}{230,230,230}

\makeatletter
\newcommand{\figcaption}[1]{\def\@captype{figure}\caption{#1}}
\newcommand{\tblcaption}[1]{\def\@captype{table}\caption{#1}}
\makeatother

\title{An Survey of Neural Network Compression}

\author{%
  James T.~O' Neill \\
  Department of Computer Science\\
  University of Liverpool\\
  Liverpool, England, L69 3BX \\
  \texttt{james.o-neill@liverpool.ac.uk} \\
}

\begin{document}

\maketitle

\begin{abstract}
Overparameterized networks trained to convergence have shown impressive performance in domains such as computer vision and natural language processing. Pushing state of the art on salient tasks within these domains corresponds to these models becoming larger and more difficult for machine learning practitioners to use given the increasing memory and storage requirements, not to mention the larger carbon footprint. Thus, in recent years there has been a resurgence in model compression techniques, particularly for deep convolutional neural networks and self-attention based networks such as the Transformer. 

Hence, in this paper we provide a timely overview of both old and current compression techniques for deep neural networks, including pruning, quantization, tensor decomposition, knowledge distillation and combinations thereof. 

\end{abstract}

\newpage
\tableofcontents

\newpage

\section{Introduction}\label{sec:mod_compress}
Deep neural networks (DNN) are becoming increasingly large, pushing the limits of generalization performance and tackling more complex problems in areas such as computer vision (CV), natural language processing (NLP), robotics and speech to name a few. For example, Transformer-based architectures~\citep{vaswani2017attention,sanh2019distilbert,liu2019roberta,yang2019xlnet,lan2019albert,devlin2018bert} that are commonly used in NLP (also used in CV to a less extent~\citep{parmar2018image}) have millions of parameters for each fully-connected layer. Tangentially, Convolutional Neural Network~\citep[(CNN)][]{fukushima1980neocognitron} based architectures~\citep{krizhevsky2012imagenet,he2016identity,zagoruyko2016wide,he2016deep} used in vision and NLP tasks~\citet{kim2014convolutional,hu2014convolutional,gehring2017convolutional}).

\begin{figure}[!bp]
    \centering
    \includegraphics[scale=0.2]{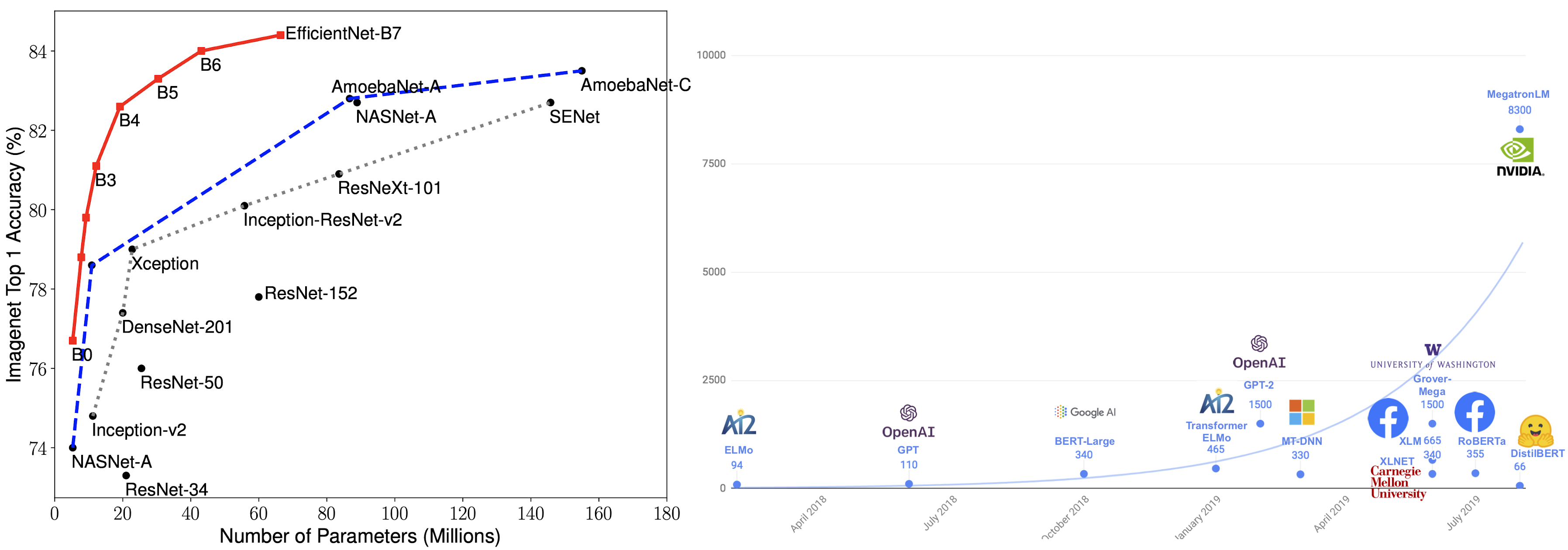}
    \caption{Accuracy vs \# Parameters for CNN architectures (source on left:~\citet{googleai}) and \# Parameters vs Years for Transformers (source on right:~\citet{hug2018face})}
    \label{fig:nets_increasing}
\end{figure}

From the left of \autoref{fig:nets_increasing}, we see that in general, larger overparameterized CNN networks generalize better for ImageNet (a large image classification benchmark dataset). However, recent architectures that aim to reduce the number of floating point operations (FLOPs) and improve training efficiency with less parameters have also shown impressive performance e.g EfficientNet~\citep{tan2019efficientnet}.

The increase in Transformer network size, shown on the right, is more pronounced given that the network consists of fully-connected layers that contain many parameters in each self-attention block~\citep{vaswani2017attention}. 
MegatronLM~\citep{shoeybi2019megatron}, shown on the right-hand side, is a 72-layer GPT-2 model consisting of 8.3 billion  parameters, trained by using 8-way model parallelism and 64-way data parallelism over 512 GPUs.
~\citet{rosset2019turing} proposed a 17 billion parameter Transformer model for natural language text generation (NLG) that consists of 78 layers with hidden size of 4,256 and each block containing 28 attention heads. They use \textit{DeepSpeed}\footnote{A library that allows for distributed training with mixed precision (MP), model parallelism, memory optimization, clever gradient accumulation, loss scaling with MP, large batch training with specialized optimizers, adaptive learning rates and advanced parameter search. See here \url{https://github.com/microsoft/DeepSpeed.git}} with \text{ZeRO}~\citep{rajbhandari2019zero} to eliminate memory redundancies in data parallelism and model parallelism and allow for larger batch sizes (e.g 512), resulting in three times faster training and less GPUs required in the cluster (256 instead of 1024). 
~\citet{brown2020language} the most recent Transformer to date, trains a GPT-3 autoregressive language model that contains 175 billion parameters. This model can perform NLP tasks (e.g machine translation, question-answering) and digit arithmetic relatively well with only few examples, closing the performance gap to similarly large pretrained models that are further fine-tuned for specific tasks and in some cases outperforming them given the large increase in the number of parameters. 
The resources required to store the aforementioned CNN and Transformer models on Graphics Processing Units (GPUs) or Tensor Processing Units (TPUs) let alone train them is out of reach for a large majority of machine learning practitioners. Moreover, these models have predominantly been driven by improving the state of the art (SoTA) and pushing the boundaries of what complex tasks can be solved using them. Therefore, we expect that the current trend of increasing network size will remain.

Thus, the motivation to compress models has grown and expanded in recent years from being predominantly focused around deployment on mobile devices, to also learning smaller networks on the same device but with eased hardware constraints i.e learning on a small number of GPUs and TPUs or the same number of GPUs and TPUs but with a smaller amount of VRAM. For these reasons, model compression can be viewed as a critical research endeavour to allow the machine learning community to continue to deploy and understand these large models with limited resources.  


Hence, this paper provides an overview of methods and techniques for compressing DNNs. This includes weight sharing (\autoref{sec:ws}), pruning (\autoref{sec:capn}), tensor decomposition (\autoref{sec:lrtd}), knowledge distillation (\autoref{sec:md}) and quantization (\autoref{sec:quantize}).

\subsection{Further Motivation for Compression}
A question that may naturally arise at this point is - \textit{Can we obtain the same or similar generalization performance by training a smaller network from scratch to avoid training a larger teacher network to begin with ?} 

Before the age of DNNs,~\citet{castellano1997iterative} found that training a smaller shallow network from random initialization has poorer generalization compared to an equivalently sized pruned network from a larger pretrained network.

More recently,~\citet{zhu2017prune} have too addressed this question, specifically in the case of pruning DNNs - that is, whether many previous work that previously report large reductions in pretrained networks using pruning were already severely overparameterized and could be achieved by simply training a smaller model equivalent in size to the pruned network. They find for deep CNNs and LSTMs that large sparsely pruned networks consistently outperform smaller dense models, achieving a compression ratio of 10 in the number of non-zero parameters with minuscule losses in accuracy. 
Even when the pruned network is not necessarily overparameterized in the pretraining stage, it still produces consistently better performance than an equivalently sized network trained from scratch~\citep{lin2017runtime}.

Essentially, having a DNN with larger capacity (i.e more degrees of freedom) allows for a larger set of solutions to be chosen from in the parameter space.  Overparameterization has also shown to have a smoothening effect on the loss space~\citep{li2018visualizing} when trained with stochastic gradient descent (SGD)~\citep{du2018gradient,cooper2018loss}, in turn producing models that generalize better than smaller models. This has been reinforced recently for DNNs after the discovery of the double descent phenomena~\citep{belkin2019two},
whereby a $2^{nd}$ descent in the test error is found for overparameterized DNNs that have little to no training errors, occurring after the \textit(critical regime) region where the test error is initially high. This $2^{nd}$ descent in test error tends to converge to an error lower than that found in the $1^{st}$ descent where the $1^{st}$ descent corresponds to the traditional bias-variance tradeoff. Moreover, the norm of the weights becomes dramatically smaller in each layer during this $2^{nd}$ descent, during the compression phase~\citep{shwartz2017opening}. Since the weights tend to be close to zero when trained far into this $2^{nd}$ region, it becomes clearer why compression techniques, such as pruning, has less effect on the networks behaviour when trained to convergence since the magnitude of individual weights becomes smaller as the network grows larger. 

~\citet{frankle2018lottery} also showed that training a network to convergence with more parameters makes it easier to find a subnetwork that when trained from scratch, maintains performance, further suggesting that compressing large pretrained overparameterized DNNs that are trained to convergence has advantages from performance and storage perspective over training and equivalently smaller DNN. Even in cases when the initial compression causes a degradation in performance, retraining the compressed model can and is commonly carried out to maintain performance. 

%
Lastly, large pretrained models are widely and publicly available\footnote{pretrained CNN models:\url{https://github.com/Cadene/pretrained-models.pytorch}}\footnote{pretrained Transformer models:\url{https://github.com/huggingface/transformers}} and thus can be easily used and compared by the rest of the machine learning community, avoiding the need to train these models from scratch. This further motivates the utility of model compression and its advantages over training equivalently smaller network from scratch. 

\subsection{Categorizations of Compression Techniques}
Retraining is often required to account for some performance loss when applying compression techniques. The retraining step can be carried out using unsupervised (including self-supervised) learning (e.g tensor decomposition) and supervised learning (knowledge distillation). Unsupervised compression is often used when there is are no particular target task/s that the model is being specifically compressed for, alternatively supervision can be used to gear the compressed model towards a subset of tasks in which case the target task labels are used as opposed to the original data the model was trained on, unlike unsupervised model comrpression. In some cases, RLg has also shown to be beneficial for maintaining performance during iterative pruning~\citep{ lin2017runtime}, knowledge distillation~\citep{ashok2017n2n} and quantization~\citep{yazdanbakhsh2018releq}.


\subsection{Compression Evaluation Metrics}
Lastly, we note that the main evaluation of compression techniques is performance metric (e.g accuracy) vs model size. When evaluating for speedups obtained from the model compression, the number of floating point operations (FLOPs) is a commonly used metric. When claims of storage improvements are made, this can be demonstrated by reporting the run-time memory footprint which is essentially the ratio of the space for storing hidden layer features during run time when compared to the original network.

We now begin to describe work for each compression type, beginning with weight sharing. 

\section{Weight Sharing}\label{sec:ws}
The simplest form of network reduction involves sharing weights between layers or structures within layers (e.g filters in CNNs). We note that unlike compression techniques discussed in later sections (Section 3-6), standard weight sharing is carried out prior to training the original networks as opposed to compressing the model after training. However,recent work which we discuss here~\citep{chen2015compressing,ullrich2017soft,bai2019deep} have also been used to reduce DNNs post-training and hence we devote this section to this straightforward and commonly used technique.

Weight sharing reduces the network size and avoids sparsity. It is not always clear how many and what group of weights should be shared before there is an unacceptable performance degradation for a given network architecture and task. For example,~\citet{inan2016tying} find that tying the input and output representations of words leads to good performance while dramatically reducing the number of parameters proportional to the size of the vocabulary of given text corpus. Although, this may be specific to language modelling, since the output classes are a direct function of the inputs which are typically very high dimensional (e.g typically greater than $10^6$). Moreover, this approach assigns the embedding matrix to be shared, as opposed to sharing individual or sub-blocks of the matrix. Other approaches include clustering weights such that their centroid is shared among each cluster and using weight penalty term in the objective to group weights in a way that makes them more amenable to weight sharing. We discuss these approaches below along with other recent techniques that have shown promising results when used in DNNs. 


\subsection{Clustering-based Weight Sharing}
~\citet{nowlan1992simplifying} instead propose a soft weight sharing scheme by learning a Gaussian Mixture Model that assigns groups of weights to a shared value given by the mixture model. By using a mixture of Gaussians, weights with high magnitudes that are centered around a broad Gaussian component are under less pressure and thus penalized less. In other words, a Gaussian that is assigned for a subset of parameters will force those weights together with lower variance and therefore assign higher probability density to each parameter. 

\autoref{eq:gmm_cost} shows the cost function for the Gaussian mixture model where $p(w_j)$ is the probability density of a Gaussian component with mean $\mu_j$ and standard deviation $\sigma_j$. Gradient Descent is used to optimize $w_i$ and mixture parameters $\pi_j$, $\mu_j$, $\sigma_j$ and $\sigma_y$.

\begin{equation}\label{eq:gmm_cost}
    C = \frac{K}{\sigma^2_y} \sum_c (y_c - d_c)^2 - \sum_i \log \Big[\sum_j \pi_j p_j (w_i)\Big]
\end{equation}

The expectation maximization (EM) algorithm is used to optimize these mixture parameters. The number of parameters tied is then proportional to the number of mixture components that are used in the Gaussian model. 

\paragraph{An Extension of Soft-Weight Sharing}
~\citet{ullrich2017soft} build on soft-weight sharing~\citep{nowlan1992simplifying} with factorized posteriors by optimizing the objective in \autoref{eq:var_compress}.
Here, $\tau=5e^{-3}$ controls the influence of the log-prior means $\mu$, variances $\sigma$ and mixture coefficients $\pi$, which are learned during retraining apart from the j-th component that are set to $\mu_j=0$ and $\pi_j=0.99$. Each mixture parameter has a learning rate set to $5 \times 10^{-4}$. Given the sensitivity of the mixtures to collapsing if the correct hyperparameters are not chosen, they also consider the inverse-gamma hyperprior for the mixture variances that is more stable during training.

\begin{equation}\label{eq:var_compress}
\mathcal{L}\big(w,\{\mu_j,\sigma_j,\pi_j \}_{j=0}^J \big) =\mathcal{L}_E +  \tau \mathcal{L}_C = - \log p \big(\tau| X, w \big) - \tau \log p\big(w,\{\mu_j,\sigma_j, \pi_j \}_{j=0}^J \big)
\end{equation}

After training with the above objective, if the components have a KL-divergence under a set threshold, some of these components are merged~\citep{adhikari2012multiresolution} as shown in \autoref{eq:kl_merge}. Each weight is set then set to the mean of the component with the highest mixture value $\argmax(\pi)$, performing GMM-based quantization.  

\begin{equation}\label{eq:kl_merge}
\pi_{\text{new}} = \pi_i + \pi_j,\quad   \mu_{\text{new}} = \frac{\pi_i \mu_i+\pi_j \mu_j}{\pi_i+\pi_j},\quad   \sigma^2_{\text{new}}= \frac{\pi_i\sigma^2_i+\pi_j \sigma^2_j}{\pi_i+\pi_j}
\end{equation}

In their experiments, 17 Gaussian components were merge to 6 quantization components, while still leading to performance close to the original LeNet classifier used on MNIST. 


\subsection{Learning Weight Sharing}
~\citet{zhang2018learning} explicitly try to learn which weights should be shared by imposing a group order weighted $\ell_1$ (GrOWL) sparsity regularization term while simultaneously learning to group weights and assign them a shared value. In a given compression step, groups of parameters are identified for weight sharing using the aforementioned sparsity constraint and then the DNN is retrained to fine-tune the structure found via weight sharing. GrOWL first identify the most important weights and then clusters correlated features to learn the values of the closest important weight throughout training. This can be considered an adaptive weight sharing technique. 

~\citet{plummer2020shapeshifter} learn what parameters groupings to share and can be shared for layers of different size and features of different modality. They find parameter sharing with distillation further improves performance for image classification, image-sentence retrieval and phrase grounding.

\paragraph{Parameter Hashing} ~\citet{chen2015compressing} use hash functions to randomly group weight connections into hash buckets that all share the same weight value. Parameter hashing~\citep{weinberger2009feature,shi2009hash} can easily be used with backpropogation whereby each bucket parameters have subsets of weights that are randomly i.e each weight matrix contains multiple weights of the same value (referred to as a \textit{virtual matrix}), unlike standard weight sharing where all weights in a matrix are shared between layers.

\subsection{Weight Sharing in Large Architectures}
\paragraph{Applications in Transformers}
\citet{dehghani2018universal} propose Universal Transformers (UT) to combine the benefits of recurrent neural networks~\citep[(RNNs)][]{rumelhart1985learning,hochreiter1997long} (recurrent inductive bias) with Transformers~\citep{vaswani2017attention} (parallelizable self-attention and its global receptive field). As apart of UT, weight sharing to reduce the network size showed strong results on NLP defacto benchmarks while . 

~\citet{dabre2019recurrent} use a 6-hidden layer Transformer network for neural machine translation (NMT) where the same weights are fed back into the same attention block recurrently. This straightforward approach surprisingly showed similar performance of an untied 6-hidden layer for standard NMT benchmark datasets. 

~\citet{xiao2019sharing} use shared attention weights in Transformer as dot-product attention can be slow during the auto-regressive decoding stage. Attention weights from hidden states are shared among adjacent layers, drastically reducing the number of parameters proportional to number of attention heads used. The Jenson-Shannon (JS) divergence is taken between self-attention weights of different heads and they average them to compute the average JS score. They find that the weight distribution is similar for layers 2-6 but larger variance is found among encoder-decoder attention although some adjacent layers still exhibit relatively JS score. Weight matrices are shared based on the JS score whereby layers that have JS score larger than a learned threshold (dynamically updated throughout training) are shared. The criterion used involves finding the largest group of attention blocks that have similarity above the learned threshold to maximize largest number of weight groups that can be shared while maintaining performance. They find a 16 time storage reduction over the original Transformer while maintaining competitive performance. 

\paragraph{Deep Equilibrium Model}
~\citet{bai2019deep} propose deep equilibrium models (DEMs) that use a root-finding method to find the equilibrium point of a network and can be analytically backpropogated through at the equilibrium point using implicit differentiation. This is motivated by the observation that hidden states of sequential models converge towards a fixed point.  Regardless of the network depth, the approach only requires constant memory because backpropogration only needs to be performed on the layer of the equilibrium point.  

For a recurrent network $f_{\mat{W}}(\vec{z}^*_{1:T};\vec{x}_{1:T})$ of infinite hidden layer depth that takes inputs $\vec{x}_{1:T}$ and hidden states $\vec{z}_{1:T}$ up to $T$ timesteps, the transformations can be expressed as,

\begin{equation}
    \lim_{i \to \infty}  z_{1:T}^{[i]} = \lim_{i \to \infty} f_{W} (\vec{Z}_{1:T}^{[i]}; \vec{x}_{1:T}) := f_{W}(\vec{z}^*_{1:T}; x_{1:T}) = \underbrace{\vec{z}^*_{1:T}}_{\text{equilibrium point}}
\end{equation}

where the final representation $\vec{z}^*_{1:T}$ is the hidden state output corresponds to the equilibrium point of the network. They assume that this equilibrium point exists for large models, such as Transformer and Trellis~\citep{bai2018trellis} networks (CNN-based architecture).

The $\frac{\partial \vec{z}^*_{1:T}}{\partial \mat{W}}$ requires implicit differentiation and \autoref{eq:deq_2} can be rewritten as \autoref{eq:deq_3}.

\begin{equation}\label{eq:deq_2}
    \frac{\partial \vec{z}^*_{1:T}}{\partial \mat{W}} =  \frac{d f_{W}(\vec{z}^*_{1:T}; \vec{x}_{1:T})}{d  \mat{W}} + \frac{\partial f_{W}(\vec{z}^*_{1:T}; \vec{x}_{1:T})}{\partial \vec{z}^*_{1:T}}  \frac{\partial \vec{z}^*_{1:T}}{\partial \mat{W}}  
\end{equation}

\begin{equation}\label{eq:deq_3}
    \Big(I -  \frac{\partial f_{W}(\vec{z}^*_{1:T}; \vec{x}_{1:T})}{\partial \vec{z}^*_{1:T}} \Big) \frac{\partial \vec{z}^*_{1:T}}{\partial \mat{W}} =  \frac{d f_{W}(\vec{z}^*_{1:T}; \vec{x}_{1:T})}{d\  \mat{W}}
\end{equation}

For notational convenience they define $g_{\mat{W}}(z^{*}_{1:T}; \vec{x}_{1:T}) =f_{\mat{W}}(z^{*}_{1:T}; \vec{x}_{1:T}) - \vec{z}^{*}_{1:T} \to 0$ and thus the equilibrium state $\vec{z}^{*}_{1:T}$ is thus the root of $g_{\mat{W}}$ found by the Broyden's method~\citep{broyden1965class}\footnote{A quasi-Newton method for finding roots of a parametric model.}.

The Jacobian of the function $g_{\mat{W}}$ at the equilibrium point $\vec{z}^*_{1:T}$ w.r.t $\mat{W}$ can then be expressed as \autoref{eq:deq_5}. Note that this is computed without having to consider how the equilibrium $\vec{z}^*_{1:T}$ was obtained.

\begin{equation}\label{eq:deq_5}
    \mat{J}_{g_\mat{W}}\Big|_{\vec{z}^*_{1:T}} = - \Big(I -  \frac{\partial f_{W}(\vec{z}^*_{1:T}; \vec{x}_{1:T})}{\partial \vec{z}^*_{1:T}} \Big)
\end{equation}

Since $f_{\mat{W}}(\cdot)$ is in equilibrium at $\vec{z}^*_{1:T}$ they do not require to backpropogate through all the layers, assuming all layers are the same (this is why it is considered a weight sharing technique). They only need to solve \autoref{eq:deq_6} to find the equilibrium points using Broydens method, 

\begin{equation}\label{eq:deq_6}
   \frac{\partial \vec{z}^{*}_{1:T}}{\partial \mat{W}}  = -  \mat{J}_{g_\mat{W}}\Big|_{\vec{z}^*_{1:T}} \frac{d\ f_{\mat{W}}(\vec{z}^*_{1:T}; \vec{x}_{1:T})}{d  \mat{W}}
\end{equation}

and then perform a single layer update using backpropogation at the equilibrium point. 
\begin{equation}\label{eq:dem_7}
    \frac{\partial \cL}{\partial \mat{W}} = \frac{\cL}{\partial \vec{z}^*_{1:T}} \frac{\partial \vec{z}^*_{1:T}}{\partial \mat{W}} = - \frac{\partial \cL}{\partial \vec{z}^*_{1:T}\Big(\mat{J}^{-1}_{g_\mat{W}}\big|_{\vec{z}^*_{1:T}}\Big)} \frac{d\ f_{\mat{W}}(\vec{z}^*_{1:T}; \vec{x}_{1:T})}{d \mat{W}}
\end{equation}

The benefit of using Broyden method is that the full Jacobian does not need to be stored but instead an approximation $\hat{\mat{J}}^{-1}$ using the Sherman-Morrison formula~\citep{scellier2017equilibrium} which can then be used as apart of the Broyden iteration:

\begin{equation}
    \vec{z}_{1:T}^{[i+1]} := \vec{z}_{1:T}^{[i]} - \alpha \hat{\mat{J}}^{-1}_{g_\mat{W}}\Big|_{\vec{z}_{1:T}^{[i]}} g_{\mat{W}}(\vec{z}_{1:T}^{[i]}; \vec{x}_{1:T}) \quad \text{for} \quad i = 0, 1, 2, \ldots
\end{equation}

where $alpha$ is the learning rate. This update can then be expressed as \autoref{eq:dem_8}

\begin{equation}\label{eq:dem_8}
    \mat{W}^+ = \mat{W} - \alpha \cdot \frac{\partial \cL}{\partial \mat{W}} = \mat{W} + \alpha \frac{\partial \cL}{\partial \vec{z}*_{1:T}\Big(\mat{J}^{-1}_{g_\mat{W}}\big|_{\vec{z}^{*}_{1:T}}\Big)} \frac{d\ f_{\mat{W}}(\vec{z}^{*}_{1:T}; \vec{x}_{1:T})}{d \mat{W}}
\end{equation}

\begin{figure}
    \centering
    \includegraphics[scale=0.38]{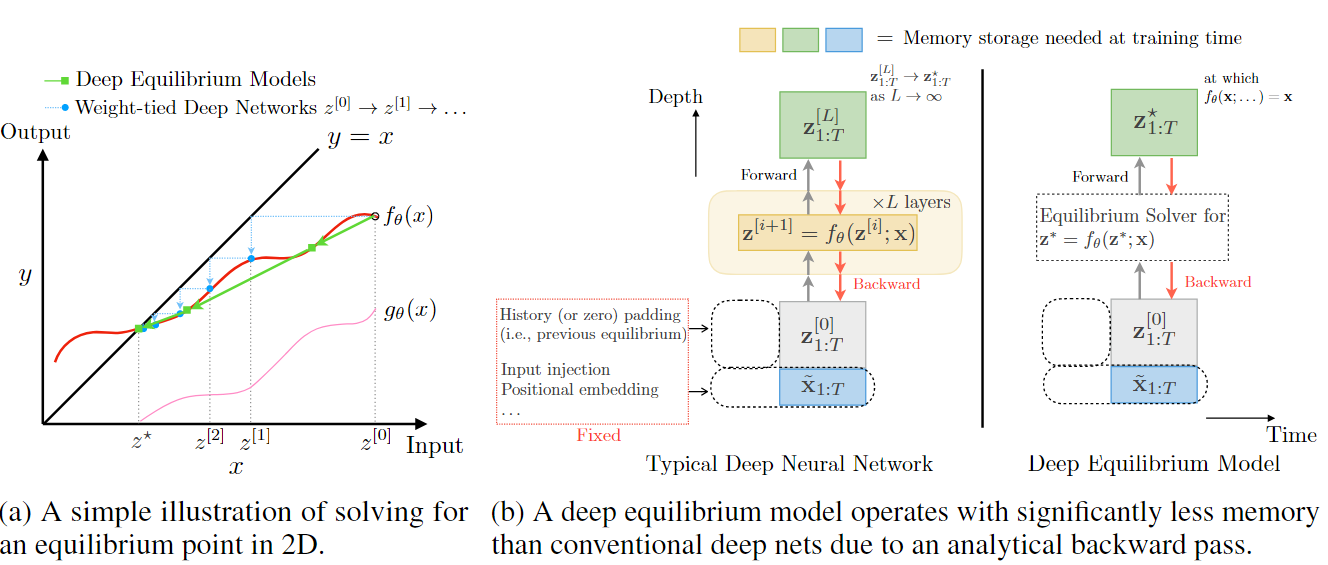}
    \caption{original source~\citet{bai2019deep}: Comparison of the DEQ with conventional weight-tied deep networks}
    \label{fig:dem}
\end{figure}

\autoref{fig:dem} shows the difference between a standard Transformer network forward pass and backward pass in comparison to DEM passes. The left figure illustrates the Broyden iterations to find the equilibrium point for inputs over successive inputs. On WikiText-103, they show that DEMs can improve SoTA sequence models and reduce memory by 88\% use for similar computational requirements as the original models. 

\subsection{Reusing Layers Recursively}
Recursively re-using layers is another form of parameter sharing. This involves feeding the output of a layer back into its input. 

~\citet{eigen2013understanding} have used recursive layers in CNNs and analyse the effects of varying the number of layers, features maps and parameters independently. They find that increasing the number of layers and number of parameters are the most significant factors while increasing the number of feature maps (i.e the representation dimensionality) improves as a byproduct of the increase in parameters. From this, they conclude that adding layers without increasing the number of parameters can increase performance and that the number of parameters far outweights the feature map dimensions with respect to performance. 

~\citet{kopuklu2019convolutional} have also focused on reusing convolutional layers using recurrency applying batch normalization after recursed layers and channel shuffling to allow filter outputs to be passed as inputs to other filters in the same block. By channel shuffling, the LRU blocks become robust with dealing with more than one type of channel, leading to improved performance without increasing the number of parameters. 
~\citet{savarese2019learning} learn a linear combination of parameters from an external group of templates. They too use recursive convolutional blocks as apart of the learned parameter shared weighting scheme. 

However, layer recursion can lead to \textit{vanishing or exploding gradients} (VEGs). Hence, we concisely describe previous work that have aimed to mitigate VEGs in parameter shared networks, namely ones which use the aforementioned recursivity.  

~\citet{kim2016deeply} have used residual connections between the input and the output reconstruction layer to avoid signal attenuation, which can further lead to vanishing gradients in the backward pass. This is applied in the context self-supervision by reconstructing high resolution images for image super-resolution. 
~\citet{tai2017image} extend the work of~\citet{kim2016deeply}. Instead of passing the intermediate outputs of a shared parameter recursive block to another convolutional layer, they use an elementwise addition of the intermediate outputs of the residual recursive blocks before passing to the final convolutional layer. The original input image is then added to the output of last convolutional layer which corresponds to the final representation of the recursive residual block outputs. 

~\citet{zhang2018residual} combine residual (skip) connections and dense connections, where skip connections add the input to each intermediate hidden layer input.

~\cite{guo2019dynamic} address VGs in recursive convolutional blocks by using a gating unit that chooses the number of self-loops for a given block before VEGs occur. They use the Gumbel Softmax trick without gumbel noise to make deterministic predictions of the number of self-loops there should be for a given recursive block throughout training. They also find that batch normalization is at the root of gradient explosion because of the statistical bias induced by having a different number of self-loops during training, effecting the calculation of the moving average. This is adressed by normalizing inputs according to the number of self-loops which is dependent on the gating unit. When used in Resnet-53 architecture, dynamically recursivity outperforms the larger ResNet-101 while reducing the number parameters by 47\%.


\section{Network Pruning}\label{sec:capn}
Pruning weights is perhaps the most commonly used technique to reduce the number of parameters in a pretrained DNN. Pruning can lead to a reduction of storage and model runtime and performance is usually maintaining by retraining the pruned network. Iterative weight pruning prunes while retraining until the desired network size and accuracy tradeoff is met. From a neuroscience perspective, it has been found that as humans learn they also carry out a similar kind of iterative pruning, removing irrelevant or unimportant information from past experiences~\citep{walsh2013peter}. Similarly, pruning is not carried out at random, but selected so that unimportant information about past experiences is discarded. In the context of DNNs, random pruning (akin to Binary Dropout) can be detrimental to the models performance and may require even more retraining steps to account for the removal of important weights or neurons~\citep{yu2018nisp}.

The simplest pruning strategy involves setting a threshold $\gamma$ that decides which weights or units (in this case, the absolute sum of magnitudes of incoming weights) are removed~\citep{hagiwara1994removal}. The threshold can be set based on each layers weight magnitude distribution, where weights centered around the mean $\mu$ are removed, or it the threshold can be set globally for the whole network. Alternatively, pruning the weights with lowest absolute value of the normalized gradient multiplied by the weight magnitude~\citep{lee2018snip} for a given set of mini-batch inputs can be used, either layer-wise or globally too. 

Instead of setting a threshold, one can predefine a percentage of weights to be pruned based on the magnitude of $w$, or a percentage aggregated by weights for each layer $w_{l},\ \forall l \in L$.  Most commonly, the percentage of weights that are closest to 0 are removed. The aforementioned criteria for pruning are all types of \textit{magnitude-based pruning} (MBP). MBP has also been combined with other strategies such as adding new neurons during iterative pruning to further improve performance~\citep{han2013structure,narasimha2008integrated}, where the number of new neurons added is less than the number pruned in the previous pruning step and so the overall number of parameters monotonically decreases. 

MBP is the most commonly used in DNNs due to its simplicity and performs well for a wide class of machine learning models (including DNNs) on a diverse range of tasks~\citep{setiono2000pruned}. In general, global MBP tends to outperform layer-wise MBP~\citep{karnin1990simple,reed1993pruning,hagiwara1994removal,lee2018snip}, because there is more flexibility on the amount of sparsity for each layer, allowing more salient layer to be more dense while less salient to contain more non-zero entries.
Before discussing more involved pruning methods, we first make some important categorical distinctions.



\subsection{Categorizing Pruning Techniques}
Pruning algorithms can be categorized into those that carry out pruning without retraining the pruning and those that do. Retraining is often required when pruning degrades performance. This can happen when the DNN is not necessarily overparameterized, in which case almost all parameters are necessary to maintain good generalization. 

Pruning techniques can also be categorized into what type of criteria is used as follows:

\begin{enumerate}
    \item The aforementioned magnitude-based pruning whereby the weights with the lowest absolute value of the weight are removed based on a set threshold or percentage, layer-wise or globally. 

    \item Methods that penalize the objective with a regularization term to force the model to learn a network with (e.g $\ell_1$, $\ell_2$ or lasso weight regularization) smaller weights and prune the smallest weights. 
    \item Methods that compute the sensitivity of the loss function when weights are removed and remove the weights that result in the smallest change in loss.
    \item Search-based approaches (e.g particle filters, evolutionary algorithms, reinforcement learning) that seek to learn or adapt a set of weights to links or paths within the neural network and keep those which are salient for the task. Unlike (1) and (2), the pruning technique does not involve gradient descent as apart of the pruning criteria (with the exception of using deep RL).
\end{enumerate}

\paragraph{Unstructured vs Structured Pruning}
Another important distinction to be made is that between structured and unstructured pruning techniques where the latter aims to preserve network density for computational efficiency (faster computation at the expense of less flexibility) by removing groups of weights, whereas unstructured is unconstrained to which weights or activations are removed but the sparsity means that the dimensionality of the layers does not change.
Hence, sparsity in unstructured pruning techniques provide good performance at the expense of slower computation. For example, MBP produces a sparse network that requires sparse matrix multiplication (SMP) libraries to take full advantage of the memory reduction and speed benefits for inference. However, SMP is generally slower than dense matrix multiplication and therefore there has been work towards preserving subnetworks which omit the need for SMP libraries (discussed in \autoref{sec:struct_prune}).

With these categorical distinctions we now move on to the following subsections that describe various pruning approaches beginning with pruning by using weight regularization. 


\subsection{Pruning using Weight Regularization}\label{eq:weight_reg}
Constraining the weights to be close to 0 in the objective function by adding a penalty term and deleting the weights closest to 0 post-training can be a straightforward yet effective pruning approach.~\autoref{eq:penalty_term_1} shows the commonly used $\ell_2$ penalty that penalizes large weights $w_m$ in the $m$-th hidden layer with a large magnitude and $\vec{v}_m$ are the output layer weights of output dimension $C$.

\begin{equation}\label{eq:penalty_term_1}
    C(\vec{w}, \vec{v}) = \frac{\epsilon}{2} \Big(\sum_{m=1}^{h}\sum_{l=1}^n \vec{w}^2_{m l} + \sum_{m=1}^{h}\sum_{p=1}^C \vec{v}^2_{pm}\Big) 
\end{equation}

However, the main issue with using the above quadratic penalty is that all parameters decay exponentially at the same rate and disproportionately penalizes larger weights. Therefore,~\citet{weigend1991generalization} proposed the objective shown in \autoref{eq:weig}.  When $f(w):= w^2/(1 + w^2)$ this penalty term is small and when large it tends to 1. Therefore, these terms can be considered as approximating the number of non-zero parameters in the network. 

\begin{equation}\label{eq:weig}
    C(\vec{w}, \vec{v}) = \frac{\epsilon}{2} \Big(\sum_{m=1}^{h}\sum_{l=1}^n \frac{\vec{w}^2_{ml}}{ 1 +\vec{w}^2_{ml}} + \sum_{m=1}^{h} \sum_{p=1}^C \frac{\vec{v}^2_{pm}}{ 1 + \vec{v}^2_{pm}}\Big)
\end{equation}

The derivative $f'(w) = 2\vec{w}/(1 + \vec{w}^2)^2$ computed during backprogation does not penalize large weights as much as \autoref{eq:penalty_term_1}. However, in the context of recent years where large overparameterized network have shown better generalization when the weights are close to 0, we conjecture that perhaps \autoref{eq:weig} is more useful in the underparameterized regime. The $\epsilon$ controls how the small weights decay faster than large weights. However, the problem of not distinguishing between large weights and very large weights is also an issue. Therefore,~\citet{weigend1991generalization} further propose the objective in \autoref{eq:penalty_term_3}.

\begin{equation}\label{eq:penalty_term_3}
    C(\vec{w}, \vec{v}) = \epsilon_1 \sum_{m=1}^{h} \Big(\sum_{l=1}^n \frac{\beta \vec{w}^2_{ml}}{ 1 + \beta \vec{w}^2_{ml}} + \sum_{p=1}^C \frac{\beta \vec{v}^2_{pm}}{ 1 + \beta \vec{v}^2_{pm}}\Big) + \epsilon_2 \sum_{m=1}^h \Big(\sum_{l=1}^n \vec{w}^2_{ml} + \sum_{p=1}^C \vec{v}_{pm}^2 \Big)
\end{equation}

~\citet{wan2009enhancing} have proposed a Gram-Schmidth (GS) based variant of backpropogation whereby GS determines which weights are updated and which ones remain frozen at each epoch. 

\begin{figure}
    \centering
    \includegraphics[scale=0.4]{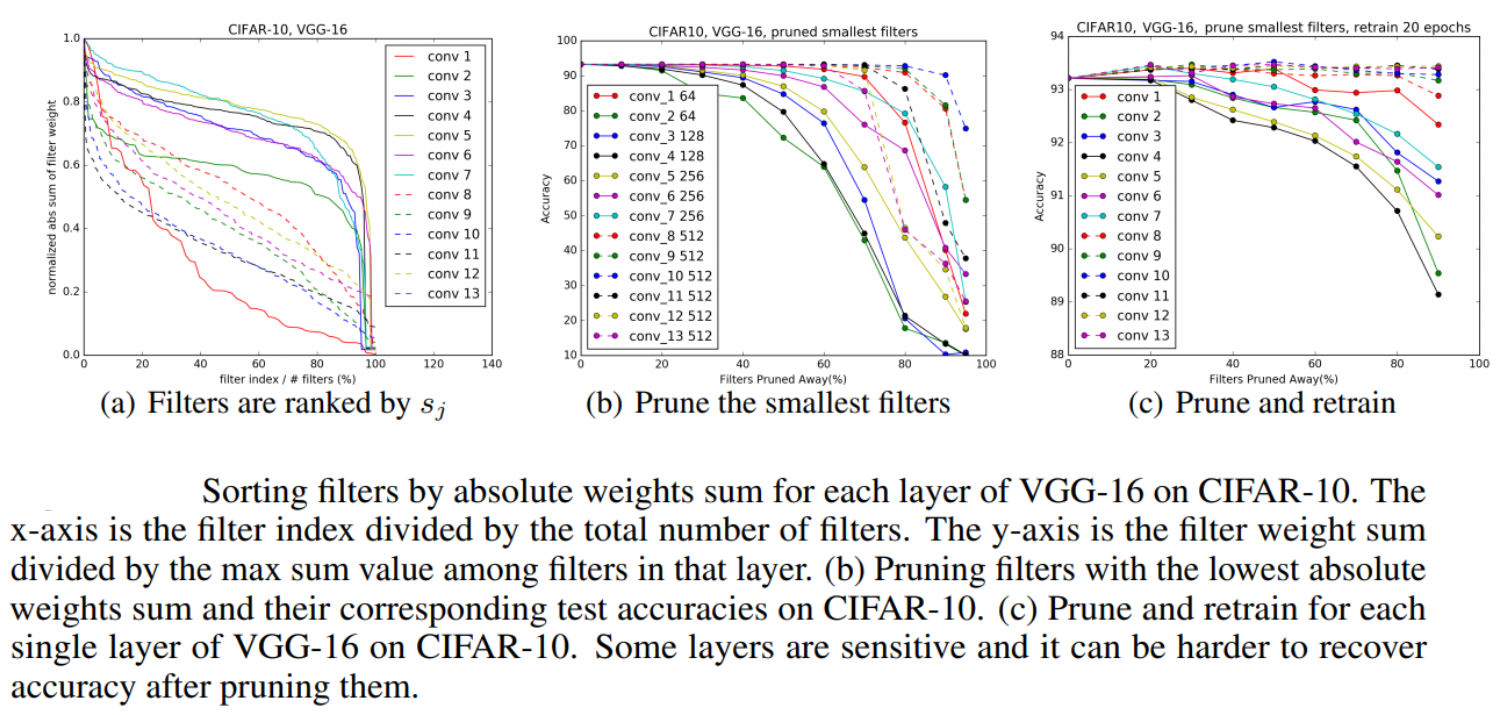}
    \caption{original source:~\citet{li2016pruning}}
    \label{fig:rank_pruned_cnn_filters}
\end{figure}

~\citet{li2016pruning} prune filters in CNNs by identifying filters which  contribute least to the overall accuracy. For a given layer, sum of the weight magnitudes are computed and since the number of channels is the same across filters, this quantity represents the average of weight value for each kernel.
Kernels with weights that have small weight activations will have weak activation and hence these will be pruned. This simple approach leads to less sparse connections and leads to 37\% accuracy reduction on average across the models tested while still being close to the original accuracy.~\autoref{fig:rank_pruned_cnn_filters} shows their figure that demonstrates that pruning filters that have the lowest sum of weight magnitudes correspond to the best maintaining of accuracy. 

\subsection{Pruning via Loss Sensitivity}
Networks can also be pruned by measuring the importance of weights or units by quantifying the change in loss when a weight or unit is removed and prune those which cause the least change in the loss.  Many methods from previous decades have been proposed based on this principle ~\citep{reed1993pruning,lecun1990optimal,hassibi1994optimal}. We briefly describe each one below in chronological order.

\paragraph{Skeletonization}
~\citet{mozer1989skeletonization} estimate which units are least important and deletes them during training.  The method is referred to as skeletonization, since it only keeps the units which preserve the main structure of the network that is required for maintaining good out-of-sample performance. Each weight $w$ in the network is assigned an importance weight $\alpha$ where $alpha=0$ the weight becomes redundant and $\alpha=1$ the weight acts as a standard hidden unit.

To obtain the importance weight for a unit, they calculate the loss derivative with respect to $\alpha$ as $\hat{\rho}_i = \partial \mathcal{L} / {\alpha_i}\big|_{\alpha_i = 1}$ where $\mathcal{L}$ in this context is the sum of squared errors. Units are then pruned when $\hat{\rho}_i$ falls below a set threshold. However, they find that $\hat{\rho}_i$ can fluctuate throughout training and so they propose an exponentially-decayed moving average over time to smoothen the volatile gradient and also provide better estimates when the squared error is very small. This moving average is given as,

\begin{equation}\label{eq:fluct_sens}
    \hat{\rho}_i(t + 1) = \beta \hat{\rho}_i(t) + (1 - \beta) \frac{\partial \mathcal{L}(t)}{\alpha_i} 
\end{equation}

where $\beta = 0.8$ in their experiments. Applying skeletonization to current DNNs  is perhaps be too slow to compute as it was originally introduced in the context of using neural networks with a relatively small amount of parameters. However, assigning importance weights for groups of weights, such as filters in a CNN is feasible and aligns with current literature~\citep{wen2016learning,anwar2017structured} on structured pruning (discussed in ~\autoref{sec:struct_prune}).

\paragraph{Pruning Weights with \textit{Low Sensitivity}}

~\citet{karnin1990simple} measure the sensitivity $S$ of the loss function with respect to weights and prune weights with low sensitivity. Instead of removing each weight individually, they approximate $S$ by the sum of changes experienced by the weight during training as

\begin{equation}\label{eq:simple_sensitivity_2}
    S_{ij} = \Big|- \sum_{n=0}^{N-1} \frac{\partial \mathcal{L} }{\partial \vec{w}_{ij}}\Delta \vec{w}_{ij}(n) \frac{\vec{w}^f_{ij}}{\vec{w}^f_{ij} - \vec{w}^i_{ij}} \Big|
\end{equation}

where $w^f$ is the final weight value at each pruning step, $w^i$ is the initial weight after the previous pruning step and $N$ is the number of training epochs. Using backpropagation to compute $\Delta w$, $\hat{S}_{ij}$ is expressed as,

\begin{equation}\label{eq:simple_sensitivity_3}
    \hat{S}_{ij} = \Big|- \sum_{n=0}^{N-1}\big[\Delta \vec{w}_{ij}(n)\big]^2 \frac{\vec{w}^f_{ij}}{\nabla (\vec{w}^f_{ij} - \vec{w}^i_{ij})} \Big|
\end{equation}

If the sum of squared errors is less than that of the previous pruning step and if a weight in a hidden layer with the smallest $S_{ij}$ changes less than the previous epoch, then these weights are pruned. This is to ensure that weight with small initial sensitivity are not pruned too early, as they may perform well given more retraining steps. If all incoming weights are removed to a unit, the unit is also removed, thus, removing all outgoing weights from that unit. Lastly, they lower bound the number of weights that can be pruned for each hidden layer, therefore, towards the end of training there may be weights with low sensitivity that remain in the network.  

\paragraph{Variance Analysis on Sensitivity-based Pruning}
~\citet{engelbrecht2001new} remove weights if its variance in sensitivity is not significantly different from zero.  If the variance in parameter sensitivities is not significantly different from zero and the average sensitivity is small, it indicates that the corresponding parameter has little or no effect on the output of the NN over all patterns considered. A hypothesis testing step then uses these variance nullity measures to statistically test if a parameter should be pruned, using the distribution.What needs to be done is to test if the expected value of the sensitivity of a parameter over all patterns is equal to zero. The expectation can be written as
$\mathcal{H}_0: \langle S_{oW, ki} \rangle^2 = 0$ where $S_{oW}$ is the sensitivity matrix of the output vector with respect to the parameter vector $\vec{W}$ and individual elements $ S_{oW, ki}$ refers to the sensitivity of output to perturbations in parameter over all samples. If the hypothesis is accepted, prune the corresponding weight at the $(k, i)$ position, otherwise check $\mathcal{H}_0: \text{var}(S_{oW, ki}) = 0$ and if this accepted also opt to prune it. They test sum-norm, Euclidean-norm and maximum-norm to compute the output sensitivity matrix. They find that this scheme finds smaller networks than OBD, OBS and standard magnitude-based pruning while maintaining the same accuracy on multi-class classification tasks. 

~\citet{lauret2006node} use a Fourier decomposition of the variance of the model predictions and rank hidden units according to how much that unit accounts for the variance and eliminates based on this variance-based spectral criterion. For a range of variation $[a_h, b_h]$ of parameter $w_h$ of layer $h$ and $N$ number of training iterations, each weight is varied as $w_h^{(n)} = (b_h + a_h / 2) + (b_h - a_h/2)\sin(\omega_h s^(n))$ where $s^{(n)} = 2\pi n/N$ and $\omega_h$ is the frequency of $w_h$ and $n$ is the training iteration. The $s_h$ is then obtained by computing the Fourier amplitudes of the fundamental frequency $\omega_h$, the first harmonic up to the third harmonic.

\subsubsection{Pruning using Second Order Derivatives}

\paragraph{Optimal Brain Damage}
As mentioned, deleting single weights is computationally inefficient and slow.~\citet{lecun1990optimal} instead estimate weight importance by making a local approximation of the loss with a Taylor series and use the $2^{nd}$ derivative of the loss with respect to the weight as a criterion to perform a type of weight sharing constraint. The objective is expressed as \autoref{eq:obd}

\begin{equation}\label{eq:obd}
\delta \mathcal{L} = \sum_i \cancelto{0}{g_i \delta \breve{w}_i} + \frac{1}{2} \sum_i h_{ii} \delta \breve{w}^2_i + \frac{1}{2} \sum_{i \neq j} h_{ij} \delta \breve{w}_i \delta w_j + \cancelto{0}{O(||\breve{W}||^2)}
\end{equation}

where $\breve{w}$ are perturbed weights of $w$, the $\delta \breve{w}_i$'s are the components of $\delta \breve{W}$, $g_i$ are the components of the gradient $\partial \mathcal{L} / \partial \breve{w}_i $ and $h_{ij}$ are the elements of the Hessian $\mat{H}$ where $\mat{H}_{ij} := \partial^2 \mathcal{L}/\partial \breve{w}_i \partial \breve{w}_j$. Since most well-trained networks will have $\mathcal{L} \approx 0$, the $1^{st}$ term is $\approx 0$. Assuming the perturbations on $W$ are small then the last term will also be small and hence ~\citet{lecun1990optimal} assume the off-diagonal values of $\mat{H}$ are 0 and hence $1/2\sum_{i \neq j} h_{ij} \delta \breve{w}_i \delta w_j := 0$. Therefore, $\delta \mathcal{L}$ is expressed as,

\begin{equation}
\delta \mathcal{L} \approx \frac{1}{2}\sum_{i} h_{ii} \delta \breve{w}_i^2    
\end{equation}

The $2^{nd}$ derivatives $\mat{h}_{kk}$ are calculated by modifying the backpropogation rule. Since $\vec{z}_i = f(\vec{a}_i)$ and $\vec{a}_i = \sum_{j} \mat{w}_{ij} \vec{z}_j$, then by substitution $\frac{\partial^2 \mathcal{L}}{\partial \mat{w}^2_{ij}} = \frac{\partial^2 \mathcal{L}}{\partial \mat{a}^2_{ij}}z_j$ and they further express the $2^{nd}$ derivative of the activation output as,

\begin{equation}
    \frac{\partial^2 \mathcal{L}}{\partial a^2_{i}} = f'(\vec{a}_i)^2 - \sum_l \mat{w}_{li}^2  \frac{\partial^2 \mathcal{L}}{\partial \vec{a}^2_{l}} - f''(\vec{a}_i)^2 \frac{\partial^2 \mathcal{L}}{\partial \vec{z}_i^2}
\end{equation}

The derivative of the mean squared error with respect to the to the last linear layer output is then 

\begin{equation}
    \frac{\partial^2 \mathcal{L}}{\partial \vec{a}^2_{i}} = 2 f'(\vec{a}_i)^2 - 2(\vec{y}_i - \vec{z}_i)f''(\vec{a}_i) 
\end{equation}

The importance of weight $w_i$ is then $s_k \approx h_{kk} \vec{w}^2_k /2 \ $ and the portion of weights with lowest $s_k$ are iteratively pruned during retraining.

\paragraph{Optimal Brain Surgeon}

~\citet{hassibi1994optimal} improve over OBD by preserving the off diagonal values of the Hessian, showing empirically that these terms are actually important for pruning and assuming a diagonal Hessian hurts pruning accuracy.

To make this Hessian computation feasible, they exploit the recursive relation for calculating the inverse hessian $\mat{H}^{-1}$ from training data and the structural information of the network. Moreover, using $\mat{H}^{-1}$ has advantages over OBD in that it does require further re-training post-pruning. 

They denote a weight to be eliminated as $w_q = 0,\ \delta w_q + w_q =0$ with the objective to minimize the following objective:

\begin{equation}\label{eq:obs_2}
\min_q \Big\{ \min_{\delta \vec{w}} \{ \frac{1}{2} \delta \vec{w}^T \cdot \mat{H} \cdot \delta \vec{w} \} \quad s.t \quad  \vec{e}^T_q \cdot \vec{w} + \vec{w}_q = 0 \Big\}  
\end{equation}

where $\vec{e}_q$ is the unit vector in parameter space corresponding to parameter $w_q$. To solve \autoref{eq:obs_3} they form a Lagrangian from \autoref{eq:obs_2}:

\begin{equation}\label{eq:obs_3}
\mathcal{L} = \frac{1}{2} \delta \vec{w}^T \cdot \mat{H} \cdot \delta \vec{w} + \lambda (\vec{e}^T_q \cdot \delta_{\vec{w}} + \vec{w}_q) 
\end{equation}

where $\lambda$ is a Lagrange undetermined multiplier. The functional derivatives are taken and the constraints of \autoref{eq:obs_2} are applied. Finally, matrix inversion is used to find the optimal weight change and resulting change in error is expressed as, 

\begin{equation}\label{eq:obs_4}
\delta \vec{w} =  \frac{w_q}{[\mat{H}^{-1}_{qq}]} \mat{H}^{-1} \vec{e}_q \quad \text{and} \quad  \mathcal{L}_q = \frac{1}{2} \frac{w_q^2}{[\mat{H}^{-1}_{qq}]} 
\end{equation}

Defining the first derivative as $\mat{X}_k := \frac{f(\vec{x}; \mat{W})}{\partial \mat{W}}$ the Hessian is expressed as,

\begin{equation}
    \mat{H} = \frac{1}{P} \sum_{k=1}^{P}\sum_{j=1}^{n} \mat{X}_{k,j} \cdot \mat{X}_{k,j}^T
\end{equation}

for an $n$-dimensional output and $P$ samples. This can be viewed as the sample covariance of the gradient and $\mat{H}$ can be recursively computed as,

\begin{equation}
    \mat{H}_{m+1}^{-1} = \mat{H}_{m}^{-1} + \frac{1}{P} \mat{X}_{m+1}^T \cdot \mat{X}_{m+1}^T 
\end{equation}

where $\mat{H}_0 = \alpha \mat{I}$ and $ \mat{H}_P = \mat{H}$. Here $10^{-8 }\leq \alpha \geq 10^{-4}$ is necessary to make $\mat{H}^{-1}$ less sensitive to the initial conditions. For OBS, $\mat{H}^{-1}$ is required and to obtain it they use a matrix inversion formula~\citep{kailath1980linear} which leads to the following update:

\begin{equation}
    \mat{H}^{-1}_{m+1} = \mat{H}^{-1}_{m} -\frac{\mat{H}^{-1}_{m} \cdot \mat{X}_{m+1} \cdot \ \mat{X}_{m+1}^T \cdot \mat{H}^{-1}_{m}}{ P + \mat{X}_{m+1}^{-1} \cdot \mat{H}^{-1}_{m} \cdot \mat{X}_{m+1}} \quad \text{where} \quad \mat{H}_0 = \alpha \mat{I}, \quad \mat{H}_P = \mat{H}
\end{equation}

This recursion step is then used as apart of \autoref{eq:obs_4}, can be computed in one pass of the training data $1 \leq m \leq P$ and computational complexity of $\mat{H}$ remains the same as $\mat{H}^{-1}$ as $\mathcal{O}(P n^2)$.
~\citet{hassibi1994optimal} have also extended their work on approximating the inverse hessian~\citep{hassibi1993second} to show that this approximation works for any twice differentiable objective (not only constrained to sum of squared errors) using the Fisher's score. 

Other methods to Hessian approximation include dividing the network into subsets to use block diagonal approximations and eigen decomposition of  $\mat{H}^{-1}$~\citep{hassibi1994optimal} and principal components of $\mat{H}^{-1}$~\citep{levin1994fast} (unlike aforementioned approximations,~\citet{levin1994fast} do not require the network to be trained to a local minimum). However the main drawback is that the Hessian is relatively expensive to compute for these methods, including OBD. For $n$ weights, the Hessian requires $\cO(n^2/2)$ elements to store and performs $\cO(P n^2)$ calculations per pruning step, where $P$ is total number of pruning steps. 

\subsubsection{Pruning using First Order Derivatives}\label{sec:pufod}
As $2^{nd}$ order derivatives are expensive to compute and the aforementioned approximations may be insufficient in representing the full Hessian, other work has focused on using $1^{st}$ order information as an alternative approximation to inform the pruning criterion. 

~\citet{molchanov2016pruning} use a Taylor expansion (TE) as a criterion to prune by choosing a subset of weights $W_s$ which have a minimal change on the cost function. They also add a regularization term that explicitly regularize the computational complexity of the network. \autoref{eq:ts_cost} shows how the absolute cost difference between the original network cost with weights $w$ and the pruned network with $w'$ weights is minimized such that the number of parameters are decreased where $||\cdot||_0$ denotes the $0$-norm bounds the number of non-zero parameters $W_s$. 

\begin{equation}\label{eq:ts_cost}
	\min_{\mat{W}'}  |\mathcal{C}(D|\mat{W}') - \mathcal{C}(D|\mat{W})| \quad \text{s.t.} \quad |\mat{W}'|_{0} \leq \mat{W}_s 
\end{equation}

Unlike OBD, they keep the absolute change $|y|$ resulting from pruning, as the variance $\sigma^2_y$ is non-zero and correlated with stability of the $\partial C / \partial h$ throughout training, where $\vec{h}$ is the activation of the hidden layer. Under the assumption that samples are independent and identically distributed, $\mathbb{E}(|y|) =\sigma\sqrt{2}/\sqrt{\pi}$ where $\sigma$ is the standard deviation of $y$, known as the expected value of the half-normal distribution. So, while $y$ tends to zero, the expectation of $|y|$ is proportional to the variance of $y$, a value which is empirically more informative as a pruning criterion.

They rank the order of filters pruned using the TE criterion and compare to an oracle rank (i.e the best ranking for removing pruned filters) and find that it has higher spearman correlation to the oracle when compared against other ranking schemes. This can also be used to choose which filters should be transferred to a target task model. They compute the importance of neurons or filters $z$ by estimating the mutual information with target variable MI$(z;y)$ using information gain $IG(y|z) = \mathcal{H}(z)  + \mathcal{H}(y) - \mathcal{H}(z, y)$ where $\mathcal{H}(z)$ is the entropy of the variable $z$, which is quantized to make this estimation tractable.

\paragraph{Fisher Pruning}~\citet{theis2018faster} extend the work of~\citet{molchanov2016pruning} by motivating the pruning scheme and providing computational cost estimates for pruning as adjacent layers are successively being pruned. Unlike OBD and OBS, they use, fisher pruning as it is more efficient since the gradient information is already computed during the backward pass. Hence, this pruning technique uses $1^{st}$ order information given by the $2^{nd}$ TE term that approximates the loss with respect to $w$. The fisher information is then computed during backpropogation and uses as the pruning criterion. 

The gradient can be formulated as \autoref{eq:fisher_1}, where $\mathcal{L}(w) = \mathbb{E}_{P}[-\log Q_{\vec{w}}(y | \vec{x})] 
$, $d$ represents a change in parameters, $P$ is the underlying distribution, $Q_{w}(y | x)$ is the posterior from the model $H$ is the Hessian matrix.

\begin{align}\label{eq:fisher_1}
g = \nabla \mathcal{L}(w), \quad  \mat{H} = \nabla^{2}\mathcal{L}(w), \quad
\mathcal{L}(w + d) - \mathcal{L}(w) \approx \vec{g}^{T} d + \frac{1}{2} \vec{d}^T \mat{H} \vec{d}  \\ 
\mathcal{L}(\mat{W} - \mat{W}_k \vec{e}_i) - \mathcal{L}(\mat{W}) + \beta \cdot (C(W - \mat{W}_k \vec{e}_i) - C(\mat{W}))
\end{align}

\paragraph{Piggyback Pruning}

~\citet{mallya2018piggyback} propose a dyanmic masking (i.e pruning) strategy whereby a mask is learned to adapt a dense network to a sparse subnetwork for a specific target task. The backward pass for binary mask is expressed as,

\begin{equation}
\frac{\partial \mathcal{L}}{\partial m_{ji}} =\Big(\frac{\partial \mathcal{L}}{\partial \hat{y}_j}\Big)\cdot \Big(\frac{\partial y_j}{\partial m_{ji}}\Big) =\partial \hat{y}_j \cdot w_{ji} \cdot x_i, 
\end{equation}

where $m_{ij}$ is an entry in the mask $m$, $\mathcal{L}$ is the loss function and $\hat{y}_j$ is the prediction when the $j-th$ mask is applied to the weights $w$. 
The matrix $m$ can then be expressed as $\frac{\partial L}{\partial m} = (\delta \vec{y} \cdot \vec{x}^T) \mat{W}$. Note that although the threshold for the mask $m$ is non-differentiable, but they perform a backward pass anyway. The justification is that the gradients of $m$ act as a noisy estimate of the gradients of the real-valued mask weights $m_r$. For every new task, $m$ is tuned with a new final linear layer.

\subsection{Structured Pruning}\label{sec:struct_prune}
Since standard pruning leads to non-structured connectivity, structured pruning can be used to reduce speed and memory since hardware is more amenable to dealing with dense matrix multiplications, with little to no non-zero entries in matrices and tensors. CNNs in particular are suitable for this type of pruning since they are made up of sparse connections. Hence, below we describe some work that use group-wise regularizers, structured variational, Adversarial  Bayesian methods to achieve structured pruning in CNNs.

\subsubsection{Structured Pruning via Weight Regularization}

\paragraph{Group Sparsity Regularization}


Group sparse regularizers enforce a subset of weight groupings, such as filters in CNNs, to be close to zero when trained using stochastic gradient descent. Consider a convolutional kernel represented as a tensor $K(i, j, s, :)$, the group-wise $\ell_2,1$-norm is given as 

\begin{equation}
    \omega_{2,1}(K) = \lambda \sum_{i,j,s}||\Gamma_{ijs}|| = \lambda \sum_{ijs} \sqrt{\sum_{t=1}^{T} K(i, j, s, t)^2}
\end{equation}

where $\Gamma_{ijs}$ is the group of kernel tensor entries $K(i, j, s, :)$ where $(i,j)$ are the pixel of $i$-th row and $j$-th column of the image for the $s$-th input feature map. This regularization term forces some $\Gamma_{ijs}$ groups to be close to zero, which can be removed during retraining depending on the amount of compression that the practitioner predefines.

\begin{wrapfigure}{R}{8cm}
    \centering
    \includegraphics[width=0.5\textwidth]{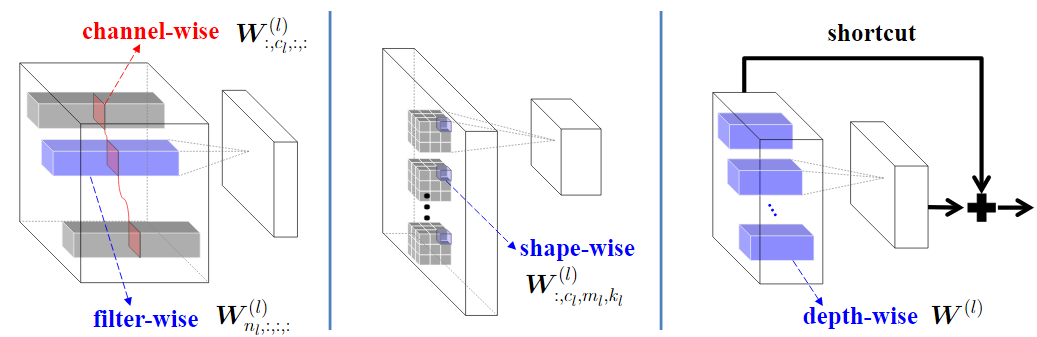}
    \caption{original source:~\citet{wen2016learning}: Structured Sparsity Learning}
    \label{fig:ssl}
\end{wrapfigure}

\paragraph{Structured Sparsity Learning}
~\citet{wen2016learning} show that their proposed structural regularization can reduce a ResNet architecture with 20 layers to 18 with 1.35 percentage point accuracy increase on CIFAR-10, which is even higher than the larger 32 layer ResNet architecture.  They use a group lasso regularization to remove whole filters, across channels, shape and depth as shown in \autoref{fig:ssl}.

\autoref{eq:ssl} shows the loss to be optimized to remove unimportant filters and channels, where $\mat{W}(l)_{n_l,c_l,:,:}$ is the $c$-th channel of the $l$-th filter for a collection of all weights $\mat{W}$ and $||\cdot||$ is the group Lasso regularization term where $||\vec{w}^{(g)} ||_g = \sqrt{\sum_{i=1}^{|\vec{w}^{(g)}|} \big(\vec{w}^{(g)}\big)^2}$ and $|\vec{w}^{(g)}|$ is the number of weights in $\vec{w}^{(g)}$.

Since zeroing out the $l$-th filter leads to the feature map output being redundant, it results in the $l+1$ channel being zeroed as well. Hence, structured sparsity learning is carried out for both filters and channels simultaneously. 

\begin{equation}\label{eq:ssl}
\mathcal{L}(\mat{W}) = \mathcal{L}_D(\mat{W}) + \lambda_n \cdot \sum_{l=1}^{L}\Big( \sum_{n_l = 1}^N ||\mat{W}^{(l)}_{m_l,:,:,:}||_g \Big) + \lambda_c \cdot \sum_{l=1}^{L} \Big(\sum_{c_l = 1}^{C_l} ||\mat{W}^{(l)}_{c_l,:,:,:}||_g \Big)
\end{equation}

\subsubsection{Structured Pruning via Loss Sensitivity}

\paragraph{Structured Brain Damage}
The aforementioned OBD has also been extended to remove groups of weights using group-wise sparse regularizers (GWSR)~\citet{lebedev2016fast}. In the case of filters in CNNs, this results in smaller reshaped matrices, leading to smaller and faster CNNs. The GWSR is added as a regularization term during retraining a pretrained CNN and after a set number of epochs, the groups with smallest $\ell_2$ norm are deleted and the number of groups are predefined as $\tau \in [0, 1]$ (a percentage of the size of the network). However, they find that when choosing a value for $\tau$, it is difficult to set the regularization influence term $\lambda$ and can be time consuming manually tuning it. Moreover when $\tau$ is small, the regularization strength of $\lambda$ is found to be too heavy, leading to many weight groups being biased towards 0 but not being very close to it. This results in poor performance as it becomes more unclear what groups should be removed. However, the drop in accuracy due to this can be remedied by further retraining after performing OBD. Hence, retraining occurs on the sparse network without using the GWSR.


\subsubsection{Sparse Bayesian Priors}

\paragraph{Sparse Variational Dropout}
Seminal work, such as the aforementioned Skeletonization~\citep{mozer1989skeletonization} technique has essentially tried to learn weight saliency. Variational dropout (VD), or more specifically Sparse Variational Dropout~\citep[(SpVD)][]{molchanov2017variational}, learn individual dropout rates for each parameter in the network using varitaional inference (VI). In Sparse VI, sparse regularization is used to force activations with high dropout rates (unlike the original VD~\citep{kingma2015variational} where dropout rates are bound at 0.5) to go to 1 leading to their removal. Much like other sparse Bayes learning algorithms, VD exhibits the Automatic relevance  determination (ARD) effect\footnote{Automatic relevance  determination provides a data-dependent prior distribution to prune away redundant features in the overparameterized regime i.e more features than samples}.~\citet{molchanov2017variational} propose a new approximation to the KL-divergence term in the VD objective and also introduce a way to reduce variance in the gradient estimator which leads to faster convergence. VI is performed by minimizing the bound between the variational Gaussian prior $q_{\phi}(w)$ and prior over the weight $p(w)$ as,


\begin{equation}
    \mathcal{L}(\phi) = \max_{\phi} \mathcal{L}_{D} - D_{\text{KL}}\Big(q_{\phi}(w) || p(w)\Big) \quad \text{where} \quad \mathcal{L}_{D}(\phi) = \sum_{n=1}^{N}\mathbb{E}_{q_{\phi}(w)}\Big[\log p(y_n | \vec{x}_n, \vec{w}_n)\Big]
\end{equation}

They use the reparameterization trick to reduce variance in the gradient estimator when $\alpha > 0.5$ by replacing multiplicative noise $1 + \sqrt{\alpha_{ij}} \cdot \epsilon_{ij}$ with additive noise $\sigma_{ij} \cdot \epsilon_{ij}$, where $\epsilon_{ij} \sim \mathcal{N}(0;1)$ and $\sigma^2_{ij}=\alpha_{ij}\cdot \theta^2_{ij}$ is tuned by optimizing the variational lower bound w.r.t $\theta$ and  $\sigma$. This difference with the original VD allow weights with high dropout rates to be removed.

Since the prior and approximate posterior are fully factorized, the full KL-divergence term in the lower bound is decomposed into a sum:

\begin{equation}
D_{\text{KL}}(q(\mat{W}|\theta, \alpha)|| p(\mat{W})) = \sum_{ij} D_{\text{KL}}(q(w_{ij} | \theta_{ij}, \alpha_{ij}) || p(w_{ij}))
\end{equation} 

Since the uniform log-prior is an improper prior, the KL divergence is only computed up to an additional constant~\citep{kingma2015variational}. 

\begin{equation}
- D_{\text{KL}}(q(w_{ij}|\theta_{ij}, \alpha_{ij}) || p(w_{ij})) =\frac{1}{2}\log \alpha_{ij}  - E \sim N(1, \alpha_{ij})\log |\cdot|+ C
\end{equation}

In the VD model this term is intractable, as the expectation $E \sim N(1,\alpha_{ij}) \log |\cdot|_in$ cannot be computed  analytically~\citep{kingma2015variational}. Hence, they approximate the negative KL. The negative KL increases as $\alpha_{ij}$ increases which means the regularization term prefers large values of $\alpha_{ij}$ and so the correspond weight $w_{ij}$ is dropped from the model. Since using SVD at the start of training tends to drop too many weights early since the weights are randomly initialized, SVD is used after an initial pretraining stage and hence this is why we consider it a pruning technique.

\paragraph{Bayesian Structured Pruning}
Structured pruning has also been achieved from a Bayesian view~\citep{louizos2017bayesian} of learning dropout rates. Sparsity inducing hierarchical priors are placed over the units of a DNN and those units with high dropout rates are pruned. Pruning by unit is more efficient from a hardware perspective than pruning weights as the latter requires priors for each individual weight, being far more computationally expensive and has the benefit of being more efficient from a hardware perspective as whole groups of weights are removed.  

If we consider a DNN as $p(D|w) =\prod_{i=1}^{N}p(y_i|x_i,w)$ where $x_i$ is a given input sample with a corresponding target $y_i$, $w$ are the weights of the network, governed by a prior distribution $p(w)$. Since computing the posterior $p(w|D) = p(D|w)p(w)/p(D)$ explicitly is intractactble, $p(w)$ is approximated with a simpler distribution, such as a Gaussian $q(w)$, parameterized by variational parameters $\phi$. The variational parameters are then optimized as,

\begin{gather}\label{eq:elbo}
\cL_E =  \mathbb{E}_{q_{\phi}(w)}[\log p(D|w)], \quad
\cL_C = \mathbb{E}_{q_{\phi}(w)}[\log p(w)] +\cH(q_{\phi}(w)) \\
L(\phi) = \cL_E + \cL_C
\end{gather}

where $\cH(\cdot)$ denotes the entropy and $\cL(\phi)$ is known as the evidence-lower-bound (ELBO). They note that $\cL_E$ is intractable for noisy weights and in practice Monte Carlo integration is used. When the simpler $q_{\phi}(w)$ is continuous the reparameterization trick is used to backpropograte through the deterministic part $\phi$ and Gaussian noise $\epsilon \sim N(0, \sigma^2 I)$. By substituting this into \autoref{eq:elbo} and using the local reparameterization trick~\citep{kingma2015variational} they can express $\cL(\phi)$ as

\begin{equation}
\mathcal{L}(\phi) = \mathbb{E}_p(\epsilon)[\log p(D|f(\phi, \epsilon))] + \mathbb{E}_{q_{\epsilon(w)}}[\log p(w)]  + \mathcal{H}(q_{\phi(w)}), \quad \text{s.t} \quad w=f(\phi, \epsilon) 
\end{equation}

with unbiased stochastic gradient estimates of the ELBO w.r.t the variational parameters $\phi$. They use mixture of a log-uniform prior and a half-Cauchy prior for $p(w)$ which equates to a horseshoe distribution~\citep{carvalho2010horseshoe}. By minimizing the negative KL divergence between the normal-Jeffreys scale prior $p(z)$ and the Gaussian variational posterior $q_{\phi}(z)$ they can learn the dropout rate $\alpha_i= \sigma^2 z_i/\mu_2 z_i$ as

\begin{equation}
-D_{\text{KL}}(\phi(z)||p(z)) \approx A \sum_i(k_1 \sigma(k_2 + k_3 \log \alpha_i) - 0.5 m(- \log \alpha_i) - k_1)
\end{equation}

where $\sigma(\cdot)$ is the sigmoid function, $m(\cdot)$ is the softplus function and $k_1 = 0.64$, $k_2= 1.87$ and $k_3= 1.49$. A unit $i$ is pruned if its variational dropout rate does not exceed threshold $t$, as $\log \alpha_i= (\log \sigma^2 zi - \log \mu_2 z_i) \geq t$.

It should be mentioned that this prior parametrization readily allows for a more flexible marginal posterior over the weights as we now have a compound distribution,
 
\begin{equation}
 q_{\phi}(W) = \int q_{\phi}(W|z) q_{\phi}(z) dz
\end{equation}

 
\paragraph{Pruning via Variational Information Bottleneck}
~\citet{dai2018compressing} minimize the variational lower bound (VLB) to reduce the redundancy between adjacent layers by penalizing their mutual information to ensure each layer contains useful and distinct information. A subset of neurons are kept while the remaining neurons are forced toward 0 using sparse regularization that occurs as apart of their variational information bottleneck (VIB) framework. They show that the sparsity inducing regularization has advantages over previous sparsity regularization approaches for network pruning. 

\autoref{eq:vib} shows the objective for compressing neurons (or filters in CNNs) where $\gamma_i$ controls the amount of compression for the $i$-th layer and $L$ is a weight on the data term that is used to ensure that for deeper networks the sum of KL factors does not result in the log likelihood term outweighed when finding the globally optimal solution. 

\begin{equation}\label{eq:vib}
    \mathcal{L}= \sum_{i=1}^L \gamma_i \sum_{j=1}^{r_i} \log \Big( 1 +\frac{\mu^2_{i,j}}{ \sigma^2_{i,j}} \Big) - L \mathbb{E}_{\{\vec{x},y\}\sim D,\ \vec{h} \sim p(\vec{h}|\vec{x})}\Big[\log q(\vec{y}|\vec{h}_L) \Big]
\end{equation}

$L$ naturally arises from the VIB formulation unlike probabilistic networks models. The $\log(1 + u)$ in the KL term is concave and non-decreasing for range $[0, \infty]$ and therefore favors solutions that are sparse with a subset of parameters exactly zero instead of many shrunken ratios $\alpha_{i,j}: \mu^2_{i,j} \sigma^{-2}_{i,j}, \ \forall i,j$.

Each layer is sampled $\epsilon_i \sim \mathcal{N}(0,I)$ in the forward pass and $\vec{h}_i$ is computed. Then the gradients are updated after backpropogation for $\{\mu_i, \sigma_i \mat{W}_i \}^L_{i=1}$ and output weights $\mat{W}_y$.

\begin{wrapfigure}{R}{8cm}
    \centering
    \includegraphics[scale=0.5]{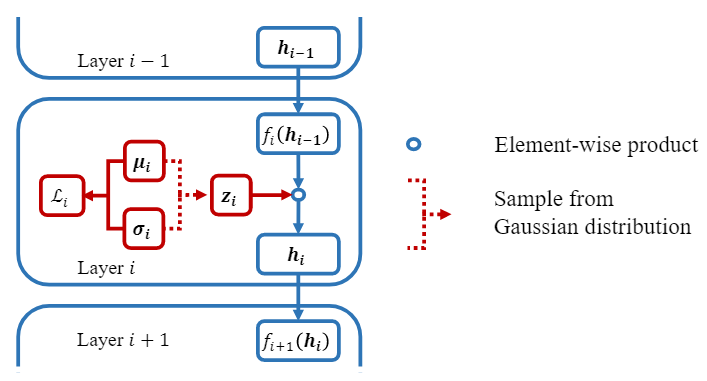}
    \caption{original source: ~\citet{dai2018compressing} Variational Information Structure}
    \label{fig:vib}
\end{wrapfigure}

\autoref{fig:vib} shows the conditional distribution $p(\vec{h}_i|\vec{h}_{i-1})$ and $\vec{h}_{i}$ sampled by multiplying $f_i(\vec{h}_{i-1})$ with a random variable $\vec{z}_i := \mu_i+ \epsilon_i \circ \sigma_i$.

They show that when using VIB network, the mutual information increases between $\vec{x}$ and $\vec{h}_1$ as it initially begins to learn and later in training the mutual information begins to drop as the model enters the compression phase.
In constrast, the mututal information for the original stayed consistently high tending towards 1.

\paragraph{Generative Adversarial-based Structured Pruning}

~\citet{lin2019towards} extend beyond pruning well-defined structures, such as filters, to more general structures which may not be predefined in the network architecture. They do so applying a soft mask to the output of each structure in a network to be pruned and minimize the mean squared error with a baseline network and also a minimax objective between the outputs of the baseline and pruned network where a discriminator network tries to distinguish between both outputs. During retraining, soft mask weights are learned over each structure (i.e filters, channels, ) with a sparse regularization term (namely, a fast iterative shrinkage-thresholding algorithm) to force a subset of the weights of each structure to go to 0. Those structures which have corresponding soft mask weight lower than a predefined threshold are then removed throughout the adversarial learning. This soft masking scheme is motivated by previous work~\citep{lin2018accelerating} that instead used hard thresholding using binary masks, which results in harder optimization due to non-smootheness.
Although they claim that this sparse masking can be performed with label-free data and transfer to other domains with no supervision, the method is largely dependent on the baseline (i.e teacher network) which implicitly provides labels as it is trained with supervision, and thus it pruned network transferability is largely dependent on this.

\subsection{Search-based Pruning}
Search-based techniques can be used to search the combinatorial subset of weights to preserve in DNNs. Here we include pruning techniques that don't rely on gradient-based learning but also evolutionary algorithms and SMC methods. 

\subsubsection{Evolutionary-Based Pruning}
\paragraph{Pruning using Genetic Algorithms}
The basic procedure for Genetic Algorithms (GAs) in the context of DNNs is as follows; (1) generate populations of parameters (or \textit{chromosones} which are binary strings), (2) keep the top-k parameters that perform the best (referred to as tournament selection) according to a predefined \textit{fitness} function (e.g classification accuracy), (3) randomly mix (i.e cross over) between the parameters of different sets within the top-k and perturb a portion of the resulting parameters (i.e mutation) and (4) repeat this procedure until convergence. This procedure can be used to find a subset of the DNN network that performs well.  

~\citet{whitley1990genetic} use a GA to find the optimal set of weights which involves connecting and reconnecting weights to find mutations that lead to the highest fitness (i.e lowest loss). They define the number of backpropogation steps as $ND + B$ where $B$ is the baseline number of steps, $N$ is the number of weights pruned and $D$ is the increase in number of backpropgation steps. Hence, if the network is heavily pruned the network is allocated more retraining steps. Unlike standard pruning techniques, weights can be reintroduced if they are apart of combination that leads to a relatively good fitness score. They assign higher reward to network which more heavily pruned, otherwise referred to as \textit{selective pressure} in the context of genetic algorithms. 

Since the cross-over operation is not specific to the task by default, interference can occur among related parameters in the population which makes it difficult to find a near optimal solution, unless the population is very large (i.e exponential with respect to the number of features).~\citet{cantu2003pruning} identify the relationship between variables by computing the joint distribution of individuals left after tournament selection and use this sub-population to generate new members of the population for the next iteration. This is achieved using 3 distribution estimation algorithms (DEA). They find that DEAs can improve GA-based pruning and that in pruned networks using GA-based pruning results in faster inference with little to no difference in performance compared to the original network.

Recently,~\citet{hu2018novel} have pruned channels from a pretrained CNN using GAs and performed knowledge distillation on the pruned network. A kernel is converted to a binary string $K$ with a length equal to the number of channels for that kernel. Then each channel is encoded as 0 or 1 where channels with a 0 are pruned and the n-th kernel $K_n$ is represented a a binary series after sampling each bit from a Bernoulli distribution for all $C$ channels. Each member (i.e channels) in the population is evaluated and top-k are kept for the next generation (i.e iteration) based on the fitness score where k corresponds to the total amount of pruning. The Roulette Wheel algorithm is used as the selection strategy~\citep{goldberg1991comparative} whereby the $n$-th member of the $m$-th generation $I_{m,n}$ has a probability of selection proportional to its fitness relative to all other members. This can simply be implemented by inputting all fitness scores for all members into a softmax. To avoid members with high fitness scores losing information post mutation and cross-over, they also copy the highest fitness scoring members to the next generation along with their mutated versions. 

The main contribution is a 2-stage fitness scoring process. First, a local TS approximation of a layer-wise error function using the aforementioned OBS objective~\citep{dong2017learning} (recall that OBS mainly revolves around efficient Hessian approximation) is used sequentially from the first layer to the last, followed by a few epochs of retraining to restore the accuracy of the pruned network. Second, the pruned network is distilled usin a cross-entropy loss and regularization term that forces the features maps of the pruned network to be similar to the distilled model, using an attention map to ensure both corresponding layer feature maps are of the same and fixed size. They achieve SoTA on ImageNet and CIFAR-10 for VGG-16 and ResNet CNN architectures using this approach. 



\paragraph{Pruning via Simulated Annlealing}
~\citet{noy2019asap} propose to reduce search time for searching neural architectures by relaxing the discrete search to continuous that allows for a differentiable simulated annealing that is optimized using gradient descent (following from the DARTS~\citep{liu2018darts} approach). This leads to much faster solutions compared to using black-box search since optimizing over the continuous search space is an easier combinatorial optimization problem that in turn leads to faster convergence. This pruning technique is not strictly consider compression in its standard definition, as it prunes during the initial training period as opposed to pruning after pretraining. This falls under the category of neural architecture search (NAS) and here they use  an annealing schedule that controls the amount of pruning during NAS to incrementally make it easier to search for sub-modules that are found to have good performance in the search process. Their (0, $\delta$)-PAC theorem guarantees under few assumptions (see paper for further details on these assumptions) that this anneal and prune approach prunes less important weights with high probability.

\subsubsection{Sequential Monte Carlo \& Reinforcement Learning Based Pruning}

\paragraph{Particle Filter Based Pruning}
~\citet{anwar2017structured} identifies important weights and paths using particle filters where the importance weight of each particle is assigned based on the misclassification rate with corresponding connectivity pattern. 
Particle filtering (PF) applies sequential Monte Carlo estimation with particle representing the probability density where the posterior is estimated with a random sample and parameters that are used for posterior estimation. PF propogates parameters with large magnitudes and deletes parameters with the smallest weight in re-sampling process, similar to MBP. They use PF to prune the network and retrain to compensate for the loss in performance due to PF pruning. When applied to CNNs, they reduce the size of kernel and feature map tensors while maintaining test accuracy.

\paragraph{Particle Swarm Optimized Pruning}
Particle Swarm Optimization (PSO) has also been combined with correlation merging algorithm (CMA) for pruning~\citep{tu2010neural}. \autoref{eq:pso} shows the PSO update formula where the velocity $\mat{V}_{id}$ for i-th position of particle $\mat{X}_id$ (i.e a parameter vector in a DNN) at the $d$-th iteration, 

\begin{equation}\label{eq:pso}
    \mat{V}_{id} := \mat{V}_{id} + c_1 u (\vec{P}_{id} - \vec{X}_{id}) + c_2 u (\vec{P}_{gd} - \vec{X}_{id}), \quad \text{where} \quad \vec{X}_{id} = \vec{X}_{id} + \mat{V}_{id}
\end{equation}

where $u \sim \text{Uniform}(0, 1)$ and $c_1, c_2$ are both learning rates, corresponding to the influence social and cognition components of the swarm respectively~\citep{kennedy1995particle}. Once the velocity vectors are updated for the DNN, the standard deviation is computed for the i-th activation as $s_i = \sum_{p=1}^{n} (\mat{V}_{ip} - \bar{\mat{V}}_i)^2$ where $\bar{v}_i$ is the mean value of $\mat{V}_i$ over training samples. 

Then compute Pearson correlation coefficient between the $i$-th an $j$-th unit in the hidden layer as $\mat{C}_{ij} = (\mat{V}_{ip} \mat{V}_{jp} - n\bar{\mat{V}}_i \bar{\mat{V}}_j)/\vec{S}_i \vec{S}_j$ and if $\mat{C}_{ij} > \tau_1$ where $\tau$ is a predefined threshold, then merge both units, delete the j-th unit and update the weights as,

\begin{equation}
    \mat{W}_{ki} = \mat{W}_{ki} + \alpha \mat{W}_{ki} \quad \text{and} \quad \mat{W}_{kb} = \mat{W}_{kb} + \beta \mat{W}_k
\end{equation}

where,

\begin{equation}
\alpha = \frac{\mat{V}_{ip} \mat{V}_{jp} - n\bar{\mat{V}}_i \bar{\mat{V}}_j}{\sum_{n=1}^p \mat{V}_{ip} \mat{V}_{jp} - \bar{\mat{V}}^{2}_i}, \quad \beta = \bar{\mat{V}}_j  - \alpha \bar{\mat{V}}_i
\end{equation}

and $\mat{W}_{ki}$ connects the last hidden layer to output unit $k$. If the standard deviation of unit $i$ is less than $\tau_2$ then it is combined with the output unit $k$. Finally, remove unit $j$ and update the bias of the output unit k as $\mat{W}_{kb}= \mat{W}_{kb} + \vec{\bar{V}}_i \mat{W}_{ki}$. This process is repeated until a maximally compressed network than maintains performance similar to the original network is found.

\paragraph{Automated Pruning}
AutoML~\citep{he2018amc} use RL to improve the efficiency of model compression performance by exploiting the fact that the sparsity of each layer is a strong signal for the overall performance. They search for a compressed architecture in a continuous space instead of searching over a discrete space. A continuous compression ratio control strategy is employed using an actor critic model (Deep Deterministic Policy Gradient~\citep{silver2014deterministic}) which is known to be relatively stable during training, compared to alternative RL models, due lower variance in the gradient estimator. The DDPG processes each consecutive layer, where for the $t$-th layer $L_t$, the network receives a layer embedding $t$ that encodes information of this layer and outputs a compression ratio $a_t$ and repeats this process from the first to last layer. The resulting pruned network is evaluated without fine-tuning, avoiding retraining to improve computational cost and time. During training, they fine-tune best explored model given by the policy search. The MBP ratio is constrained such that the compressed model produced by the agent is below a resource constrained threshold in resource constrained case. Moreover, the maximum amount of pruning for each layer is constrained to be less than 80\%, 
When the focus is to instead maintain accuracy, they define the reward function to incorporate accuracy and the available hardware resources.

By only requiring 1/4 number of the FLOPS they still manage to achieve a 2.7\% increase in accuracy for MobileNet-V1. This also corresponds to a 1.53 times speed up on a Titan Xp GPU and 1.95 times speed up on Google Pixel 1 Android phone.

\subsection{Pruning Before Training}
Thus far, we have discussed pruning pretrained networks. Recently, the \text{lottery ticket} hypothesis~\citep[LTH][]{frankle2018lottery} showed that there exists sparse subnetworks that when trained from scratch with the same initialized weights can reach the same accuracy as the full network. The process can be formalized as:

\begin{enumerate}
    \itemsep0em 
    \item  Randomly initialize a neural network $f(\vec{x}; \theta_0)$ (\text{where} $\theta_0 \sim D_\theta $).
    \item Train the network for $j$ iterations, arriving at parameters $\theta_j$
    \item Prune $p$ \% of the parameters in $\theta_j$, creating a mask $m$.
    \item Reset the remaining parameters to their values in $\theta_0$, creating the winning ticket $f(\vec{x}; m \otimes \theta_0 )$.
\end{enumerate}

~\citet{liu2018rethinking} have further shown that the network architecture itself is more important than the remaining weights after pruning pretrained networks, suggesting pruning is better perceived as an effective architecture search. This coincides with Weight Agnostic Neural Networks~\citep[WANN;][]{gaier2019weight} search which avoids weight training. Topologies of WANNs are searched over by first sampling single shared weight for a small subnetwork and evaluated over several randomly shared weight rollout. For each rollout the cumulative reward over a trial is computed and the population of networks are ranked according to the resulting performance and network complexity. This highest ranked networks are probabilistically selected and mixed at random to form a new population. The process repeats until the desired performance and time complexity is met. 

The two aforementioned findings (there exists smaller sparse subnetworks that perform well from scratch and the importance of architecture design) has revived interest in finding criteria for finding sparse and trainable subnetworks that lead to strong performance. 

However, the original LTH paper was demonstrated on relatively simpler CV tasks such as MNIST and when scaled up it required careful fine-tuning of the learning rate for the lottery ticket subnetwork to achieve the same performance as the full network. To scale up LTH to larger architectures
~\cite{frankle2019lottery} in a stable way without requiring any additional fine-tuning, they relax the restrictions of reverting to the lottery ticket being found at initialization but instead revert back to the $k$-th epoch. This $k$ typically corresponds to only few training epochs from initialization. Since the lottery ticket (i.e subnetwork) no longer corresponds to a randomly initialized subnetwork but instead a network trained from $k$ epochs, they refer to these subnetworks as \textit{matching tickets} instead. This relaxation on LTH allows tickets to be found on CIFAR-10 with ResNet-20 and ImageNet with ResNet-50, avoiding the need for using optimizer warmups to precompute learning rate statistics.

~\citet{zhou2019deconstructing} have further investigate the importance of the three main factors in pruning from scratch: (1) the pruning criteria used, (2) where the model is pruned from (e.g from initialization or $k$-th epoch) and (3) the type of mask used. They find that the measuring the distance between the weight value at intialization and its value after training is a suitable criterion for pruning and performs at least as well as preserving weights based on the largest magnitude. They also note that if the sign is the same after training, these weights can be preserved. Lastly, they find for (3) that using a binary mask and setting weights to 0 is plays an integral part in LTH. 
Given that these LTH based pruning masks outperform random masks at initialization, leads to the question whether we can search for architectures by pruning as a way of learning instead of traditional backpropogation training. In fact,~\cite{zhou2019deconstructing} have also propose to use REINFORCE~\citep{sutton2000policy} to optimize and search for optimal wirings at each layer. In the next subsection, we discuss recent work that aims to find optimal architectures using various criteria.

\subsubsection{Pruning to Search for Optimal Architectures}
Before LTH and the aforementioned line of work, Deep Rewiring~\citep[DeepR;][]{bellec2017deep} was proposed to adaptively prune and reappear periodically during training by drawing stochastic samples of network configurations from a posterior. The update rule for all active connections is given as,

\begin{equation}
    \mat{W}_k \gets \mat{W}_k - \eta \frac{\partial E}{\partial \mat{W}_k}- \eta \alpha  + \sqrt{2 \eta \Gamma} v_k
\end{equation}

for $k$-th connection. Here, $\eta$ is the learning rate, $\Gamma$ is a temperature term, $E$ is the error function and the noise $v_k \sim \mathcal{N}(0, I\sigma^{2})$ for each active weight $\mat{W}$. If the $\mat{W}_k < 0$ then the connection is frozen. When the set the number of dormant weights exceeds a threshold, they reactivate dormant weights with uniform probability. The main difference between this update rule and SGD lies in the noise term $\sqrt{2 \eta \Gamma} v_k$ whereby the $v_k$ noise and the amount of it controlled by $\Gamma$ performs a type of random walk in the parameter space. Although unique, this approach is computationally expensive and challenging to apply to large networks and datasets.

Sparse evolutionary training~\citep[SET;][]{mocanu2018scalable} simplifies prune–regrowth cycles by replacing the top-k lowest magnitude weights with newly randomly initialized weights and retrains and this process is repeated throughout each epoch of training.
~\citet{dai2019nest} carry out the same SET but using gradient magnitude as the criterion for pruning the weights. Dynamic  Sparse  Reparameterization~\citep[DSR;][]{mostafa2019parameter} implements  a  prune–redistribute–regrowth cycle where target sparsity levels are redistributed among layers, based on loss gradients (in contrast to SET, which uses fixed, manually configured, sparsity levels). SparseMomentum~\citep[SM;][]{dettmers2019sparse} follows the same cycle but instead using the mean momentum magnitude of each layer during the redistribute phase.  SM outperforms DSR on ImageNet for unstructured pruning by a small margin but has no performance difference on CIFAR experiments. Our approach also falls in the dynamic category but we use error compensation mechanisms instead of hand crafted redistribute–regrowth cycles

~\citet{ramanujan2020s}\footnote{This approach also is also relevant to \autoref{sec:pufod} as it relies on $1^{st}$ order derivatives for pruning.} propose an \texttt{edge-popup} algorithm to optimize towards a pruned subnetwork from a randomly initialized network that leads to optimal accuracy. The algorithm works by switching edges until the optimal configuration is found. Each weight is assigned a ``popup'' score $s_{uv}$ from neuron $u$ to $v$. The top-k \% percentage of weights with the highest popup score are preserved while the remaining weights are pruned. Since the top-k threshold is a step function which is non-differentiable, they propose to use a straight-through estimator to allow gradients to backpropogate and differentiate the loss with respect to $s_{uv}$ for each respective weight i.e the activation function $g$ is treated as the identity function in the backward pass. The scores $s_{uv}$ are then updated via SGD. Unlike,~\citet{theis2018faster} that use the absolute value of the gradient, they find that preserving the direction of momentum leads to better performance. 
During training, removed edges that are not within the top-k can switch to other positions of the same layer as the scores change. They show that this shuffling of weights to find optimal permutation leads to lower cross-entropy loss throughout training. Interestingly, this type of adaptive pruning training leads to competitive performance on ImageNet when compared to ResNet-34 and can be performed on pretrained networks.

\subsubsection{Few-Shot and Data-Free Pruning Before Training}
Pruning from scratch requires a criterion that when applied, leads to relatively strong out-of-sample performance compared to the full network. LTH established this was possible, but the method to do so requires an intensive number of pruning-retraining steps to find this subnetwork. Recent work, has focused trying to find such subnetworks without any training, of only a few mini-batch iterations. 
~\citet{lee2018snip} aim to find these subnetworks in a single shot i.e a single pass over the training data. This is referred to as Single-shot Network Pruning (SNIP) and as in previously mentioned work it too constructs the pruning mask by measuring connection sensitivities and identifying structurally important connections. 

~\citet{you2019drawing} identify to as `early-bird' tickets (i.e winning tickets early on in training) using a combination of early stopping, low-precision training and large learning rates. Unlike, LTH that use unstructured pruning, `early-bird' tickets are identified using structured pruning whereby whole channels are pruned based on their batch normalization scaling factor. Secondly, pruning is performed iteratively within a single training epoch, unlike LTH that performs pruning after numerous retraining steps. The idea of pruning early is motivated by ~\citet{saxe2019information} that describe training in two phase: (1) a label fitting phase where most of the connectivity patterns form and (2) a longer compression phase where the information across the networks is dispersed and lower layers compress the input into more generalizable representations. Therefore, we may only need phase (1) to identify important connectivity patterns and in turn find efficient sparse subnetworks.~\citet{you2019drawing} conclude that this hypothesis in fact the case when identifying channels to be pruned based on the hamming distance between consequtive pruning iterations. Intuitively, if the hamming distance is small and below a predefined threshold, channels are removed.

~\citet{tanaka2020pruning} have further investigated whether tickets can be identified without any training data. They note that the main reason for performance degradation with large amounts of pruning is due to \text{layer collapse}. Layer collapse refers when too much pruning leads to a cut-off of the gradient flow (in the extreme case, a whole layer is removed), leading to poor signal propogation and maximal compression while allowing the gradient to flow is referred to as \textit{critical compression}. 

\begin{wrapfigure}{R}{7cm}
    \centering
    \includegraphics[scale=0.37]{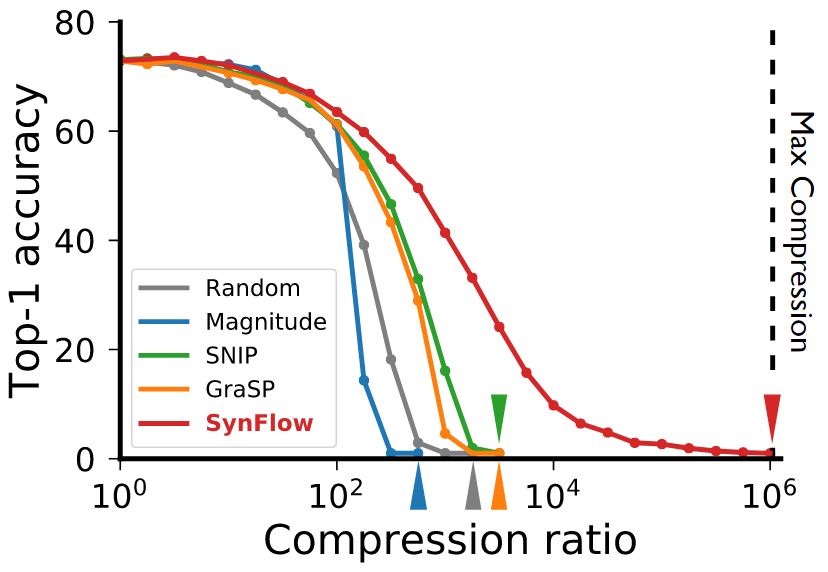}
    \caption{original source:~\citet{tanaka2020pruning} - Layer collapse in VGG-16 network for different pruning criteria on CIFAR-100}
    \label{fig:syn_flow}
\end{wrapfigure}

They show that retraining with MBP avoids layer-wise collapse because gradient-based optimiziation encourages compression with high signal propogation. From this insight, they propose a measure for measuring synaptic flow, expressed in \autoref{eq:syn_flow}. The parameters are first masked as $\theta_{\mu} \gets \mu \odot \theta_0$. Then the iterative synaptic flow pruning objective is evaluated as,

\begin{equation}\label{eq:syn_flow}
   \mathcal{L} = \mathbf{1}^{T}(\prod_{l=1}^{T} |\theta[l]_{\mu}|) \mathbf{1}
\end{equation}

where $\mathbf{1}$ is a vectors of ones. The score $\mathcal{S}$ is then computed as $\mathcal{S} = \frac{\partial \mathcal{R}}{\partial \theta_{\mu}} \odot \theta_{\mu}$ and the threshold $\tau$ is defined as $\tau = (1 - \rho - k/n)$ where $n$ is the number of pruning iterations and $\rho$ is the compression ratio. If $\mathcal{S}) > \tau$ then the mask $\mu$ is updated.

The effects of layer collapse for various random pruning, MBP, SNIP and synaptic flow (SynFlow) are shown in \autoref{fig:syn_flow}. We see that SynFlow achieves far higher compression ratio for the same test accuracy without requiring any data. 



\section{Low Rank Matrix \& Tensor Decompositions}\label{sec:lrtd}
DNNs can also be compressed by decomposing the weight tensors ($2^{nd}$ order tensor in the case of a matrix) into a lower rank approximation which can also removed redundancies in the parameters. Many works on applying TD to DNNs have been predicated on using SVD~\citep{xue2013restructuring,sainath2013low,xue2014singular,xue2014singular,novikov2015tensorizing}. Hence, before discussing different TD approaches, we provide an introduction to SVD.

A matrix $\mathbb{A} \in \mathbb{R}^{m \times n}$ of full rank $r$ can be decomposed as $\mat{A} = \mat{W}\mat{H}$ where $\mat{W}\in \mathbb{R}^{m \times r}$ and $\mat{H} \in \mathbb{R}^{r \times n}$. The change in space complexity as $\cO(\text{mn}) \to \cO(r(m + n))$ at the expense of some approximation error after optimizing the following objective,

\begin{equation}\label{eq:lrm}
\min_{\mat{W}, \mat{H}} \frac{1}{2}||\mat{A} - \mat{W} \mat{H}||^2_F    
\end{equation}

where for a low rank $k < r$, $\mat{W} \in \mathbb{R}^{m \times k}$ and $\mat{H} \in \mathbb{R}^{k \times n}$ and $||\cdot||_F$ is the Frobenius norm.

A common technique for achieving this low rank TD is Singular Value Decomposition (SVD). For orthogonal matrices $\mat{U} \in \mathbb{R}^{m \times r}$, $\mat{V} \in \mathbb{R}^{n\times r}$ and a diagonal matrix $\Sigma \in \mathbb{R}^{r \times r}$ of singular values, we can express $\mat{A}$ as

\begin{equation}\label{eq:svd_1}
\mat{A} =\mat{U} \Sigma \mat{V}^T
\end{equation}

where if $k < r$ then this is called truncated SVD. The nonzero elements of $\Sigma$ are the sorted in decreasing order and the top k $\Sigma_k \in \mathbb{R}^{k \times k}$ are used as $\mat{A} \approx \mat{U}_k \Sigma_k \mat{V}_k^T$.

Randomized SVD~\citep{halko2011finding} has also been introduced for faster approximation using ideas from random matrix theory. An approximation of the range $\mat{A}$ by finding $\mat{Q}$ with $r$ othornomal columns and $\mat{A} \approx \mat{Q} \mat{Q}^T \mat{A}$. Then the SVD is found by constructing a matrix $\mat{B} = \mat{Q}^T \mat{A}$ and SVD is instead computed on $\mat{B}$ as before using~\autoref{eq:svd_1}. Since $\mat{A} \approx \mat{Q} \mat{B}$, we can see $\mat{U} = \mat{Q} \mat{S}$ computes a LRD $\mat{A} \approx U \mat{S} \mat{V}^T$

Then as $\mat{A} \approx \mat{Q} \mat{Q}^T \mat{A} = \mat{Q} (\mat{S} \Sigma \mat{V}^T)$, we see that taking $\mat{U} = \mat{Q} \mat{S}$, we have computed a low rank approximation $\mat{A} \approx U \mat{S} \mat{V}^T$. Approximating $\mat{Q}$ is achieved by forming a Gaussian random matrix $\omega \in \mathbb{R}^{n \times l}$ and computing $\mat{Z} = \mat{A} \omega$, and using QR decomposition of $\mat{Z}, \mat{Q} \mat{R} = \mat{Z}$, then $\mat{Q} \in \mathbb{R}^{m \times l}$ has columns that are an orthonormal basis for the range of $\mat{Z}$.

Numerical precision is maintained by taking intermediate QR and LU decompositions during $o$ power iterations of $\mat{A}\mat{A}^T$ to reduce $Y$'s spectrum because if the singular values of $\mat{A}$ are $\Sigma$, then the singular values of $(\mat{A}\mat{A}^T)^{o}$ are $\Sigma^{2o + 1}$. With each power iteration the spectrum decays exponentially, therefore it only requires very few iterations. 

\subsection{Tensor Decomposition}
Generalizing $\mat{A}$ to higher order tensors, which we can refer to as an $\ell$-way array $\mathcal{A} \in \mathbb{R}^{n_a \times n_b \ldots \times n_{x}}$, the aim is to find the components $\mathcal{A} = \sum_{i}^{r} a \circ b \circ x  = [[ \mat{A}, \mat{B} \ldots, \mat{X} ]]$. 

Before discussing the TD we first define three important types of matrix products used in tensor computation: 

\begin{itemize}
    \item The Kronecker product between two arbitrarily-sized matrices $\mat{A} \in \mathbb{R}^{I \times J}$ and $\mat{B} \in \mathbb{R}^{K \times L}$,$\mat{A} \otimes \mat{B} \in \mathbb{R}^{(I )\times (JL)}$ , is a generalization of the outer product from vectors to 3 matrices $\mat{A} \otimes \mat{B} := [\vec{a}_1 \otimes \vec{b}_1, \vec{a}_2 \otimes \vec{b}_2, \ldots \vec{a}_J \otimes \vec{b}_{L-1}, \vec{a}_J \otimes \vec{b}_{L}]$.
    \item The Khatri-Rao product between two matrices $\mat{A} \in \mathbb{R}^{I \times K}$ and $\mat{B} \in \mathbb{R}^{J \times K}$, $\mat{A}\mat{B} \in \mathbb{R}^{(IJ) \times K}$, corresponds to the column-wise Kronecker product. $\mat{A}\mat{B} := [\vec{a}_1 \otimes \vec{b}_1 \vec{a}_2 \otimes b_2 \ldots \vec{a}_K \otimes \vec{b}_K]$.
    \item The Hadamard product is the elementwise product between 2 matrices $\mat{A, B} \in \mathbb{R}^{I \times J}$ and $\mat{A} * \mat{B} \in \mathbb{R}^{I \times J}$.
\end{itemize}

These products are used when performing Canonical Polyadic~\citep[(CP)][]{hitchcock1927expression}, Tucker decompositions~\citep{tucker1966some}, Tensor Train~\citep[TT][]{oseledets2011tensor} to find the factor matrices $\mathcal{X} := [[ \mat{A}, \mat{B} \ldots, \mat{C} ]]$. For the sake of simplicity we'll proceed with 3-way tensors. As before in \autoref{eq:lrm}, we can express the optimization objective as

\begin{equation}\label{eq:td_1}
 \min_{\mat{A}, \mat{B}, \mat{C}} \sum_{i,j, k} || \vec{x}_{ijk} - \sum_{l} \vec{a}_{il} \vec{b}_{jl} \vec{c}_{\text{kl}}  ||^2
\end{equation}

Since the components $\vec{a}, \vec{b},  \vec{c}$ are not orthogonal, we cannot compute SVD as was the case for matrices. The rank $r$ of $\mat{A}$ is also NP-hard and the solutions found for lower rank approximations may not be apart of the solution for higher ranks. Unlike, when you rotate the row or column vectors of a matrix and apply dimensionality reduction (e.g PCA) and still get the same solution, this is not the case for TD.  Unlike matrices where there can be many low rank matrices, a tensor is requires to have a low-rank matrix that is compatible for all tensor slices. This interconnection between different slices results in tensor being more restrictive and hence the for weaker uniqueness conditions. 

One way to perform TD using \autoref{eq:td_1} is to use alternating least squares (ALS) which involves minimizing  $\min_{\mat{A}}$ while fixing $\mat{B}, \mat{C}$ and repeating this for $\min_{\mat{B}}$ and $\min_{\mat{C}}$. ALS is suitable because it is a nonconvex optimization problem but with convex subproblems. CP can be generalized to different objectives apart from the squared loss, such as Rayleigh (when entries are non-negative), Boolean (entries are binary) and Poisson (when entries are counts) losses. Similar to randomized SVD described in the previous subsection, they have also been successful when scaling up TD. 

With this introduction, we now move onto how low rank TD has been applied to DNNs for reducing the size of large weight matrices. 

\subsection{Applications of Tensor Decomposition to Self-Attention and Recurrent Layers}
\subsubsection{Block-Term Tensor Decomposition (BTD)}
Block-Term Tensor Decomposition ~\citep[(BTD)][]{de2008decompositions} combines CP decomposition and Tucker decomposition. Consider an $n-th$ order tensor as $\mathcal{A} \in \mathbb{R}^{A_1 x \ldots x A_n}$ that can be decomposed into $N$ block terms and each block consist of $k$ elements between a core tensor $\mathcal{G}_n \in \mathbb{R}^{G_1 \times \ldots G_d}$ and $d$ and factor matrices $\mathcal{C}_{n}^{(k)} \in \mathbb{R}^{A_k \times G_k}$ along the $k$-th dimension where $n \in [1, N]$ and $k \in [1, d]$~\citep{de2008decompositions}. BTD can then be defined as, 

\begin{equation}
\mathcal{A} = N \sum_{n=1} \mathcal{G}_n \otimes \mathcal{C}^{(1)}_n  \otimes \mathcal{C}^{(2)}_n \ldots \mathcal{C}^{(d)}_n
\end{equation}

The $N$ here is the CP-rank, $G_1$, $G_2$, $G_3$ is the Tucker-rank and $d$ the core-order.

\paragraph{BTD RNNs}
~\citet{ye2018learning} also used BTD to learn small and dense RNNs by first tensorizing the RNN weights and inputs to a 3-way tensor as $\mathcal{X}$ and $\mathcal{W}$ respectively. BTD is then performed on the weights $\mathcal{W}$ and a tensorized backpropogation is computed when updating the weights. The core-order $d$ is important when deciding the total number of parameters and they recommend $d = [3, 5]$ which is a region that corresponds to orders of magnitude reduction in the number of parameters. When $d > 5$ the number of parameters begins to increase again since the number of parameters is defined as $p_{BTD} = N \sum_{k=1}^d Y_k Z_k R + R^d$ where $Y_k$ is the row of the $k$-th matrix, $Z_k$ is the number of columns and $d$ is responsible for exponential increase in the core tensors. If $d$ is too high, it results in the loss of spatial information. For a standard forward and backward pass, the time complexity and memory required is $\mathcal{O}(YZ)$ for $W$. 

For BTD-RNN, the time complexity for the forward pass is $\mathcal{O}(N d Y R^d Z_{\text{max}})$ and $\mathcal{0}(N d^2 Y R^d Z_{\text{max}})$ on the backward pass where $J$ is the product of the number of BT parameters for each $d$-th order tensor. For spatial complexity it is $\mathcal{O}(R^d Y)$ on both passes. They find significant improvements over an LSTM baseline network and improvements over a Tensor-Train LSTM~\cite{yu2017long}.

\subsection{Applications of Tensor Decompositions to Convolutional Layers}

\subsubsection{Filter Decompositions}
~\citet{rigamonti2013learning} reduce computation in CNNs by learning a linear combination of separable filters, while maintaining performance. 

For $N$ 2-d filters $\{f^j\}_{1 \leq j \leq N}$, one can obtain a shared set of separable (rank-1) filters by minimizing the objective in \autoref{eq:sep_filt_obj}

\begin{equation}\label{eq:sep_filt_obj}
\argmin_{\{f^j\},\{m^j_i \}} \sum_i \Big( ||x_i - \sum_{j=1}^N f^j * m^j_i||^2_2 + \lambda \sum_{j=1}^{N} ||m_i^j||_1 \Big)
\end{equation}

where $\mat{x}_i$ is an input image, $*$ denotes the convolution product operator, $\{m^j_i\}_{j=1\ldots N}$ are the feature maps obtained during training and $\lambda_1$ is a regularization coefficient. This can be optimized using stochastic gradient descent (SGD) to optimize for $m^j_i$ latent features and $f^j$ filters.

In the first approach they identify low-rank filters using the objective in \autoref{eq:sep_filt_obj_2} to penalize high-rank filters. 

\begin{equation}\label{eq:sep_filt_obj_2}
\text{argmin}_{\{s^j\},\{m^j_i \}} \sum_i \Big( ||x_i - \sum_{j=1}^N s_j * m_j^i||^2_2 + \lambda_1 \sum_{j=1}^{N} ||m_i^j||_1 + \lambda_{*} \sum_{j=1}^{N} ||s^j||_{*}  \Big)
\end{equation}

where the $s_j$ are the learned linear filters, $||*||$ is the
sum of singular values (convex relaxation of the rank), and $\lambda_{*}$ is an additional regularization parameter. The second approach involves separating the optimization of squared difference between the original filter $f^i$ and the weighted combination of learned linear filters $w^j_k s_k$, and the sum of singular values of learned filters $s$. 

\begin{equation}\label{eq:sep_filt_obj_3}
\text{argmin}_{\{s_k\},\{w^j_k \}} \sum_j \Big( ||f^i - \sum_{k=1}^M w^j_k s_k||^2_2 + \lambda_{*} \sum_{k=1}^{M} ||s^j||_{*} \Big)
\end{equation}

They find empirically that decoupling the computation of the non-separable filters
from that of the separable ones leads to better results compared to jointly optimizing over $s^j, m^j_i$ and $w^j_k$ which is a difficult optimization problem. 

\subsubsection{Channel-wise Decompositions}
~\citet{jaderberg2014speeding} propose to approximate filters in convolutional layers using a low-rank basis of filters that have good separability in the spatial filter dimensions but make the contribution of removing redundancy across channels by performing channel-wise low-rank (LR) decompositions (LRD), leading to further speedups. This approach showed significant 2.5x speed ups and maintained performance on character recognition leading to SoTA on standard benchmarks. 

\begin{figure}
    \centering
    \includegraphics[scale=0.34]{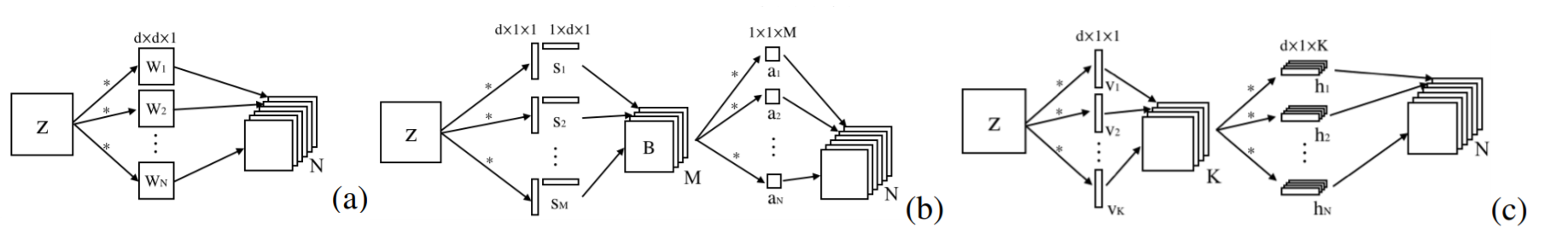}
    \caption{original source:~\citet{jaderberg2014speeding} - Low Rank Expansion Methods: (a) standard CNN filter, (b) LR approximation along the spatial dimension of 2d seperable filters and (c) extending to 3D filters where each conv. layer is factored as a sequence ot two standard conv. layers but with rectangular filters.}
    \label{fig:lre}
\end{figure}


\subsubsection{Combining Filter and Channel Decompositions}
~\citet{yu2017compressing} argue that sparse and low rank decompositions (LRDs) of weight filters should be combined as filters often exhibit both and ignoring either sparsity or LRDs requires iterative retraining and lower comrpession rates. Feature maps are reconstructed using fast-SVD. This approach allowed accuracy to be maintained for higher compression rates in few retraining steps when compared to single approaches (e.g pruning) for AlexNet, VGG-16 (15 time reduction) and LeNet CNN architectures.

\section{Knowledge Distillation}\label{sec:md}
Knowledge distillation involves learning a smaller network from a large network using supervision from the larger network and minimizing the entropy, distance or divergence between their probabilistic estimates. 

To our knowledge,~\citet{bucilua2006model} first explored the idea of reducing model size by learning a student network from an ensemble of models. They use a teacher network to label a large amount of unlabeled data and train a student network using supervision from the pseudo labels provided by the teacher. They find performance is close to the original ensemble with 1000 times smaller network. 

~\citet{hinton2015distilling} a neural network knowledge distillation approach where a relatively small model (2-hidden layer with 800 hidden units and ReLU activations) is trained using supervision (class probability outputs) for the original ``teacher'' model (2-hidden layer, 1200 hidden units). They showed that learning from the larger network outperformed the smaller network learning from scratch in the standard supervised classification setup. In the case of learning from ensemble, the average class probability is used as the target.

The cross entropy loss is used between the class probability outputs of the student output $y^S$ and one-hot target $y$ and a second term is used to ensure that the student representation $z^s$ is similar to the teacher output $z^T$. This is expressed in terms of KL divergence as,

\begin{equation}
\cL_{\text{KD}} = (1 - \alpha)\mathbb{H}(y, y^S) + \alpha \rho^2 \mathbb{H}\Bigg(\phi\Big(\frac{z^T}{\rho}\Big), \phi\Big(\frac{z^S}{\rho}\Big)\Bigg)
\end{equation}

where $\rho$ is the temperature, $\alpha$ balances between both terms, and $\phi$ represents the softmax function. The $\mathbb{H}\big(\phi(\frac{z^T}{\rho}), \phi(\frac{z^S}{\rho})\big)$ is further decomposed into $D_{\text{KL}}\big(\phi(\frac{z^T}{\rho})|\phi(\frac{z^S}{\rho})\big)$ and a constant entropy $\mathbb{H}\big(\phi(\frac{z^T}{\rho})\big)$. 

The idea of training a student network on the logit outputs (i.e log of the predicted probabilities) of the teacher to gain more information from the teacher network can be attributed to the work of~\citet{ba2014deep}. By using logits, as opposed to a softmax normalization across class probabilities for example, the student network better learns the relationship between each class on a log-scale which is more forgiving than the softmax when the differences in probabilities are large.



\subsection{Analysis of Knowledge Distillation}
The works in this subsection provide insight into the relationship between the student and teacher networks for various tasks, teacher size and network size. We also discuss work that focuses on what is required to train a well-performing student network e.g use of early stopping~\citep{tarvainen2017mean} and avoiding training the teacher network with label smoothing~\citep{muller2019does}.

\paragraph{Theories of Why Knowledge Distillation Works}

For a distilled linear classifier, ~\citet{phuong2019towards} prove a generalization bound that shows the fast convergence of the expected loss. In the case where the number of samples is less than the dimensionality of the feature space, the weights learned by the student network are projections of the weights in the student network onto the data span. Since gradient descent makes updates that are within the data space, the student network is bounded in this space and therefore it is the best student network can approximate of the teacher network weights w.r.t the Euclidean norm. 
From this proof, they identify 3 important factors that contribute that explain the success of knowledge distillation - (1)
the geometry of the data distribution that makes up the separation between classes greatly effects the student networks convergence rate (2), gradient descent is biased towards a desirable minimum in the distillation objective and (3) the loss monotonically decreases proportional to size of the training set.

\paragraph{Teacher Assistant Knowledge Distillation}

~\citet{mirzadeh2019improved} show that the performance of the student network degrades when the gap between the teacher and the student is too large for the student to learn from. Hence, they propose an intermediate `teaching assistant' network to supervise and distil the student network where the intermediate networks is distilled from the teacher network. 

\begin{figure}
    \centering
    \includegraphics[scale=0.4]{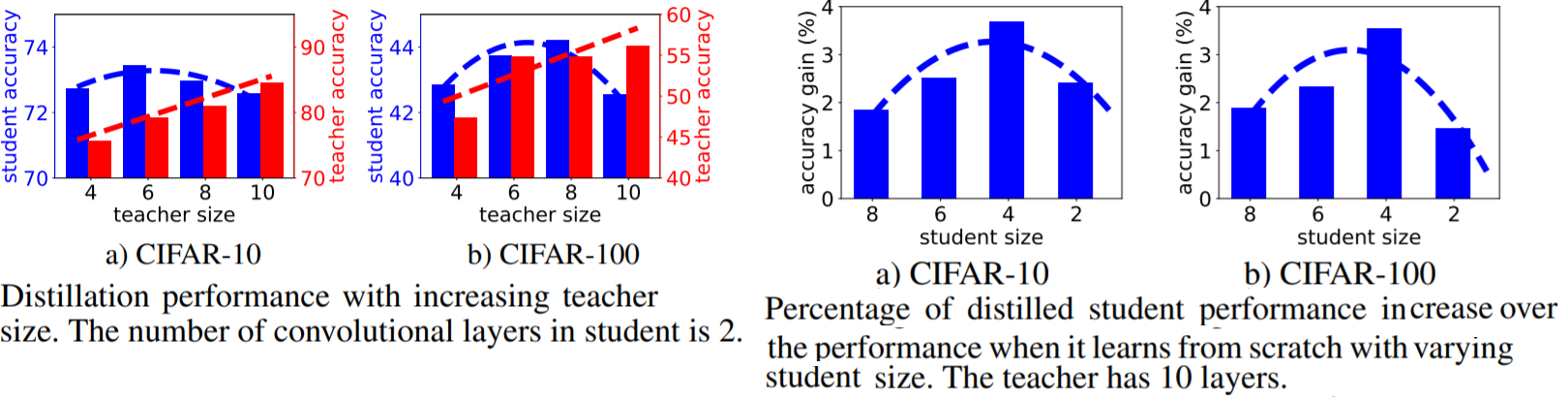}
    \caption{original source~\citet{mirzadeh2019improved}}
    \label{fig:distil_perf}
\end{figure}

~\autoref{fig:distil_perf} shows their plot, where on the left side a) and b) we see that as the gap between the student and teacher networks widen when the student network size is fixed, the performance of student network gradually degrades. Similarly, on the right hand side, a similar trend is observed when the student network size is increased with a fixed teacher network.

Theoretical analysis and extensive experiments on CIFAR-10,100 and ImageNet datasets and on CNN and ResNet architectures substantiate the effectiveness of our proposed approach.

\begin{wrapfigure}{R}{8cm}
    \centering
    \includegraphics[scale=0.33]{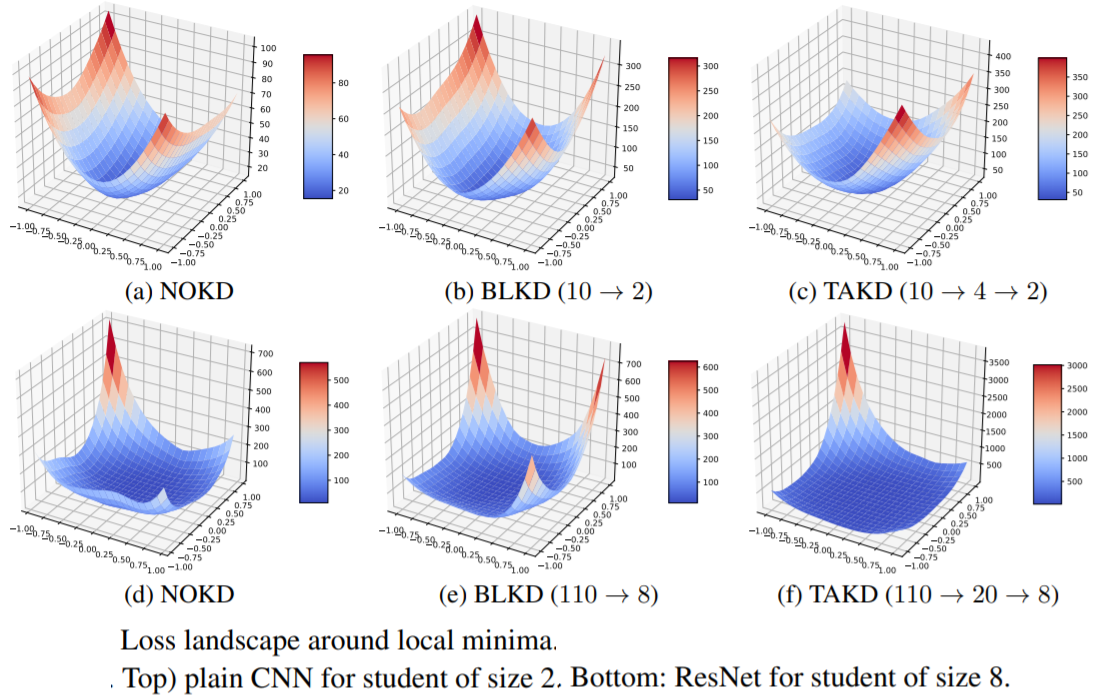}
    \caption{original source:~\citet{mirzadeh2019improved}}
    \label{fig:loss_landscape}
\end{wrapfigure}

Their \autoref{fig:loss_landscape} shows the loss surface of CNNs trained on CIFAR-100 for 3 different approaches: (1) no distillation, (2) standard knowledge distillation and (3) teaching assisted knowledge distillation. As shown, the teaching assisted knowledge distillation has a smoother surface around the local minima, corresponding to more robustness when the inputs are perturbed and better generalization. 

\paragraph{On the Efficacy of Knowledge Distillation}

~\citet{cho2019efficacy} analyse what are some of the main factors in successfully using a teacher network to distil a student network. Their main finding is that when the gap between the student and teacher networks capacity is too large, distilling a student network that maintains performance or close to the teacher is either unattainable or difficult. They also find that the student network can perform better if early stopping is used for the teacher network, as opposed to training the teacher network to convergence. 

\begin{figure}
    \centering
    \includegraphics[scale=0.4]{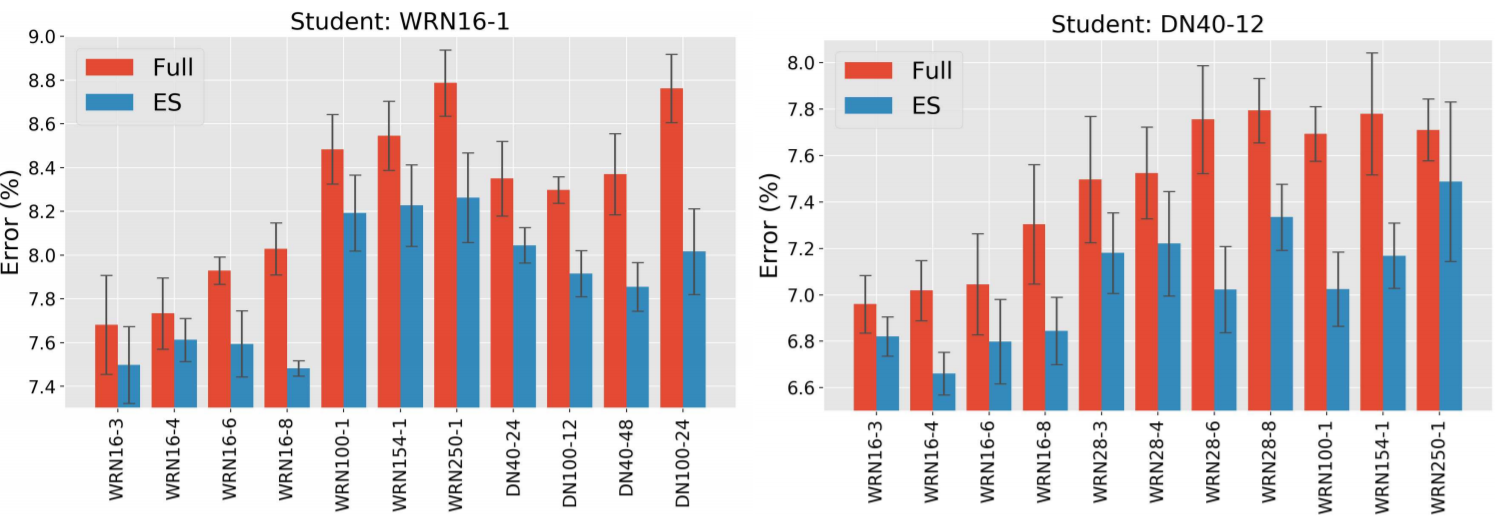}
    \caption{original source~\citet{cho2019efficacy}: Early Stopping Teacher Networks to Improve Student Network Performance}
    \label{fig:early_stop_teacher}
\end{figure}

\autoref{fig:early_stop_teacher} shows that teachers (DenseNet and WideResNet) trained with early stopping are better suited as supervisors for the student network (DenseNet40-12 and WideResNet16-1).

\paragraph{Avoid Training the Teacher Network with Label Smoothing}
~\citet{muller2019does} show that because label smoothing forces the same class sample representations to be closer to each other in the embedding space, it provides less information to student network about the boundary between each class and in turn leads to poorer generlization performance. They quantify the variation in logit predictions due to the hard targets using mutual information between the input and output logit and show that label smoothing reduces the mutual information. Hence, they draw a connection between label smoothing and information bottleneck principle and show through experiments that label smoothing can implicitly calibrate the predictions of a DNN. 


\paragraph{Distilling with Noisy labels}
~\citet{sau2016deep} propose to use noise to simulate learning from multiple teacher networks by simply adding Gaussian noise the logit outputs of the teacher network, resulting in better compression when compared to training with the original logits as targets for the teacher network. They choose a set of samples from each mini-batch with a probability $\alpha$ to perturbed by noise while the remaining samples are unchanged. They find that a relatively high $\alpha=0.8$ performed the best for image classification task, corresponding to 80\% of teacher logits having noise.

~\citet{li2017learning} distil models with noisy labels and use a small dataset with clean labels, alongside a knowledge graph that contains the label relations, to estimate risk associated with training using each noisy label. 
A model is trained on the clean dataset $D_c$ and the main model is trained over the whole dataset $D$ with noisy labels using the loss function,

\begin{equation}
\mathcal{L}_D(\vec{y}_i, f(\vec{x}_i)) = \lambda l(\vec{y}_i, f(\vec{x}_i)) + (1 - \lambda)l(\vec{s}_i, f(\vec{x}_i))
\end{equation}

where $s_i=\delta[f^D_c(\vec{x}i)]$. The first loss term is cross entropy between student noisy and noisy labels and the second term is the loss between the the hard target $s_i$ given by the model trained on clean data and the model trained on noisy data. 

They also use pseudo labels $\hat{y}\lambda_i= \lambda \vec{y}_i+ (1 - \lambda)\vec{s}_i$ that combine noisy label $\vec{y}_i$ with the output $\vec{s}_i$ trained on $D_c$. This motivated by the fact that both noisy label and the predicted labels from clean data are independent and this can be closer to true labels $y^{*}_i$ under conditions which they further detail in the paper. 

To avoid the model trained on $D_c$ overfitting, they assign label confidence score based on related labels from a knowledge graph, resulting in a reduction in model variance during knowledge distillation.

\paragraph{Distillation of Hidden Layer Activation Boundaries}
Instead of transferring the outputs of the teacher network,~\citet{heo2019knowledge} transfer activation boundaries, essentially outputs which neurons are activated and those that are not. They use an activation loss that minimizes the difference between the student and teacher network activation boundaries, unlike previous work that focuses on the activation magnitude. Since gradient descent updates cannot be used on the non-differentiable loss, they propose an approximation of the activation transfer loss that can be minimized using gradient descent. The objective is given as, 

\begin{equation}\label{eq:trans_act_loss}
\cL(I) =||\rho(\cT(\cI))\sigma \big(\mu \mathbf{1} - r(\cS(\cI))\big)+ \big(1 - \rho (\cT(\cI))\big) \circ \sigma ( \mu \mathbf{1} + r(\cS(\cI))\big)||_2^2
\end{equation}

where $\cS(\cI)$ and $\cT(\cI)$ are the neuron response tensors for student and teacher networks,  $\rho(\cT(I))$ is the the activation of teacher neurons corresponding to class labels, $r(\cS(\cI))$ is the , $r$ is a connector function (a fully connected layer in their experiments) that converts a neuron response vector of student to the same size as the teacher vector, $\circ$ is elementwise product of vectors and $\mu$ is the margin to stabilize training.



\paragraph{Simulating Ensembled Teachers Training}
~\citet{park2020improved} have extended the idea of student network learning from a noisy teacher to speech recognition and similarly found high compression rates. 
~\citet{han2018co} have pointed out that co-teaching (where two networks learn from each other where one has clean outputs and the other has noisy outputs) avoids a single DNN from learning to memorize the noisy labels and select samples from each mini-batch that the networks should learn from and avoid those samples which correspond to noisy labels. Since both networks have different ways of learning, they filter different types of error occurring from the noisy labels and this information is communicated mutually. This strategy could also be useful for using the teacher network to provide samples to a smaller student network that improve the learning of the student. 

\subsection{Data-Free Knowledge Distillation}
~\citet{lopes2017data} aim to distill in the scenario where it is not possible to have access to the original data the teacher network was trained on. This can occur due to privacy issues (e.g personal medical data, models trained case-based legal data) or the data is no longer available or some way corrupted. They store the sufficient statistics (e.g mean and covariance) of activation outputs from the original data along with the pretrained teacher network to reconstruct the original training data input. This is achieved by trying to find images that have the highest representational similarity to those given by the representations from the activation records of the teacher network. Gaussian noise is passed as input to the teacher and update gradients to the noise to minimize the difference between the recorded activation outputs and those of the noisy image and repeat this to reconstruct the teachers view of the original data. 

\begin{figure}
    \centering
    \includegraphics[scale=0.4]{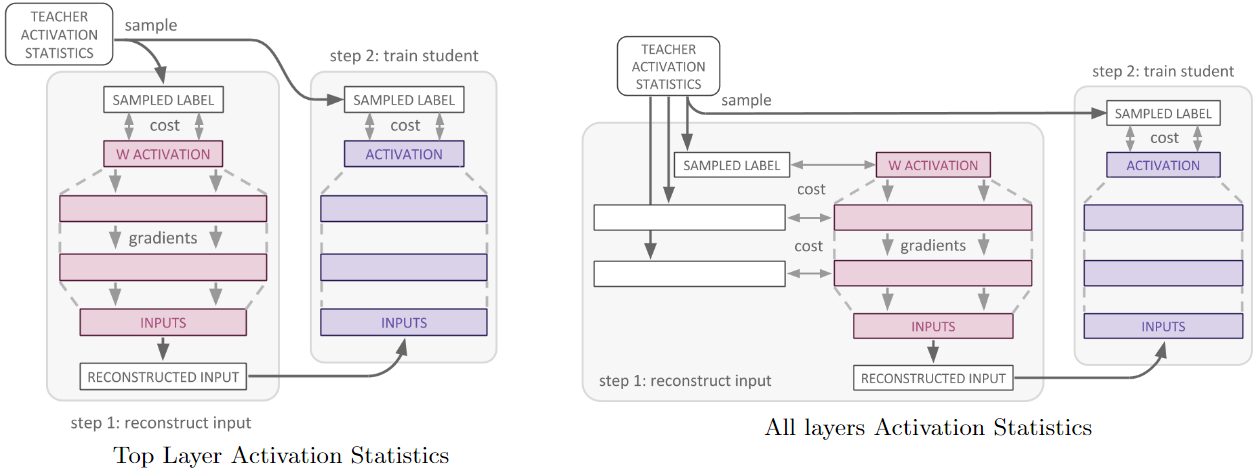}
    \caption{original source~\citet{lopes2017data}: Data Free Knowledge Distillation. The left shows }
    \label{fig:data_free_md}
\end{figure}

The left figure in \autoref{fig:data_free_md} shows the activation statistics for the top layer and a sample drawn that is used to optimize the input to teacher network to reconstruct the activations. The reconstructed input is then fed to the student network. On the right, the same procedure follows but for reconstructing activations for all layers of the teacher network. 

They manage to compress the teacher network to half the size in the student network using the reconstructed inputs constructed from using the metadata. The amount of compression achieved is contingent on the quality of the metadata, in their case they only used activation statistics. We posit that the notion of creating synthetic data from summary statistics of the original data to train the student network is worth further investigation. 

\paragraph{Layer Fusion}
Layer Fusion (LF)~\citep{neill2020compressing} is a technique to identify similar layers in very deep pretrained networks and fuse the top-k most similar layers during retraining for a target task. Various alignments measures are proposed that have desirable properties of for layer fusion and freezing, averaging and dynamic mixing of top-k layer pairs are all experimented with for fusing the layers. This can be considered as unique approach to knowledge distillation as it does aim to preserve the knowledge in the network while preserving network density, but without having to train a student network from scratch.

\subsection{Distilling Recurrent (Autoregressive) Neural Networks}

Although the work by ~\citet{bucilua2006model} and ~\citet{hinton2015distilling} has often proven successful for reducing the size of neural models in other non-sequential tasks, many sequential tasks in NLP and CV have high-dimensional outputs (machine translation, pixel generation, image captioning etc.). This means using the teachers probabilistic outputs as targets can be expensive.


~\citet{kim2016sequence} use the teachers hard targets (also 1-hot vectors) given by the highest scoring beam search prediction from an encoder-decoder RNN, instead of the soft output probability distribution. The  teacher distribution $q(y_t|x)$ is approximated  by  its  mode:$q(y_s|x) \approx 1{t= \argmax_{y_t \in \mathcal{Y}} q(y_t|x)}$ with the following objective

\begin{equation}
\mathcal{L}_{SEQ-MD} = - \mathbb{E}_{x \sim D} \sum_{y_t \in \mathcal{Y}} p(y_t|x) \log p(y_t|x) \approx - \mathbb{E}_{x \sim D},\hat{y}_s = \argmax_{y_t \in \mathcal{Y}} q(y_t|x)[\log p(y_t =\hat{y}_s|x)]    
\end{equation}

where $y_t \in \mathcal{Y}$ are teacher targets (originally defined by the predictions with the highest scoring beam search) in the space of possible target sequences. When the temperature $\tau \to 0$, this is equivalent to standard knowledge distillation. 

In sequence-level interpolation, the targets from the teacher with the highest \textit{similarity} with the ground truth are used as the targets for the student network. Experiments on NMT showed performance improvements compared to soft targets and further pruning the distilled model results in a pruned student that has 13 times fewer parameters than the teacher network with a 0.4 decrease in BLEU metric. 

\subsection{Distilling Transformer-based (Non-Autoregressive) Networks}\label{sec:distil_trans}
Knowledge distillation has also been applied to very large transformer networks, predominantly on BERT~\citep{devlin2018bert} given its wide success in NLP. Thus, there has been a lot of recent work towards reducing the size of BERT and related models using knowledge distillation. 

\paragraph{DistilBERT}~\citet{sanh2019distilbert} achieves distillation by training a smaller BERT on very large batches using gradient accumulation, uses dynamic masking, initializes the student weights with teacher weights and removes the next sentence prediction objective. They train the smaller BERT model on the original data BERT was trained on and fine that DistilBERT is within 3\% of the original BERT accuracy while being 60\% faster when evaluated on the GLUE~\citep{wang2018glue} benchmark dataset.

\paragraph{BERT Patient Knowledge Distillation}
Instead of minimizing the soft probabilities between the student and teacher network outputs,~\citet{sun2019patient} propose to also learn from the intermediate layers of the BERT teacher network by minimizing the mean squared error between adjacent and normalized hidden states. This loss is combined with the original objective proposed by~\citet{hinton2015distilling} which showed further improves in distilling BERT on the GLUE benchmark datasets~\citep{wang2018glue}.

\paragraph{TinyBERT}
TinyBERT~\citep{jiao2019tinybert} combines multiple Mean Squared Error (MSE) losses between embeddings, hidden layers, attention layers and prediction outputs between $S$ and $T$. The TinyBERT distillation objective is shown below, where it combines multiple reconstruction errors between $S$ and $T$ embeddings (when m=0), between the hidden and attention layers of $S$ and $T$ when $ M \geq m > 0$ where $M$ is index of the last hidden layer before prediction layer and lastly the cross entropy between the predictions where $t$ is the temperature of the softmax. 

\[
L_{layer}\big(S_m, T_g(m)  \big) =
\begin{cases}
  \text{MSE}(\mat{E}^S \mat{W}_e \mat{E}^T) & m = 0 \\
    \text{MSE}(\mat{H}^S \mat{W}_h, \mat{H}^T) + \frac{1}{h} \sum_{i=1}^h \text{MSE}(\mat{A}^{S}_i, \mat{A}^T_i)  & M \geq m > 0 \\
     \text{softmax}(\vec{z}^T) \cdot \text{log-softmax}(\vec{z}^S/t) & m = M + 1 \\
\end{cases}
\]

Through many ablations in experimentations, they find distilling the knowledge from multi-head attention layers to be an important step in improving distillation performance.

\paragraph{ALBERT}~\citet{lan2019albert} proposed factorized embeddings to reduce the size of the vocabulary embeddings and parameter sharing across layers to reduce the number of parameters without a performance drop and further improve performance by replacing next sentence prediction with an inter-sentence coherence loss. ALBERT is. 5.5\% the size of original BERT and has produced state of the art results on top NLP benchmarks such as GLUE~\citep{wang2018glue}, SQuAD~\citep{rajpurkar2016squad} and RACE~\citep{lai2017race}.

\begin{wrapfigure}{r}{8cm}
    \centering
    \includegraphics[scale=0.3]{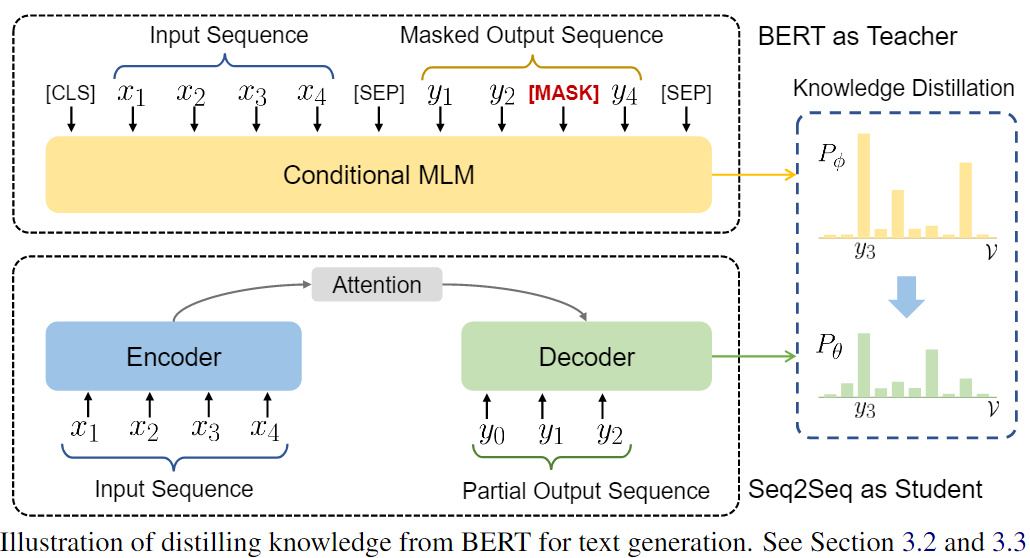}
    \caption{original source \citep{chen2019distilling}: BERT Distillation for Text Generation}
    \label{fig:bert_distil_text_gen}
\end{wrapfigure}

\paragraph{BERT Distillation for Text Generation}
~\citet{chen2019distilling} use a conditional masked language model that enables BERT to be used on generation tasks. The outputs of a pretrained BERT teacher network are used to provide sequence-level supervision to improve Seq2Seq model and allow them to plan ahead.~\autoref{fig:bert_distil_text_gen} illustrates the process, showing where the predicted probability distribution for the remaining tokens is minimized with respect to the masked output sequence from the BERT teacher.

\paragraph{Applications to Machine Translation}
~\citet{zhou2019understanding} seek to better understand why knowledge distillation leads to better non-autoregressive distilled models for machine translation. They find that the student network finds it easier to model variations in the output data since the teacher network reduces the complexity of the dataset.

\subsection{Ensemble-based Knowledge Distillation}

\paragraph{Ensembles of Teacher Networks for Speech Recognition}
~\citet{chebotar2016distilling} use the labels from an ensemble of teacher networks to supervise a student network trained for acoustic modelling. To choose a good ensemble, one can select an ensemble where each individual model potentially make different errors but together they provide the student with strong signal for learning. Boosting weights each sample based proportional to its misclassification rate. Similarly this can used on the ensemble to learn which outputs from each model to use for supervision. 
Instead of learning from a combination of teachers that are best by using an oracle that approximates the best outcome of the ensemble for automatic speech recognition (ASR) as

\begin{equation}
P_{\text{oracle}}(s|x) =\sum^N_{i=1}[O(u) =i]P_i(s|x) =P_O(u)(s|x)
\end{equation}

where the oracle $O(u) \in 1 \ldots N$ that contains $N$ teachers assigns all the weight to the model that has the lowest word errors for a given utterance $u$. Each model is an RNN of different architecture trained with different objectives and the student $s$ is trained using the Kullbeck Leibler (KL) divergence between oracle assigned teachers output and the student network output. They achieve an 8.9\% word error rate improvement over similarly structured baseline models.

~\citet{freitag2017ensemble} apply knowledge distillation to NMT by distilling an ensemble of networks and oracle BLEU teacher network into a single NMT system. The find a student network of equal size to the teacher network outperforms the teacher after training. They also reduce training time by only updating the student networks with filtered samples based on the knowledge of the teacher network which further improves translation performance.

~\citet{cui2017knowledge} propose two strategies for learning from an ensemble of teacher network; (1) alternate between each teacher in the ensemble when assigning labels for each mini-batch and (2) simultaneously learn from multiple teacher distributions via data augmentation. They experiment on both approaches where the teacher networks are deep VGG and LSTM networks from acoustic models.


~\citet{cui2017knowledge} extend knowledge distillation to multilingual problems. They use multiple pretrained teacher LSTMs trained on multiple low-resource languages to distil into a smaller standard (fully-connected) DNN. They find that student networks with good input features makes it easier to learn from the teachers labels and can improve over the original teacher network. Moreover, from their experiments they suggest that allowing the ensemble of teachers learn from one another, the distilled model further improves.

\paragraph{Mean Teacher Networks}
~\citet{tarvainen2017mean} find that averaging the models weights of an ensemble at each epoch is more effective than averaging label predictions for semi-supervised learning. This means the Mean Teacher can be used as unsupervised learning distillation approach as the distiller does not need labels. than methods which rely on supervision for each ensemble model. They find this straightforward approach to outperform previous ensemble based distillation approaches~\citep{laine2016temporal} when only given 1000 labels on the Street View House View Number~\citep[SVHN;][]{goodfellow2013multidig} dataset. 
Moreover, using Mean Teacher networks with Residual Networks achieved SoTA with 4000 labels from 10.55\% error to 6.28\% error.

\paragraph{on-the-fly native ensemble}
~\citet{zhu2018knowledge} focus on using distillation on the fly in a scenario where the teacher may not be fully pretrained or it does not have a high capacity. This reduces compression from a two-phase (pretrain then distil) to one phase where both student and teacher network learn together. They propose an On the fly Native Ensemble (ONE) learning strategy that essentially learns a strong teacher network that assists the student network as it is learning. Performance improvements for on the fly distillation are found on the top benchmark image classification datasets.

\paragraph{Multi-Task Teacher Networks}
~\citet{liu2019improving} perform knowledge distillation for performing multi-task learning (MTL), using the outputs of teacher models from each natural language understanding (NLU) task as supervision for the student network to perform MTL. The distilled MT-DNN outperforms the original network on 7 out of 9 NLU tasks (includes sentence classification, pairwise sentence classification and pairwise ranking) on the GLUE~\citep{wang2018glue} benchmark dataset. 

\subsection{Reinforcement Learning Based Knowledge Distillation}
Knowledge distillation has also been performed using reinforcement learning (RL) where the objective is to optimize for accumulated of rewards where the reward function can be task-specific. Since not all problems optimize for the log-likelihood, standard supervised learning can be a poor surrogate, hence RL-based distillation can directly optimize for the metric used for evaluaion.

\paragraph{Network2Network Compression}
~\citet{ashok2017n2n} propose Network to Network (N2N) compression in policy gradient-based models using a RNN policy network that removes layers from the `teacher' model while another RNN policy network then reduces the size of the remaining layers. The resulting policy network is trained to find a locally optimal student network and accuracy is considered the reward signal. The policy networks gradients are updated accordingly, achieving a compression ratio of $10$ for ResNet-34 while maintaining similar performance to the original teacher network. 


\paragraph{FitNets}
~\citet{romero2014fitnets} propose a student network that has deeper yet smaller hidden layers compared to the teacher network. They also constrain the hidden representations between the networks to be similar. Since the hidden layer size for student and teacher will be different, they project the student layer to into an embedding space of fixed size so that both teacher and student hidden representations are of the same size.
~\autoref{eq:fitnet_loss} represents the Fitnet loss where the first term represents the cross-entropy between the target $y_{\text{true}}$ and the student probability $P_S$, while $H(P^{\tau}_T,P^{\tau}_S)$ represents the cross entropy between 
the normalized and flattened teachers hidden representation $P^{\tau}_T$ and the normalized student hidden representation $P^{\tau}_S$ where $\gamma$ controls the influence of this similarity constraint. 

\begin{equation}\label{eq:fitnet_loss}
    \cL_{\text{MD}}(\mat{W}_S) = H(y_{\text{true}},P_S) + \gamma H(P^{\tau}_T,P^{\tau}_S)
\end{equation}

\autoref{eq:conv_regressor} shows the loss between the teacher weights $\mat{W}_{Guided}$ for a given layer and the reconstructed weights $\mat{W}_r$ which are the weights of a corresponding student network projected using a convolutional layer (cuts down computation compared to a fully-connected projection layer) to the same hidden size of the teacher network weights. 

\begin{equation}\label{eq:conv_regressor}
    \cL_{\text{HT}} (\mat{W}_{T},\mat{W}_r) = \frac{1}{2}||u_h(\vec{x}; \mat{W}_{\text{Hint}}) - r(v_g(\vec{x};\mat{W}_{\text{T}});\mat{W}_r)||^2
\end{equation}

where $u_h$ and $v_g$ are the teacher/student deep nested functions up to their respective hint/guided layers with parameters $\mat{W}_{\text{Hint}}$ and $\mat{W}_{\text{Guided}}$, $r$ is the regressor function on top of the guided layer with parameters $\mat{W}_r$. Note that the outputs of uh and r have to be comparable, i.e., $u_h$ and $r$ must
be the same non-linearity. The teacher tries to imitate the flow matrices from the teacher which are defined as the inner product between feature maps, such as layers in a residual block.

\subsection{Generative Modelling Based Knowledge Distillation}
Here, we describe how two commonly used generative models, variational inference (VI) and generative adversarial networks (GANs), have been applied to learning a student networks.

\subsubsection{Variational Inference Learned Student}

\begin{wrapfigure}{R}{6.5cm}
\vspace{-3em}
    \centering
    \includegraphics[scale=0.25]{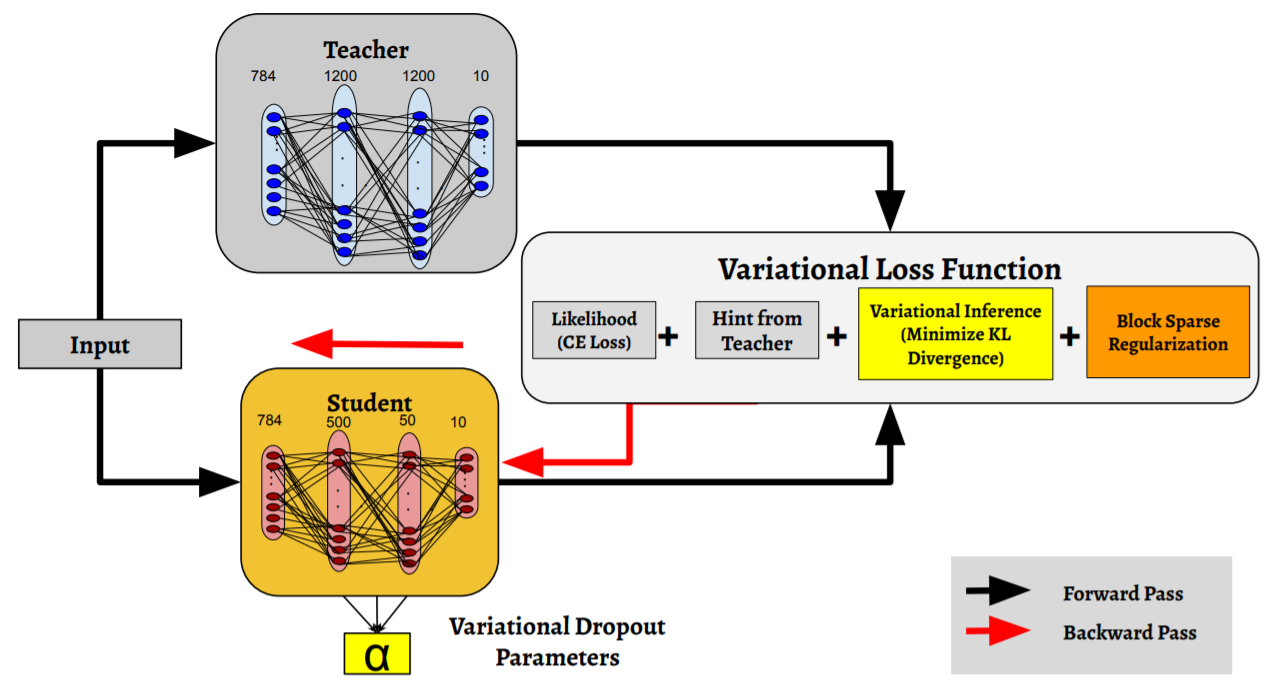}
    \caption{Variational Student Framework (original source:~\citet{hegde2019variational})}\label{fig:var_student}
\end{wrapfigure}

~\citet{hegde2019variational} propose a variational student whereby VI is used for knowledge distillation. The parameters induced by using VI-based least squares objective are sparse, improving the generalizability of the student network. Sparse Variational Dropout (SVD)~\citet{kingma2015variational,molchanov2017variational} techniques can also be used in this framework to promote sparsity in the network. The VI objective is shown in \autoref{eq:var_student}, where $\vec{z}^s$ and $\vec{z}^t$ are the output logits from student and teacher networks.

\begin{multline}\label{eq:var_student}
    \cL(\vec{x}, \vec{y},\mat{W}_s,\mat{W}_t, \alpha) = - \frac{1}{N}\sum_{n=1}^N \vec{y}_n \log(\vec{z}^s_n) +  \lambda_T \Bigg[ 2T^2 D_{\text{KL}}\Bigg(  \sigma^{'} \Big(\frac{\vec{z}^s}{T}\Big) \Big|\Big|  \sigma^{'}  \Big(\frac{\vec{z}^t}{T}\Big)\Bigg)\Bigg] \\
    + \lambda_V \cL_{\text{KL}}(\mat{W}_s, \alpha) + \lambda_g \sum_{m=1}^M\Big|\max_{n,k,h,l} W_{T:S}(m, n, k, h, l)\Big|
\end{multline}

\autoref{fig:var_student} shows their training procedure and loss function that consist of the learning compact and sparse student networks. The roles of different terms in variational loss function are:
likelihood - for independent student network’s learning; hint - learning induced from teacher network; variational term - promotes sparsity by optimizing variational dropout parameters, $\alpha$; Block Sparse Regularization - promotes and transfers sparsity from the teacher network.

\subsubsection{Generative Adversarial Student}
GANs train a binary classifier $f_w$ to discriminate between real samples $x$ and generated samples $g_{\theta}(z)$ that are given by a generator network $g_{\theta}$ and $z$ is sampled from $p_g$ a known distribution e.g a Gaussian. A minimax objective is used to minimize the misclassifications of the discriminator while maximizing the generators accuracy of tricking the discriminator. This is formulated as, 

\begin{equation}\label{eq:gan_obj}
\min_{\theta \in \Theta} \max_{\vec{w} \in W} \mathbb{E}_{\vec{x} \sim p_{\text{data}}}[\log(f_{\vec{w}}(\vec{x})] + \mathbb{E}_{\vec{z} \sim p_{\vec{z}}}[\log(1  - f_{\vec{w}}( g_{\theta}(\vec{z}))) ]
\end{equation}

where the global minimum is found when the generator distribution $p_g$ is similar 
to the data distribution $p_{data}$ (referred to as the nash equilibrium).

~\citet{wang2018kdgan} learn a Generative Adversarial Student Network where the generator learns from the teacher network using the minimax objective in \autoref{eq:gan_obj}. They reduce the variance in gradient updates which leads less epochs requires to train to convergence, by using the Gumbel-Max trick in the formulation of GAN knowledge distillation.


First they propose Naive GAN (NaGAN) which consists of a classifier $C$ and a discriminator $D$ where $C$ generates pseudo labels given a sample $x$ from a categorical distribution  $p_c(\vec{y}|\vec{x})$ and $D$ distinguishes between the true targets and the generated ones. The objective for NaGAN is express as,

\begin{equation}
\min_{c} \max_{d} V(c, d) = \mathbb{E}_{\vec{y} \sim p_u} [\log p_d(\vec{x}, \vec{y})] + \mathbb{E}_{\vec{y} \sim p_c}[\log(1 - p^{\varrho}_d(\vec{x}, \vec{y}))]    
\end{equation}

where $V(c, d)$ is the value function. The scoring functions of $C$ and $D$ are $h(\vec{x}, y)$ and $g(x,y)$ respectively. Then $p_c(y|\vec{x})$ and $p^{\varrho}_d(\vec{x},y)$ are expressed as,

\begin{equation}
p_c(y|\vec{x}) = \phi(h(\vec{x},y)) ,\quad p^{\varrho}_d(\vec{x},y) = \sigma(g(\vec{x},y))  
\end{equation}

where $\phi$ is the softmax function and $\sigma$ is the sigmoid function. However, NaGAN requires a large number of samples and epochs to converge to nash equilibrium using this objective, since the gradients from $D$ that update $C$ can often vanish or explode. 

This brings us to their main contribution, Knowledge Distilled GAN (KDGAN).

KDGAN somewhat remedy the aforementioned converegence problem by introducing a pretrained teacher network $T$ along with $C$ and $D$. The objective then consists of a distillation $\ell_2$ loss component between $T$ and $C$ and adversarial loss between $T$ and $D$. Therfore, both $C$ and $T$ aim to fool $D$ by generating fake labels that seem real, while $C$ tries to distil the knowledge from $T$ such that both $C$ and $T$ agree on a good fake label. 

\begin{figure}
    \centering
    \includegraphics[scale=0.4]{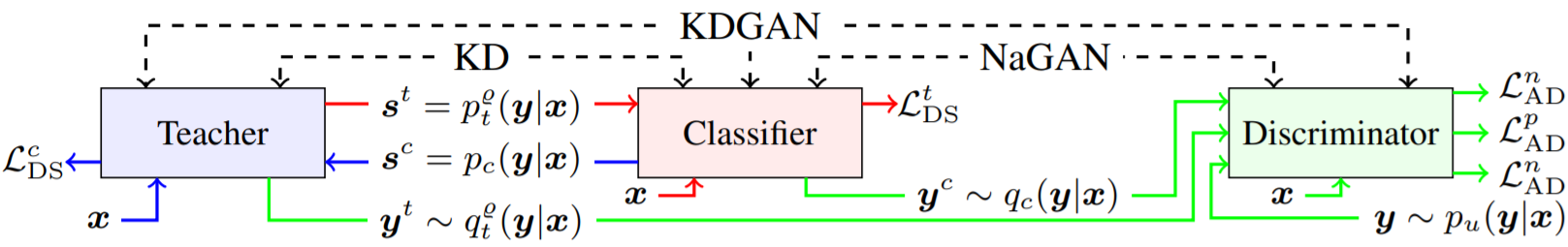}
    \caption{original source~\citet{wang2018kdgan}: Comparison among KD, NaGAN, and KDGAN}
    \label{fig:kdgan}
\end{figure}

The student network convergence is tracked by observing the generator outputs and loss changes. Since the gradient from $T$ tend to have low variance, this can help $C$ converge faster, reaching a nash equilibrium. The difference between these models is illustrated in \autoref{fig:kdgan}.

\paragraph{Compressing Generative Adversarial Networks}

\begin{wrapfigure}{R}{7cm}
\vspace{-1em}
    \centering
    \includegraphics[scale=0.3]{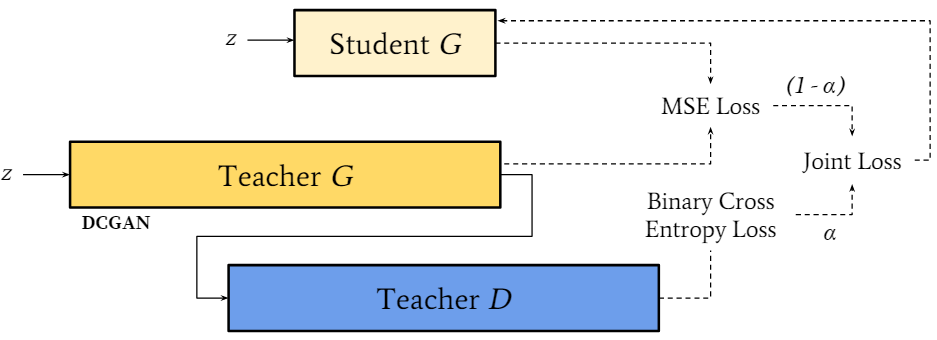}
    \caption{original source~\citet{aguinaldo2019compressing}: Student Teacher GAN Training}
    \label{fig:stud_gan}
\end{wrapfigure}

~\citet{aguinaldo2019compressing} compress GANs achieving high compression ratios (58:1 on CIFAR-10 and 87:1 CelebA) while maintaining high Inception Score (IS) and low Frechet Inception Distance (FID). They're main finding is that a compressed GAN can outperform the original overparameterized teacher GAN, providing further evidence for the benefit of compression in very larrge networks.~\autoref{fig:stud_gan} illustrates the student-teacher training using a joint loss between the student GAN discriminator and teacher generator DCGAN. 

Student-teacher training framework with joint loss for student training. The teacher generator was trained using deconvolutional GAN~\citep[DCGAN;][]{radford2015unsupervised} framework.

They use a joint training loss to optimize that can be expressed as,

\begin{equation}
\min_{\theta \in \Theta} \max_{\vec{w} \in W} \mathbb{E}_{\vec{x} \sim p_{\text{data}}}[\log(f_{\vec{w}}(\vec{x})] + \mathbb{E}_{z \sim p_z} \Big[ \alpha \log (1 - f_{\vec{w}}(g_{\theta}(z))) + (1 - \alpha) g_{\text{teacher}}||(z) - g_{\theta}(z)||^2 \Big]
\end{equation}

where $\alpha$ controls the influence of the MSE loss between the logit predictions $g_{\text{teacher}}(z)$ and $g_{\theta}(z)$ of teacher and student respectively. The terms with expectations correspond to the standard adversarial loss.

\subsection{Pairwise-based Knowledge Distillation}
Apart from pointwise classification tasks, knowledge distillation has also been performed for pairwise tasks. 

\paragraph{Similarity-preserving Knowledge Distillation}

Semantically similar inputs tend to have similar activation patterns. Based on this premise,~\citet{tung2019similarity} have propose knowledge distillation such that input pair similarity scores from the student network are similar to those from the teacher network. This can be a pairwise learning extension of the standard knowledge distillation approaches. 

They aim to preserve similarity between student and pretrained teacher activations for a given batch of similar and dissimilar input pairs. For a batch $b$, a similarity matrix $G(l^{'})_S \in \mathbb{R}^{b \times b}$ is produced between their student activations $A^{(l^{'})}_S$ at the $l^{'}$ layer and teacher activations $A^{(l)}_T$ at the l-th layer. The objective is then defined as the cross entropy between the student logit output $\sigma(\vec{z}_s)$
and target $y$ summed with the similarity preserving distillation loss component on the RHS of \autoref{eq:spdl},

\begin{equation}\label{eq:spdl}
 \mathcal{L} = \mathcal{L}_{\text{ce}} (\vec{y}, \phi(\vec{Z}_S)) + \frac{\gamma}{b^2} \sum_{(l, l^{'}) \in \mathcal{I}} ||\mat{G}^{(l)}_T  - \mat{G}^{(l')}_S ||^2_F
\end{equation}

where $||\cdot||_F$ denotes the Frobenius norm, $\mathcal{I}$ is the total number of layer pairs considered and $\gamma$ controls the influence of similarity preserving term between both networks.  

In the transfer learning setting, their experiments show that similarity preserving can be a robust way to deal with domain shift. Moreover, this method complements the SoTA attention transfer~\citep{zagoruyko2016paying} approach.

\paragraph{Contrastive Representation Distillation}
Instead of minimizing the KL divergence between the scalar outputs of teacher network $T$ and student network $S$,~\citet{tian2019contrastive} propose to preserve structural information of the embedding space. Similar to~\citet{hinton2012improving}, they force the representations between the student and teacher network to be similar but instead use a constrastive loss that moves positive paired representations closer together while positive-negative pairs away. This contrastive objective is given by,

\begin{multline}
    f^{S*} = \argmax_{f^S} \max_h \cL_{\text{critic}}(h) = \\
    \argmax_{f^S} \max_h   \mathbb{E}_q(T ,S|C=1)[\log h(T, S)] +  N \mathbb{E}_{q}(T ,S|C=0)[\log(1 - h(T, S))]
\end{multline}

where $h(T, S) = \frac{e^{g^T(T)' g^S(S)'/\tau}}{ e^{g^T(T) g^S(S)/\tau} + NM}$, $M$ is number of data samples, $\tau$ is the temperature. If the dimensionality of the outputs from $g^T$ and $g^S$ are not equal, a linear transformation is made to fixed size followed by an $\ell_2$ normalization.

\begin{figure}
    \centering
    \includegraphics[scale=0.4]{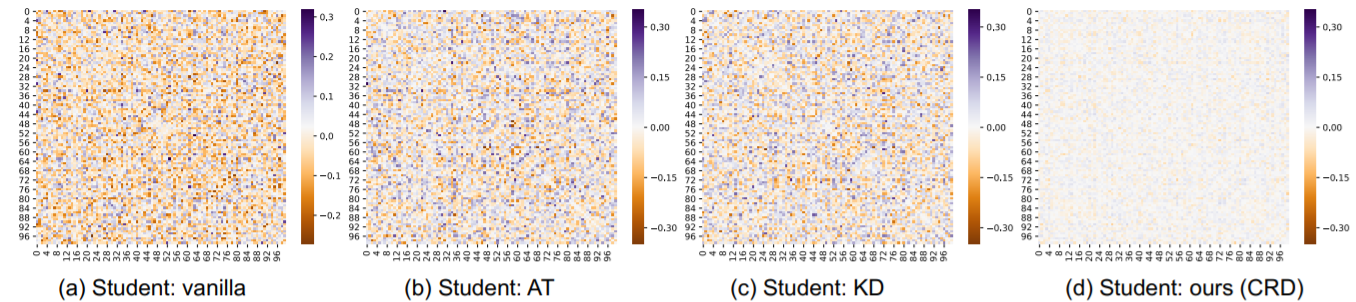}
    \caption{original source:~\citet{tian2019contrastive}}
    \label{fig:constrastive_distil_corr_plot}
\end{figure}

~\autoref{fig:constrastive_distil_corr_plot} demonstrates how the correlations between student and teacher network are accounted for in CRD (d) while in standard teacher-student networks (a) ignores the correlations and to a less extent this is also found for attention transfer (b)~\citep{zagoruyko2016wide} and the student network distilled by KL divergence (c)~\citep{hinton2015distilling}.


\paragraph{Distilling SimCLR}
~\citet{chen2020big} shows that an unsupervised learned constrastive-based CNN requires 10 times less labels to for fine-tuning on ImageNet compared to only using a supervised contrastive CNN (ResNet architecture). They find a strong correlation between the size of the pretrained network and the amount of labels it requires for fine-tuning. Finally, the constrastive network is distilled into a smaller version without sacrificing little classification accuracy. 

\paragraph{Relational Knowledge Distillation}
~\citet{park2019relational} apply knowledge distillation to relational data and propose distance (huber) and angular-based (cosine proximity) loss functions that account for different relational structures and claim that metric learning allows the student relational network to outperform the teacher network on achieving SoTA on relational datasets.

\begin{figure}
    \centering
    \includegraphics[scale=0.36]{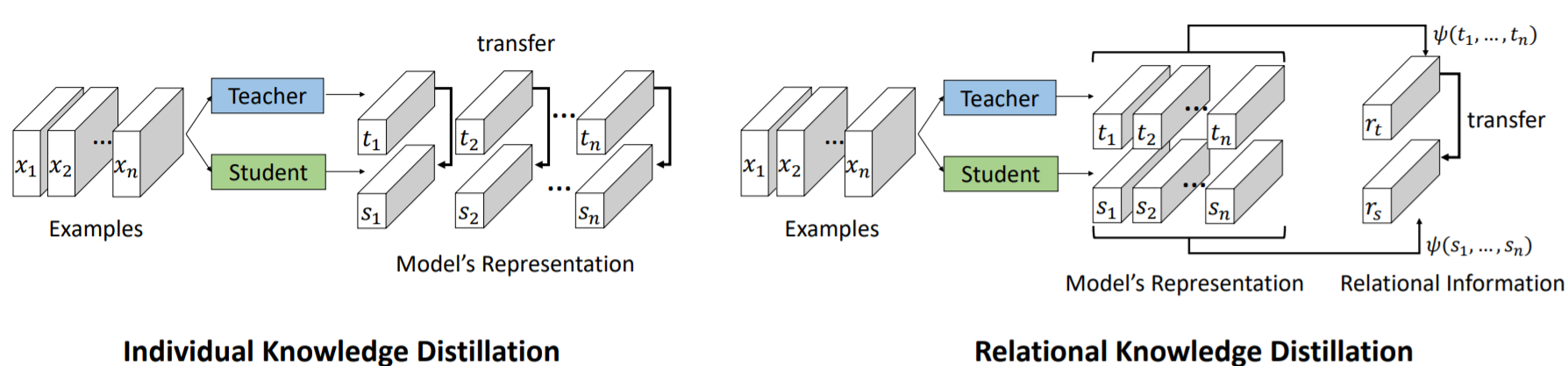}
    \caption{original source~\citet{park2019relational}: Individual knowledge distillation (IKD) vs. relational knowledge distillation (RKD)}
    \label{fig:rkd}
\end{figure}

The $\psi(\cdot)$ similarity function from the relation teacher network outputs a score that is transferred to as a pseudo target for the teacher network to learn from as,

\[
\delta(x, y) =
\begin{cases}
   \frac{1}{2} \sum_{i=1}^{N}(x - y)^2 & \text{for} \quad |x - y| \leq 1 \\
    |x - y| - 1  & \text{otherwise}
\end{cases}
\]

In the case of the angular loss shown in \autoref{eq:ang_loss}, $\vec{e}^{ij} = \frac{t_i - t_j}{|| t_i - t_j ||_2}$, $e^{kj} = \frac{t_k - t_j}{
||t_k - t_{j}||^2}$.

\begin{equation}\label{eq:ang_loss}
    \psi_A(t_i, t_j , t_k) = \cos \angle t_i t_j t_k = \langle \vec{e}_{ij}, \vec{e}_{kj} \rangle
\end{equation}

They find that measuring the angle between teacher and student outputs as input to the huber loss $\cL_{delta}$ leads to improved performance when compared to previous SoTA on metric learning tasks.

\begin{equation}\label{eq:ang_loss_2}
\cL_{rmd-a} = \sum_{(x_i,x_j ,x_k) \in \cX^3} l_{\delta} \psi_A(t_i, t_j , t_k), \psi_A(s_i, s_j , s_k)
\end{equation}

This is then used as a regularization terms to the task specific loss as,

\begin{equation}\label{eq:total_loss}
\cL_{\text{task}} + \lambda_{\text{MD}} \cL_{MD} 
\end{equation}

When used in metric learning the triplet loss shown in \autoref{eq:triplet_loss} is used. 

\begin{equation}\label{eq:triplet_loss}
\cL_{\text{triplet}} = \Big[ || f(\vec{x}_a) - f(\vec{x}_p) ||^2_2 - || f(\vec{x}_a) - f(\vec{x}_n)||^2_2 + m \Big]_{+}    
\end{equation}

\begin{wrapfigure}{R}{6.5cm}
    \centering
    \vspace{-1em}
    \includegraphics[scale=0.4]{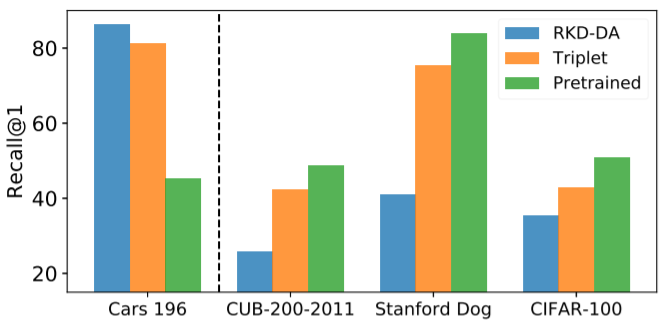}
    \caption{original source:~\citep{park2019relational}}
    \label{fig:recall_rd}
\end{wrapfigure}

\autoref{fig:recall_rd} shows the test data recall@1 on tested relational datasets. The teacher network is trained with the triplet loss and student distils the knowledge using \autoref{eq:total_loss}. Left of the dashed line are results on the training domain while on the right shows results on the remaining domains.

~\citet{song2018neural} use attention-based knowledge distillation for fashion matching that jointly learns to match clothing items while incorporating domain knowledge rules defined by clothing description where the attention learns to assign weights corresponding to the rule confidence.

\section{Quantization}\label{sec:quantize}
Quantization is the process of representing values with a reduced number of bits. In neural networks, this corresponds to weights, activations and gradient values. Typically, when training on the GPU, values are stored in 32-bit floating point (FP) single precision. \textit{Half-precision} for floating point (FP-16) and integer arithmetic (INT-16) are also commonly considered. INT-16 provides higher precision but a lower dynamic range compared to FP-16. In FP-16, the result of a multiplication is accumulated into a FP-32 followed by a down-conversion to return to FP-16. 


To speed up training, faster inference and reduce bandwidth memory requirements, ongoing research has focused on training and performing inference with lower-precision networks using integer precision (IP) as low as INT-8 INT-4, INT-2 or 1 bit representations~\citep{dally2015high}. Designing such networks makes it easier to train such networks on CPUs, FPGAs, ASICs and GPUs. 

Two important features of quantization are the range of values that can be represented and the bit spacing. For the range of signed integers with $n$ bits, we represent a range of [-2n-1, 2n-2] and for full precision (FP-32) the range is $+/- 3.4 \times 1038$. For signed integers, there $2n$ values in that range and approximately $4.2 \times 109$ for FP-32. FP can represent a large array of distributions which is useful for neural network computation, however this comes at larger computational costs when compared to integer values. For integers to be used to represent weight matrices and activations, a FP scale factor is often used, hence many quantization approaches involve a hybrid of mostly integer formats with FP-32 scaling numbers. This approach is often referred to as mixed-precision (MP) and different MP strategies have been used to avoid overflows during training and/or inference of low resolution networks given the limited range of integer formats. 

In practice, this often requires the storage of hidden layer outputs with full-precision (or at least with represented with more bits than the lower resolution copies). The main forward-pass and backpropogation is carried out with lower resolution copies and convert back to the full-precision stored ``accumulators'' for the gradient updates. 

In the extreme case where binary weights (-1, 1) or 2-bit ternary weights (-1, 0, 1) are used in fully-connected or convolutional layers, multiplications are not used, only additions and subtractions. For binary activations, bitwise operations are used~\citep{rastegari2016xnor} and therefore addition is not used. For example, ~\citet{rastegari2016xnor} proposed XNOR-Networks, where binary operations are used in a network made up of xnor gates which approximate convolutions leading to 58 times speedup and 32 times memory savings.


\subsection{Approximating High Resolution Computation}

Quantizing from FP-32 to 8-bit integers with retraining can result in an unacceptable drop in performance. Retraining quantized networks has shown to be effective for maintaining accuracy in some works~\citep{gysel2018ristretto}.
Other work~\citep{dettmers2015} compress gradients and activations from FP-32 to 8 bit approximations to maximize bandwidth use and find that performance is maintained on MNIST, CIFAR10 and ImageNet when parallelizing both model and data.  

The quantization ranges can be found using k-means quantization~\citep{lloyd1982least}, product quantization~\citep{jegou2010product} and residual quantization~\citep{buzo1980speech}. Fixed point quantization with optimized bit width can reduce existing networks significantly without reducing performance and even improve over the original network with retraining~\citep{lin2016fixed}. 

~\citet{courbariaux2014training} instead scale using shifts, eliminating the necessity of floating point operations for scaling. This involves an integer or fixed point multiplication, as apart of a dot product, followed by the shift. 

~\citet{dettmers2015} have also used FP-32 scaling factors for INT-8 weights and where the scaling factor is adapted during training along with the activation output range. They also consider not adapting the min-max ranges online and clip outlying values that may occur as a a result of this in order to drastically reduce the min-max range. They find SoTA speedups for CNN parallelism, achieving a 50 time speedup over baselines on 96 GPUs.

~\citet{gupta2015deep} show that stochastic rounding techniques are important for FP-16 DNNs to converge and maintain test accuracy compared to their FP-32 counterpart models. In stochastic rounding the weight $x$ is rounded to the nearest target fixed point representation $[x]$ with probability $1 - (x - [x])/\epsilon$ where $\epsilon$ is the smallest positive number representable in the fixed-point format, otherwise $x$ is rounded to $x + \epsilon$. Hence, if $x$ is close to $[x]$ then the probability is higher of being assigned $[x]$.~\citet{wang2018training} train DNNs with FP-8 while using FP-16 chunk-based accumulations with the aforementioned stochastic rounding hardware. 

The necessity of stochastic rounding, and other requirements such as loss scaling, has been avoided using customized formats such as Brain float point~\citep[(BFP)][]{kalamkar2019study} which use FP-16 with the same number of exponent bits as FP-32. ~\citet{cambier2020shifted} recently propose a shifted and squeezed 8-bit FP format (S2FP-8) to also avoid the need of stochastic rounding and loss scaling, while providing dynamic ranges for gradients, weights and activations. Unlike other related 8-bit techniques~\citep{mellempudi2019mixed}, the first and last layer do not need to be in FP32 format, although the accmulator converts the outputs to FP32. 

\subsection{Adaptive Ranges and Clipping}
\begin{wrapfigure}{R}{7cm}
    \centering
    \includegraphics[scale=0.26]{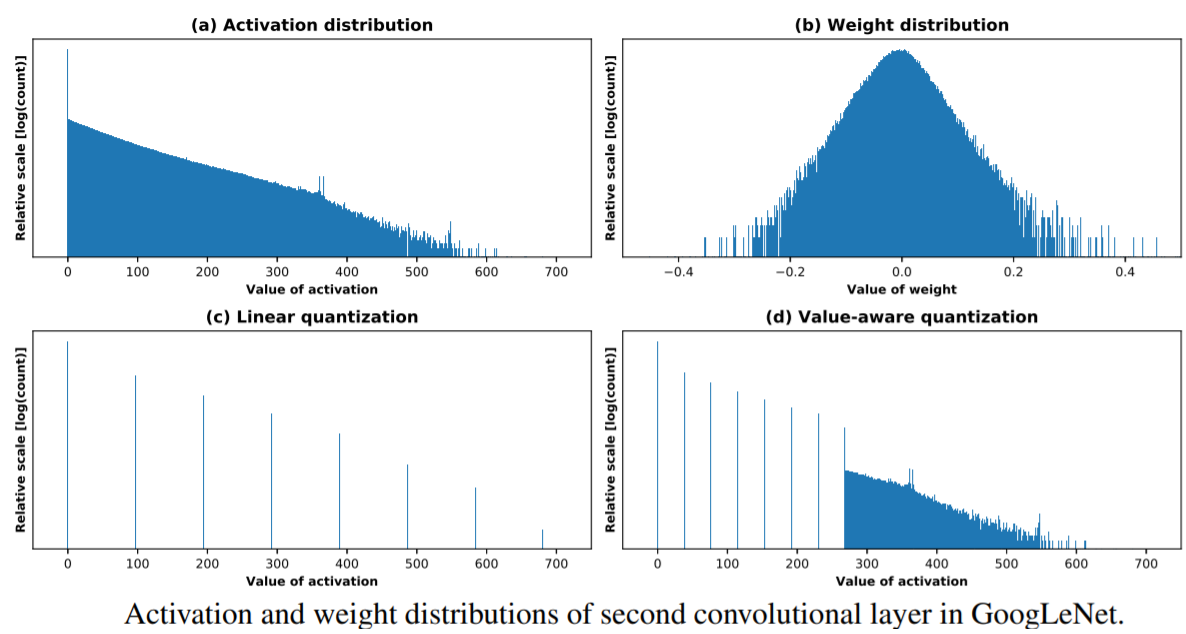}
    \caption{original source~\citet{park2018adversarial}: Weight and Activation Distributions Before and After Quantization}
    \label{fig:quant_weight_dist}
\end{wrapfigure}

~\citet{park2018value} exploit the fact that most the weight and activation values are scattered around a narrower region while larger values outside such region can be represented with higher precision. The distribution is demonstrated in \autoref{fig:quant_weight_dist}, which displays the weight distribution for the $2^{nd}$ layer in the LeNet CNN network. Instead of using linear quantization shown in (c), a smaller bit interval is used for the region of where most values lie (d), leading to less quantization errors. 

They propose 3-bit activations for training quantized ResNet and Inception CNN architectures during retraining. For inference on this retrained low precision trained network, weights are also quantized to 4-bits for inference with 1\% of the network being 16-bit scaling factor scalars, achieving accuracy within 1\% of the original network. This was also shown to be effective in LSTM network on language modelling, achieving similar perplexities for bitwidths of 2, 3 and 4. 

~\citet{migacz2017} use relative entropy to measure the loss of information between two encodings and aim minimize the KL divergence between activation output values. For each layer they store histgrams of activations, generate quantized distributions with different saturation thresholds and choose the threshold that minimizes the KL divergence between the original distribution and the quantized distribution. 

~\citet{banner2018aciq} analyze the tradeoff between quantization noise and clipping distortion and derive an expression for the mean-squared error degradation due to clipping. Optimizing for this results in choosing clipping values that improve 40\% accuracy over standard quantization of VGG16-BN to 4-bit integer. 

Another approach is to use scaling factors per group of weights (e.g channels in the case of CNNs or internal gates in LSTMs) as opposed to whole layers, particularly useful when the variance in weight distribution between the weight groupings is relatively high.

\subsection{Robustness to Quantization and Related Distortions}

~\citet{merolla2016deep} have studied the effects of different distortions on the weights and activations, including quantization, multiplicative noise (aking to Gaussian DropConnect), binarization (sign) along with other nonlinear projections and simply clipping the weights. This suggests that neural networks are robust to such distortions at the expense of longer convergence times.


In the best case of these distortions, they can achieve 11\% test error on CIFAR-10 with 0.68 effective bits per weight. They find that training with weight projections other than quantization performs relatively well on ImageNet and CIFAR-10, particularly their proposed stochastic projection rule that leads to 7.64\% error on CIFAR-10.

Others have also shown DNNs robustness to training binary and ternary networks~\citep{gupta2015deep,courbariaux2014training}, albeit a larger number of bit weight and ternary weights that are required.

\subsection{Retraining Quantized Networks}
Thus far, these post-training quantization (PTQ) methods without retraining are mostly effective on overparameterized models. For smaller models that are already restricted by the degrees of freedom, PTQ can lead to relatively large performance degradation in comparison to the overparameterized regime, which has been reflected in recent findings that architectures such as MobileNet suffer when using PTQ to 8-bit integer formats and lower~\citep{jacob2018quantization,krishnamoorthi2018quantizing}.

Hence, retraining is particularly important as the number of bits used for representation decreases e.g 4-bits with range [-8, 8]. However, quantization results in discontinuities which makes differentiation during backpropogation difficult. 

To overcome this limitation,~\citet{zhou2016dorefa} quantized gradients to 6-bit number and stochastically propogate back through CNN architectures such as AlexNet using straight through estimators, defined as \autoref{eq:ste_quant}. Here, a real number input $r_i \in [0, 1]$ to a n-bit number output $r_o \in [0, 1]$ and $\mathcal{L}$ is the objective function.

\begin{gather}\label{eq:ste_quant}
\textbf{\text{Forward}}: r_o = \frac{1}{2^n - 1} \text{round}((2n - 1)r_i) \\
\textbf{\text{Backward}}: \frac{\partial \mathcal{L}}{\partial r_i} = \frac{\partial \mathcal{L}}{\partial r_o}
\end{gather}

To compute the integer dot product of $r_0$ with another $n-bit$ vector, they use \autoref{eq:int_dp}, with a computational complexity of $\mathcal{O}(MK)$, directly proportional to bitwidth of $x$ and $y$. Furthermore, bitwise kernels can also be used for faster training and inference

\begin{gather}\label{eq:int_dp}
 x \cdot y = \sum_{m=0}^{M-1} \sum_{k=0}^{K-1}2^{m+k} \text{bitcount}[\text{and}(c_m(\vec{x}), c_k(\vec{y}))] \\
c_m(\vec{x})i, c_k(\vec{y}) \quad i \in \{0, 1\} \quad \forall i, m, k
\end{gather}

\paragraph{Model Distilled Quantization}

An overview of our incremental network quantization method. (a) Pre-trained full precision model used as a reference. (b) Model update with three proposed operations: weight partition, group-wise quantization (green connections) and re-training (blue connections). (c) Final low-precision model with all the weights constrained to be either powers of two or zero. In the figure, operation (1) represents a single run of (b), and operation (2) denotes the procedure of repeating operation (1) on the latest re-trained weight group until all the non-zero weights are quantized. Our method does not lead to accuracy loss when using 5-bit, 4-bit and even 3-bit approximations in network quantization. For better visualization, here we just use a 3-layer fully connected network as an illustrative example, and the newly re-trained weights are divided into two disjoint groups of the same size at each run of operation (1) except the last run which only performs quantization on the re-trained floating-point weights occupying 12.5\% of the model weights.

 \begin{figure}
     \centering
     \includegraphics[scale=0.35]{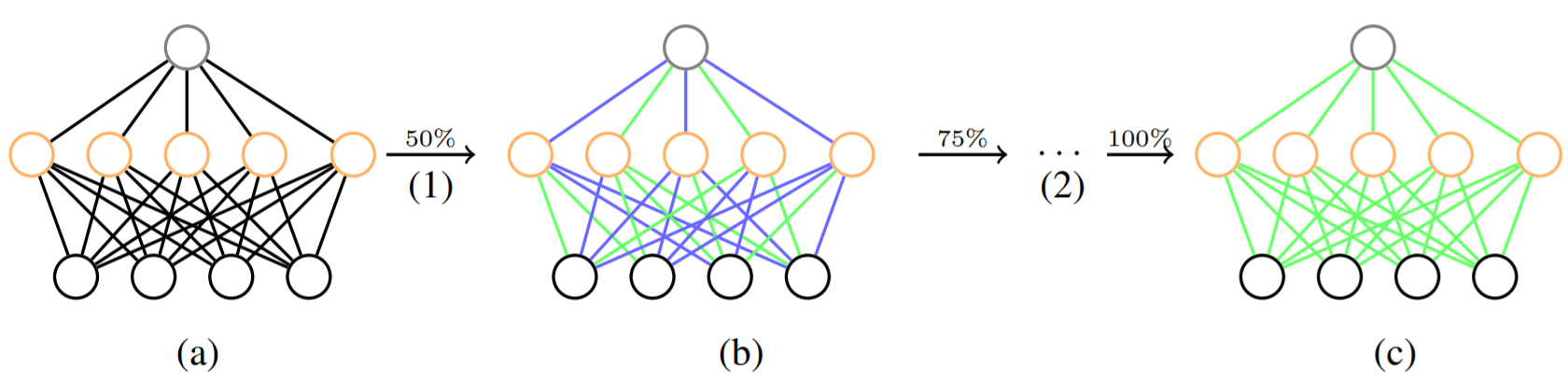}
     \caption{Quantized Knowledge Distillation (original source:~\citep{zhou2017incremental})}
     \label{fig:quant_mod_dist}
 \end{figure}

~\citet{polino2018model} use a distillation loss with respect to the teacher network whose weights are quantized to set number of levels and quantized  teacher trains the `student'. They also propose differentiable quantization, which optimizes the location of quantization points through stochastic gradient descent, to better fit the behavior of the teacher model.


\paragraph{Quantizing Unbounded Activation Functions} 
When the nonlinear activation unit used is not bounded in a given range, it is difficult to choose the bit range. Unlike sigmoid and tanh functions that are bounded in $[0, 1]$ and $[-1, 1]$ respectively, the ReLU function is unbounded in $[0, \infty]$. Obviously, simply avoiding such unbounded functions is one option, another is to clip values outside an upper bound~\citep{zhou2016dorefa,mishra2017wrpn} or dynamically update the clipping threshold for each layer and set the scaling factor for quantization accordingly~\citep{choi2018pact}.

\paragraph{Mixed Precision Training}
Mixed Precision Training (MPT) is often used to train quantized networks, whereby some values remain in full-precision so that performance is maintained and some of the aforementioned problems (e.g overflows) do not cause divergent training. It has also been observed that activations are more sensitive to quantization than weights~\citep{zhou2016dorefa}

~\citet{micikevicius2017mixed} use half-precision (16-bit) floating point accuracy to represent weights, activations and gradients, without losing model accuracy or having to modify hyperparameters, almost halving the memory requirements. They round a single-precision copy of the weights for forward and backward passes after performing gradient-updates,  use loss-scaling to preserve small magnitude gradient values and perform half-precision computation that accumulates into single-precision outputs before storing again as half-precision in memory.

\begin{figure}
    \centering
    \includegraphics[scale=0.3]{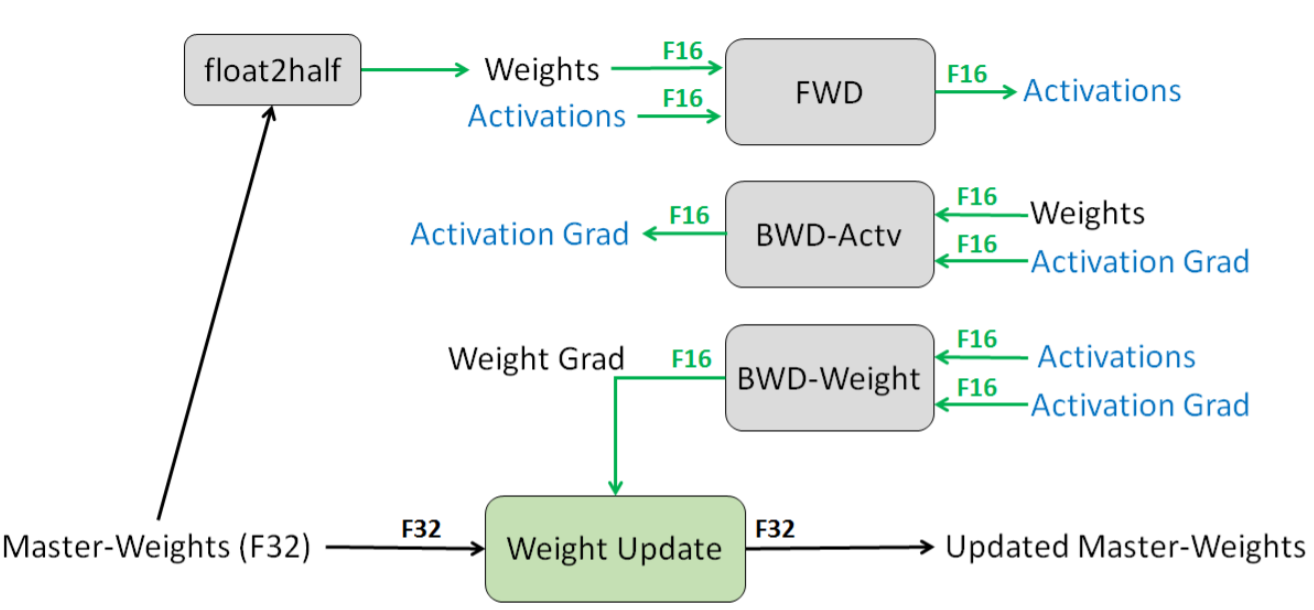}
    \caption{Mixed Precision Training (original source:~\citet{micikevicius2017mixed}}
    \label{fig:mpt}
\end{figure}

~\autoref{fig:mpt} illustrates MPT, where the forward and backward passes are performed with FP-16 precision copies. Once the backward pass is performed the computed FP-16 gradients are used to update the original FP-32 precision master weight. After training, the quantized weights are used for inference along with quantized activation units. This can be used in any type of layer, convolutional or fully-connected.

Others have focused solely on quantizing weights, keeping the activations at FP32 ~\citep{li2016ternary,zhu2016trained}.During gradient descent,~\citet{zhu2016trained} learn both the quantized ternary weights and pick which of these values is assigned to each weight, represented in a codebook.

~\citet{das2018mixed} propose using Integer Fused-Multiply-and-Accumulate (FMA) operations to accumulate results of multiplied INT-16 values into INT-32 outputs and use dynamic fixed point scheme to use in tensor operations. This involves the use of a shared tensor-wide exponent and down-conversion on the maximum value of an output tensor at each given training iteration using stochastic, nearest and biased rounding. They also deal with overflow by proposing a scheme that accumulates INT-32 intermediate results to FP-32 and can trade off between precision and length of the accumulate chain to improve accuracy on the image classification tasks. They argue that previous reported results on mixed-precision integer training report on non-SoTA architectures and less difficult image tasks and hence they also report their technique on SoTA architectures for the ImageNet 1K dataset. 


\paragraph{Quantizing by Adapting the Network Structure}
To further improve over mixed-precision training, there has been recent work that have aimed at better simulating the effects of quantization during training. 

~\citet{mishra2017apprentice} combine low bit precision and knowledge distillation using three different schemes: (1) a low-precision (4-bit) ResNet network is trained from a full-precision ResNet network both from scratch, (2) a full precision trained network is transferred to train a low-precision network from scratch and (3) a trained full-precision network guides a smaller full-precision student randomly initialized network which is gradually becomes lower precision throughout training. They find that (2) converges faster when supervised by an already trained network and (3) outperforms (1) and set at that time was SoTA for Resnet classifiers at ternary and 4-bit precision. 

 
~\citet{lin2017towards} replace FP-32 convolutions with multiple binary convolutions with various scaling factors for each convolution, overall resulting in a large range.
 
~\citet{zhou2016dorefa} and~\citet{choi2018pact} have both reported that the first and last convolutional layers are most sensitive to quantization and hence many works have avoided quantization on such layers. However,~\citet{choi2018pact} find that if the quantization is not very low (e.g 8-bit integers) then these layers are expressive enough to maintain accuracy.

~\citet{zhou2017incremental} have overcome this problem by iteratively quantizing the network instead of quantize the whole model at once. During the retraining of an FP-32 model, each layer is iteratively quantized over consecutive epochs. They also consider using supervision from a teacher network to learn a smaller quantized student network, combining knowledge distillation with quantization for further reductions.

\paragraph{Quantization with Pruning \& Huffman Coding}

\begin{wrapfigure}{r}{8cm}
    \centering
    \includegraphics[scale=0.35]{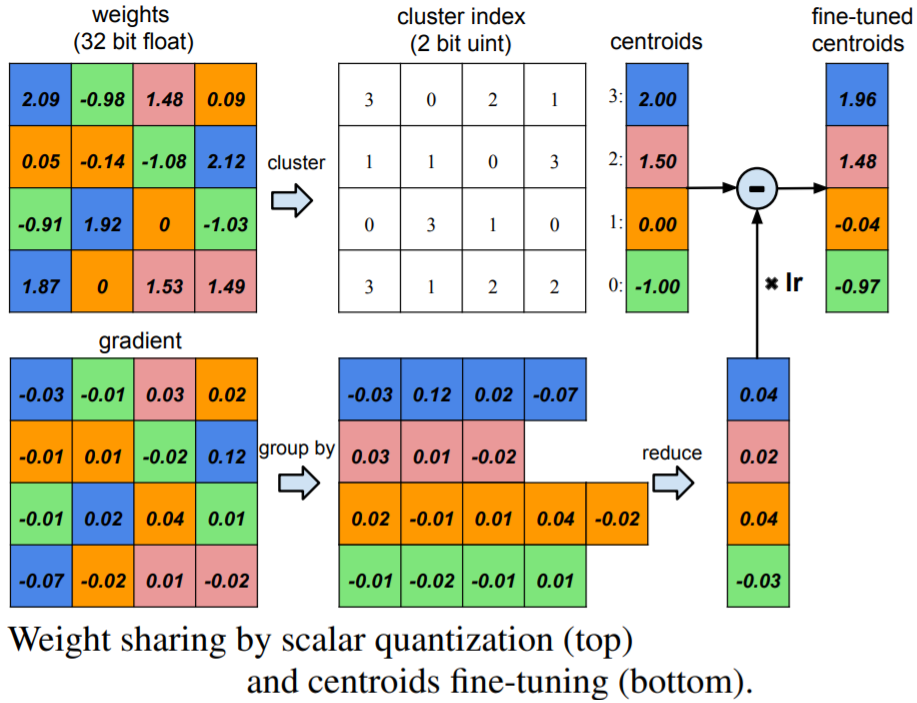}
    \caption{original source:~\citet{han2015deep}}
    \label{fig:sqft}
\end{wrapfigure}

Coding schemes can be used to encode information in an efficient manner and construct codebooks that represent weight values and activation bit spacing.~\citet{han2015deep} use pruning with quantization and huffman encoding for compression of ANNs by 35-49 times (9-13 times for pruning, quantization represents the weights in 5 bits instead of 32) the original size without affecting accuracy.

Once the pruned network is established, the parameter are quantized to promote parameter sharing. This multi-stage compression strategy is illustrated in \autoref{fig:sqft}, showing the combination of weight sharing (top) and fine-tuning of centroids (bottom). They note that too much pruning on channel level sparsity (as opposed to kernel-level) can effect the network's representational capacity.

\subsubsection{Loss-aware quantization}
~\citet{hou2016loss} propose a proximal Newton algorithm with a diagonal Hessian approximation to minimize the loss with respect to the binarized weights $\hat{\vec{w}} = \alpha \vec{b}$, where $\alpha > 0$ and $\vec{b}$ is binary. During training, $\alpha$ is computed for the $l$-th layer at the $t$-th iteration as $\alpha^t_l = || \vec{d}^{t - 1} \otimes \vec{w}^t_l ||_1 / || \vec{d}^{t - 1 }_l ||_1$ where $\vec{d}^{t - 1}_l := \text{diag}(\mat{D}^{t - 1}_{l})$  and $\vec{b}^{t}_l = \sign(\vec{w}^t_l)$. The input is then rescaled for layer $l$ as  $\tilde{\vec{x}}^t_l = \alpha^t_l \vec{x}^t_{l-1}$ and then compute $\vec{z}^t_l$ with input $\tilde{x}^t_{l-1}$ and binary weight $\vec{b}^{t}_l$.

~\autoref{eq:proximal_newton} shows the proximal newton update step where $w^{t}_{l}$ is the weight update at iteration $t$ for layer $l$, $\mat{D}$ is an approximation to the diagonal of the Hessian which is already given as the $2^{nd}$ momentum of the adaptive momentum (adam) optimizer. The $t$-th iteration of the proximal Newton update is as follows:

\begin{equation}\label{eq:proximal_newton}
\begin{split}
\min_{\hat{\vec{w}}^t} \nabla \ell(\hat{w}^{t-1})^{T}(\hat{\vec{w}}^{t} - \hat{\vec{w}}^{t-1}) + (\hat{\vec{w}}^{t} - \hat{\vec{w}}^{l-1}) D^{t-1} (\hat{\vec{w}}^{t} - \hat{\vec{w}}^{t-1}) \\
s.t.\hat{\vec{w}}_t^l= \alpha^t_l \vec{b}^t_l, \alpha^t_l>0, \ \vec{b}^t_l \in \{+/-1\}n_l,  l= 1,\ldots, L.  
\end{split}
\end{equation}

where the loss $\ell$ w.r.t binarized version of $\ell(w_t)$ is expressed in terms of the $2^{nd}$-order TS expansion using a diagonal approximation of the Hessian $\mat{H}^{t-1}$, which estimates of the Hessian at $\vec{w}^{t-1}$. Similar to the $2^{nd}$ order approximations discussed in~\autoref{eq:weight_reg}, the Hessian is essential since $\ell$ is often flat in some directions but highly curved in others.


\paragraph{Explicit Loss-Aware Quantization}
~\citet{zhou2018explicit} propose an Explicit Loss-Aware Quantization (ELQ) method that minimizes the loss perturbation from quantization in an incremental way for very low bit precision i.e binary and ternary. Since going from FP-32 to binary or ternary bit representations can cause considerable fluctuations in weight magnitudes and in turn the predictions, ELQ directly incorporates this quantization effect in the loss function as

\begin{equation}
    \min_{\hat{\mat{W}}_{l}} + a_1 \cL_p (\mat{W}_l, \hat{\mat{W}}_l) + E(\mat{W}_l, \hat{\mat{W}}_l) \quad s.t. \hat{W} \in \{a_l c_k | 1 \leq k \leq K \}, \ 1 \leq l \leq L
\end{equation}

where $L_p$ is the loss difference between quantized and the original model $||\cL(\mat{W}_l) - \cL(\hat{\mat{W}}_l)||$, $E$ is the reconstruction error between the quantized and original weights $||\mat{W}_l - \hat{\mat{W}_l}||^2$, $a_l$ a regularization coefficient for the $l$-th layer and $c_k$ is an integer and $k$ is the number of weight centroids.

\paragraph{Value-aware quantization}
~\citet{park2018value} like prior work mentioned in this work have also succeeded in reduced precision by reducing the dynamic range by narrowing the range where most of the weight values concentrate. Different to other work, they assign higher precision to the outliers as opposed to mapping them to the extremum of the reduced range. This small difference allow 3-bit activations to be used in ResNet-152 and DenseNet-201, leading to a 41.6\% and 53.7\% reduction in network size respectively.

\subsubsection{Differentiable Quantization}
When considering fully-differentiable training with quantized weight and activations, it is not obvious how to back-propagate through the quantization functions. These functions are discrete-valued, hence their derivative is 0 almost everywhere. So, using their gradients as-is would severely hinder the learning process. 
A commonly used approximation to overcome this issue is the ``straight-through estimator'' (STE)~\citep{hinton2012neural,bengio2013estimating}, which simply passes the gradient through these functions as-is, however there has been a plethora of other techniques proposed in recent years which we describe below. 

\begin{figure}[!bp]
\centering
\begin{subfigure}{.5\textwidth}
  \centering
    \includegraphics[width=1.0\textwidth]{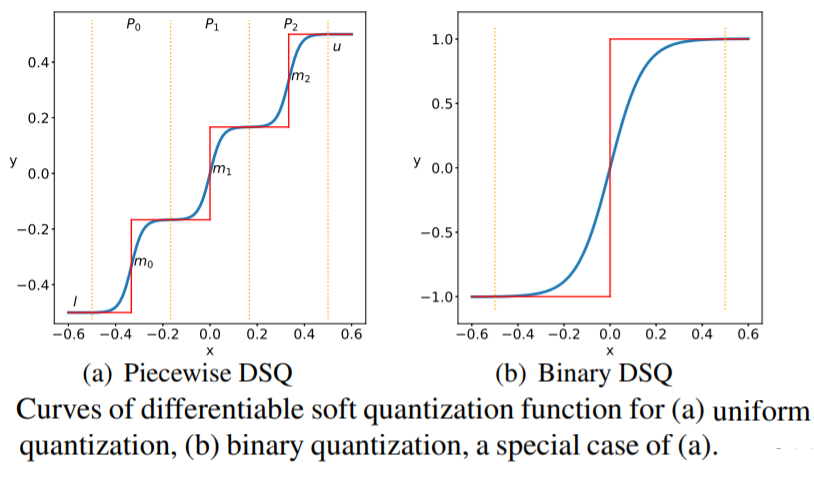}
  \caption{original source:~\citet{gong2019differentiable}}
  \label{fig:diff_soft_quant}
\end{subfigure}%
\begin{subfigure}{.5\textwidth}
  \centering
    \includegraphics[width=1.0\textwidth]{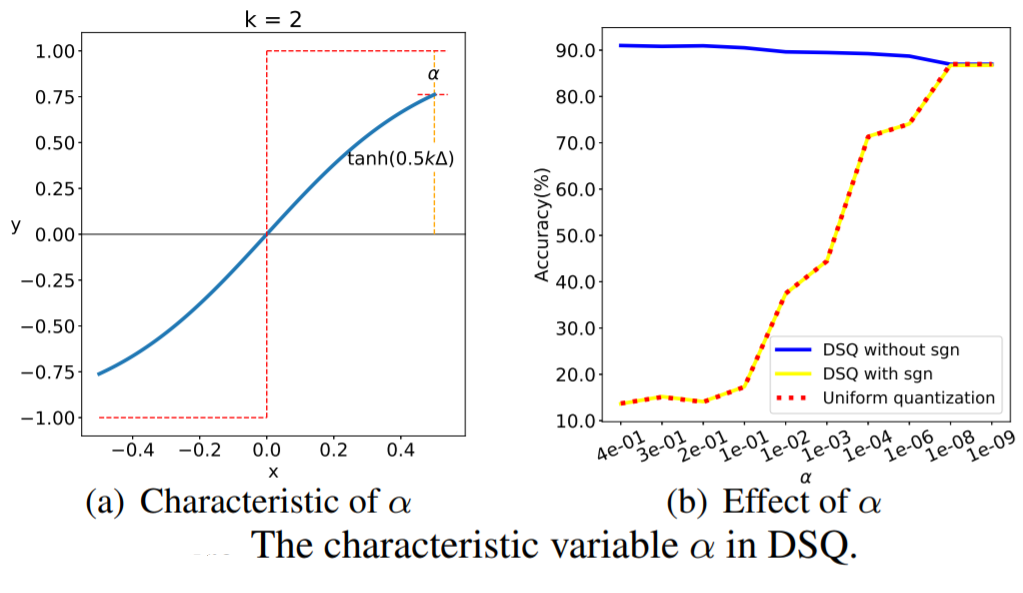}
  \caption{original source:~\citet{gong2019differentiable}}
  \label{fig:dsq}
\end{subfigure}
\caption{Differentiable Soft Quantization}
\label{fig:test}
\end{figure}

\paragraph{Differentiable Soft Quantization}

~\citet{gong2019differentiable} have proposed differentiable soft quantization (DSQ) learn clipping ranges in the forward pass and approximating gradients in the backward pass. To approximate the derivative of a binary quantization function, they propose a differentiable asymptotic function (i.e smooth) which is closer to the quantization function that it is to a full-precision $\mathtt{tanh}$ function and therefore will result in less of a degradation in accuracy when converted to the binary quantization function post-training.

For multi-bit uniform quantization, given the bit width $b$ and the floating-point activation/weight $\vec{x}$ following in the range $(l, u)$, the complete quantization-dequantization process of uniform quantization can be defined as:
$Q_U(\vec{x}) = \text{round}(\vec{x} \Delta)\Delta$ where the original range $(l, u)$ is divided into $2^b - 1$ intervals $\cP_i, i \in (0, 1, \ldots 2^b - 1)$, and $\Delta = \frac{u-l}{2^b-1}$ is the interval length.

The DSQ function, shown in \autoref{eq:soft_quant_tanh}, handles the point $\vec{x}$ depending what interval in $\cP_i$ lies.

\begin{equation}\label{eq:soft_quant_tanh}
\phi(x) = s \tanh (k(\vec{x} - m_i)), \quad \text{if} \quad \vec{x} \in \mathcal{P}_i
\end{equation}
\vspace{-0.5em}
with
\vspace{-0.5em}
\begin{equation}
m_i = l + (i + 0.5)\Delta \quad \text{and} \quad s = 1 \tanh(0.5 k \Delta)     
\end{equation}

The scale parameter $s$ for the tanh function $\varphi$ ensures a smooth transitions between adjacent bit values, while $k$ defines the functions shape where large $k$ corresponds close to consequtive step functions given by uniform quantization with multiple piecewise levels, as shown in \autoref{fig:diff_soft_quant}. The DSQ function then approximates the uniform quantizer $\varphi$ as follows:

\[
    \mathcal{Q}_S(\vec{x}) = 
\begin{cases}
   l, & \vec{x} < l, \\
   u, & \vec{x} > u, \\ 
   l + \Delta (i + \frac{\varphi(x) + 1}{2} ), & \vec{x} \in \mathcal{P}_i
\end{cases}
\]

The DSQ can be viewed as aligning the data with the quantization values with minimal quantization error due to the bit spacing that is carried out to reflect the weight and activation distributions. \autoref{fig:dsq} shows the DSQ curve without [-1, 1] scaling, noting standard quantization is near perfectly approximated when the largest value on the curve bounded by +1 is small. They introduce a characteristic
variable $\alpha: = 1 - \tanh(0.5 k \Delta) = 1 - \frac{1}{s}$ and given that,

\begin{gather}
\Delta = \frac{u-l}{2^b - 1} \\
\varphi(0.5\Delta) = 1 \quad \Rightarrow \quad k = \frac{1}{\Delta}\log( 2/\alpha - 1)
\end{gather}

DSQ can be used as a piecewise uniform quantizer and when only one interval is used, it is the equivalent of using DSQ for binarization.

\paragraph{Soft-to-hard vector quantization}


~\citet{agustsson2017soft} propose to compress both the feature representations and the model by gradually transitioning from soft to hard quantization during retraining and is end-to-end differentiable. They jointly learn the quantization levels with the weights and show that vector quantization can be improved over scalar quantization. 


\begin{equation}
    H(E(\vec{Z})) = - X_{e \in [L]} m P(E(\vec{Z}) = e) \log(P(E(\vec{Z}) = e))
\end{equation}

They optimize the rate distortion trade-off between the expected loss and the entropy of $\mathbb{E}(Z)$:

\begin{equation}
\min_{E,D,\vec{W}}\mathbb{E}_{X,Y}[\ell(\hat{F}(X), Y) + \lambda R(\mat{W})] + \beta H(E(\vec{Z}))    
\end{equation}

\paragraph{Iterative Product Quantization (iPQ)}
Quantizing a whole network at once can be too severe for low precision (< 8 bits) and can lead to \textit{quantization drift} - when scalar or vector quantization leads to an accumulation of reconstruction errors within the network that compound and lead to large performance degradations. To combat this,~\citet{stock2019and} iteratively quantize the network starting with low layers and only performing gradient updates on the rest of the remaining layers until they are robust to the quantized layers. This is repeated until quantization is carried out on the last layer, resulting in the whole network being amenable to quantization. 
The codebook is updated by averaging the gradients of the weights within the block $b_{\text{KL}}$ as

\begin{equation}
    \vec{c} \leftarrow \vec{c} - \eta \frac{1}{|J_c|} \sum_{(k, l) \in J_c} \frac{\partial \mathcal{L}}{ \partial b_{\text{KL}}} \quad \text{where} \quad J_{\vec{c}} = \{ (k,l) \ | \ c[\mat{I}_{\text{KL}}] = \vec{c} \}
\end{equation}

where $\mathcal{L}$ is the loss function, $I_{\text{KL}}$ is an index for the $(k, l)$ subvector and $\eta > 0$ is the codebook learning rate. This adapts the upper layers to the drift appearing in their inputs, reducing the impact of the quantization approximation on the overall performance.

\paragraph{Quantization-Aware Training}

\begin{figure}
    \centering
    \includegraphics[scale=.35]{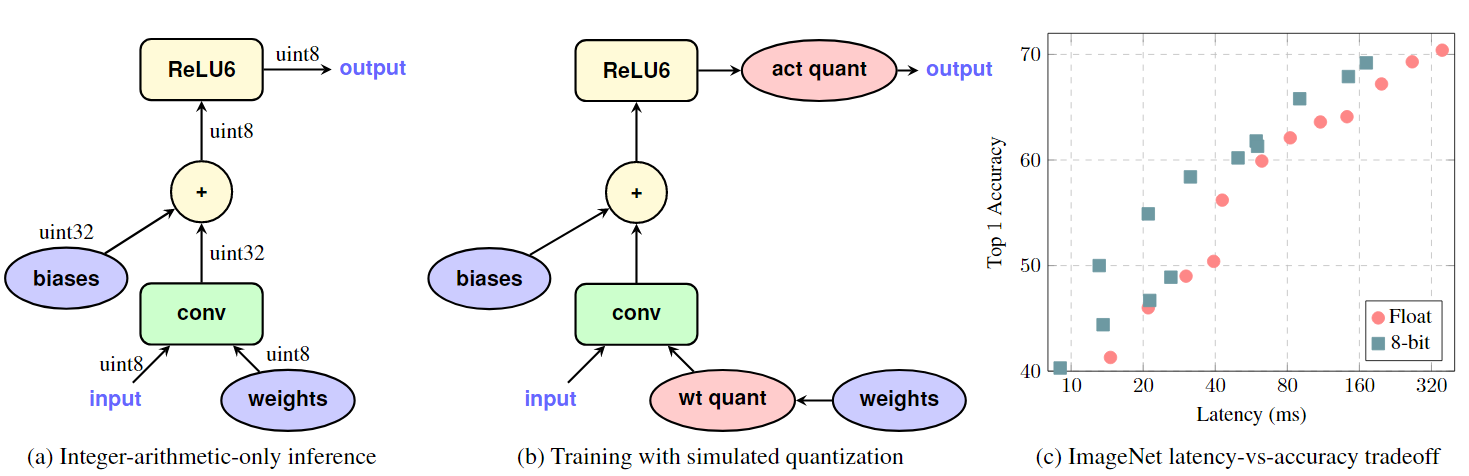}
    \caption{original source (~\citep{jacob2018quantization}): Integer-arithmetic-only quantization}
    \label{fig:iaoq}
\end{figure}

Instead of iPQ,~\citet{jacob2018quantization} use a straight through estimator ~\citep[(STE)][]{bengio2013estimating} to backpropogate through quantized weights and activations of convolutional layers during training.~\autoref{fig:iaoq} shows the 8-bit weights and activations, while the accumulator is represented in 32-bit integer.

They also note that in order to have a challenging architecture to compress, experiments should move towards trying to compress architectures which are already have a minimal number of parameter and perform relatively well to much larger predecessing architectures e.g EfficientNet, SqueezeNet and ShuffleNet.

\paragraph{Quantization Noise}
~\citet{fan2020training} argue that both iPQ and QAT are less suitable for very low precision such as INT4, ternary and binary. They instead propose to randomly simulate quantization noise on a subset of the network and only perform backward passes on the remaining weights in the network. Essentially this is a combination of DropConnect (instead of the Bernoulli function, it is a quantization noise function) and Straight Through Estimation is used to backpropogate through the sample of subvectors chosen for quantization for a given mini-batch.

Estimating quantization noise through randomly sampling blocks of weights to be quantized allows the model to become robust to very low precision quantization without being too severe, as is the case with previous quantization-aware training~\citep{jacob2018quantization}. The authors show that this iterative quantization approach allows large compression rates in comparison to QAT while staying close to (few perplexity points in the case of language modelling and accuracy for image classification) the uncompressed model in terms of performance. They reach SoTA compression and accuracy tradeoffs for language modelling (compression of Transformers such as RoBERTa on WikiText) and image classification (compressing EfficientNet-B3 by 80\% on ImageNet).

\paragraph{Hessian-Based Quantization}
The precision and order (by layer) of quantization has been chosen using $2^{nd}$ order information from the Hessian~\citep{dong2019hawq}. They show that on already relatively small CNNs (ResNet20, Inception-V3, SqueezeNext) that Hessian Aware Quantization (HAWQ) training leads to SoTA compression on CIFAR-10 and ImageNet with a compression ratio of 8 and in some cases exceed the accuracy of the original network with no quantization. 

Similarly,~\citet{shen2019q} quantize transformer-based models such as BERT with mixed precision by also using $2^{nd}$ order information from the Hessian matrix. They show that each layer exhibits varying amount of information and use a sensitivity measure based on mean and variance of the top eigenvalues. They show the loss landscape as the two most dominant eigenvectors of the Hessian are perturbed and suggest that layers that shower a smoother curvature can undergo lower but precision. In the cases of MNLI and CoNLL datasets, upper layers closer to the output show flatter curvature in comparison to lower layers. From this observation, they are motivated to perform a group-wise quantization scheme whereby blocks of a matrix have different amounts of quantization with unique quantization ranges and look up table. A Hessian-based mixed precision scheme is then used to decide which blocks of each matrix are assigned the corresponding low-bit precisions of varying ranges and analyse the differences found for quantizing different parts of the self-attention block (self-attention matrices and fully-connected feedforward layers) and their inputs (embeddings) and find the highest compression ratios can be attributed to most of the parametes in the self-attention blocks.

\section{Summary}

The above sections have provided descriptions of old and new compression methods and techniques. We finish by providing general recommendations for this field and future research directions that I deem to be important in the coming years. 

\subsection{Recommendations}
\paragraph{Old Baselines May Still Be Competitive}
Evidently, there has been an extensive amount of work in pruning, quantization, knowledge distillation and combinations of the aforementioned for neural networks. We note that many of these approaches, particularly pruning, were proposed in decades past~\citep{cleary1984data,mozer1989skeletonization,hassibi1994optimal,lecun1990optimal,whitley1990genetic,karnin1990simple,reed1993pruning,fahlman1990cascade}. The current trend of deep neural networks growing ever larger means that keeping track of new innovations on the topic of reducing network size becomes increasingly important. Therefore, we suggest that comparing past and present techniques for compression should be  standardized across models, datasets and evaluation metrics such that these comparisons are made direct. Ideally this would be carried out using the same libraries in the same language (e.g PyTorch or Tensorflow in Python) to further minimize any implementation differences that naturally occur. 

\paragraph{More Compression Work on Large Non-Sparse Architectures}
The majority of the aforementioned compression techniques that have been proposed are the context of CNNs since they have been used extensively over the past 3 decades, predominantly for image-based tasks. We suggest that future and existing techniques can now also be extended to recent architectures such as Transformers and applied to other important tasks (e.g text generation, speech recognition). In fact, this already becoming apparent by the rise in the
number of papers around compressing transformers in the NLP community. More specifically, reducing the size BERT and related models, as discussed in \autoref{sec:distil_trans}.

\paragraph{Challenging Compression on Already Parameter Efficient Architectures}
The importance of trying to compress already parameter efficient architectures (e.g EffecientNet, SqueezNet or MobileNet for CNNs or DistilBERT for Transformers), such as those discussed in \autoref{sec:lra}, makes for more challenging compression problem. Although compressing large overparameterized network have a large and obvious capacity for compression, compressing already parameter efficient network provides more insight into the advantages and disadvantages of different compression techniques.


\subsection{Future Research Directions}
The field of neural network compression has seen a resurgence in activity given the growing size of state of the art of models that are pushing the boundaries of hardware and practitioners resources. However, compression techniques are still in a relatively early stage of development. Below, I discuss a few research directions I think are worth exploring for the future of model compression.

\paragraph{\textit{What Combination of Compression Techniques To Use ?}}

Most of the works discussed here have not used multiple compression techniques for retraining (e.g pruning with distillation and quantization) nor have they figured out what order is optimal for a given set of tasks and architectures.~\citet{han2015deep} is a prime example of combining compression techniques, combining quantization, pruning and huffman coding. However, it still remains unclear what combination and what order should be used to get the desired compression tradeoff between performance, speed and storage. A strong ablation study on many different architectures with various combinations and orders would be greatly insightful from a practical standpoint. 

\paragraph{Automatically Choosing Student Size in Knowledge Distillation}
Current knowledge distillation approaches use fixed sized students during retraining. However, to get the desired tradeoff between performance versus student network size it requires a manual iteration over different student size in retraining. This is often used to visualize this tradeoff in papers, however automatically searching for student architecture during knowledge distillation is certainly an area of future research worth considering. In this context, meta learning and neural architecture search becomes important topics to bridge this gap between manually found student architectures to automatic techniques for finding the architectures. 

\paragraph{Few-Shot Knowledge Distillation}
In cases where larger pretrained models are required for a set (or single) of target tasks where there are only few samples, knowledge distillation can be used to distill the knowledge of teacher specifically for that transfer domain. The advantage of doing so is that we benefit from the transferability of teacher network while also distilling these large feature sets into a smaller network. 

\paragraph{Meta Learning Based Compression}
Meta learning~\citep{schmidhuber1987evolutionary,andrychowicz2016learning} have been successfully used for \textit{learning to learn}. Meta-learning how these larger teacher networks learn could be beneficial for improving the performance and convergence of a distilled student network. To date, I believe this is an unexplored area of research.


\paragraph{Further Theoretical Analysis}
Recent work has aided in our understanding of generalization in deep neural networks~\citep{neyshabur2018towards,wei2018margin,nakkiran2019deep,belkin2019reconciling,derezinski2019exact,saxe2019information} and proposed measures for tracking generalization performance while training DNNs. Further theoretical analysis of compression generalization is a worthwhile endeavour considering the growing importance and usage of compressing already trained neural networks. This is distinctly different than training models from random initialization and requires a new generalization paradigm to understand how compression works for each type (i.e pruning, quantization etc.).


\bibliography{jmlr2e}

\appendix
\appendixpage

\section{Low Resource and Efficient CNN Architectures}\label{sec:lra}

\subsubsection{MobileNet}
~\citet{howard2017mobilenets} propose compression of convolutional neural networks for embedded and mobile vision applications using depth-wise separable convolutions (DSC) and use two hyperparameters that tradeoff latency and accuracy. DSCs factorize a standard convolution into a depthwise convolution and $1\times 1$ pointwise convolution. Each input channel is passed through a DSC filter followed by a pointwise $1\times 1$ convolution that combines the outputs of the DSC. Unlike standard convolutions, DSCs split the convolution into two steps, first filtering then combining outputs of each DSC filter, which is why this is referred to as a factorization approach. 

Experiments on ImageNet image classification demonstrated that these smaller networks can achieve ac curacies similar to much larger networks.

\subsubsection{SqueezeNet}
~\citet{iandola2016squeezenet} reduce the network architecture by reducing $3\times 3$ filters to $1 \times 1$ filters (squeeze layer), reduce the number of input channels to $3 \times 3$ filters using squeeze layers and downsample later in the network to avoid the bottleneck of information through the network too early and in turn lead to better performance. A \textit{fire} module is made up of the squeeze layer is into an \textit{expand} layer that is a mix of $1 \times 1$ and $3\times 3$ convolution filters and the number of filters per \textit{fire} module is increased as it gets closer to the last layer. 

By using these architectural design decisions, SqueezeNet can compete with AlexNet with 50 times smaller network and even outperforms layer decomposition and pruning for deep compression. When combined with INT8 quantization,  SqueezeNet yields a 0.66 MB model which is 363 times smaller than 32-bit AlexNet, while still maintaining performance.

\subsubsection{ShuffleNet}
ShuffleNet~\citep{zhang2018shufflenet} uses pointwise group convolutions~\citep{krizhevsky2012imagenet}(i.e using a different set of convolution filter groups on the same input features, this allows for model parallelization) and channel shuffles (randomly shuffling helps information flow across feature channels) to reduce compute while maintaining accuracy. 
ShuffleNet is made up economical $3\times 3$ depthwise convolutional filters and replace $1 \times 1$ layer with pointwise group convolutional followed by the channel shuffle. 
Unlike predecessor models~\citep{xie2017aggregated,chollet2017xception}, ShuffleNet is efficient for smaller networks as they find big improvements when tested on ImageNet and MSCOCO object detection using 40 Mega FLOPs and achieves 13 times faster training over AlexNet without sacrificing much accuracy.  

\subsubsection{DenseNet}
Gradients can vanish in very deep networks because the error becomes more difficult to backpropogate as the number of matrix multiplications increase. DenseNets~\citep{huang2017densely} address gradient vanishing connecting the feature maps of the previous layer to the inputs of the next layer, similar to ResNet skip connections. This reusing of features mean the network efficient with its use of parameters. Although, deep and thin DenseNetworks can be parameter efficient, they do tradeoff with memory/speed efficiency in comparison to shallower yet wider network (~\citep{zagoruyko2016wide}) because all layer outputs need to be stored to perform backpropogation. However, DenseNets too can be made wider and shallower to become more memory effecient if required.  

%

%

\subsubsection{Fast Sparse Convolutional Networks}

~\citet{elsen2019fast} replace dense convolutional layers with sparse ones by introducing efficient sparse kernels for ARM and WebAssembly and show that sparse versions of  MobileNet v1, MobileNet v2 and EfficientNet architectures substantially outperform strong dense baselines on the efficiency-accuracy curve. 

\section{Low Resource and Efficient Transformer Architectures}\label{sec:lra_transformer}
In this section we describe some work that tries to find efficient architectures during training and hence are not considered compressed networks in the traditional definition as they are not already pretrained before the network is reduced. 

\paragraph{Transformer Architecture Search}
Most neural architecture search (NAS) methods learn to apply modules in the network with no regard for the computational cost of adding them, such as Neural architecture optimization~\citep{luo2018neural} which uses an encoder-decoder model to reconstruct an architecture from a continuous space.~\citet{guo2019nat} instead have proposed to learn a transformer architecture while minimizing the computational burden, avoiding modules with large number of parameters if necessary. However, solving such problem is NP-hard. Therefore, they propose to treat the optimization problem as a Markov Decision Process (MDP) and optimize the policies w.r.t. to the different architectures using reinforcement learning. These different architectures are replace redundant transformations with more efficient ones such as skip connections or removing connections altogether.

\end{document}